\documentclass[10pt,twocolumn,letterpaper]{article}

\usepackage[pagenumbers]{cvpr} 

\usepackage[cachedir=.]{minted}
\usepackage[most]{tcolorbox}
\tcbuselibrary{minted}
\usepackage{helvet}
\setminted{
  fontsize=\footnotesize,
  breaklines=true,
  breakanywhere=true
}
\usepackage{cuted}
\usepackage[T1]{fontenc}
\tcbuselibrary{skins,breakable,listings}

\newtcblisting{evaluationprompt}{
title={Evaluation Prompt},
enhanced,
breakable,         
colback=black!2,
colframe=black!60,
boxrule=0.4pt,
listing only,
width=\textwidth,    
listing engine=minted,
minted language=text,
minted options={
    breaklines=true,
    breakanywhere=true,
    fontsize=\footnotesize,
    tabsize=2,
    obeytabs=true,
    autogobble=false,
    showspaces=false,
    showtabs=false,
    fontfamily=phv
}
}

\newtcblisting{preprocessprompt}{
title={Preprocess Prompt},
enhanced,
breakable,      
colback=black!2,
colframe=black!60,
boxrule=0.4pt,
listing only,
width=\textwidth,   
listing engine=minted,
minted language=text,
minted options={
    breaklines=true,
    breakanywhere=true,
    fontsize=\footnotesize,
    tabsize=2,
    obeytabs=true,
    autogobble=false,
    showspaces=false,
    showtabs=false,
    fontfamily=phv
}
}

\newtcblisting{captionconcepts}{
  title={MeViS Caption-Concept Pairs},
  enhanced,
  breakable,         
  colback=black!2,
  colframe=black!60,
  boxrule=0.4pt,
  listing only,
  width=\textwidth,    
  listing engine=minted,
  minted language=text,
  minted options={
    breaklines=true,
    breakanywhere=true,
    fontsize=\tiny,
    tabsize=2,
    obeytabs=true,
    autogobble=false,
    showspaces=false,
    showtabs=false,
    fontfamily=phv
  }
}

\newtcbox{\boxlabel}{
  on line,       
  colback=black!60,   
  colframe=black!60,    
  coltext=white,    
  boxrule=0.4pt,
  arc=2pt,           
  boxsep=1pt,
  left=2pt,
  right=2pt,
  top=1pt,
  bottom=1pt
}

\newtcblisting{sanavideoprompt}{
title={SANA-Video Prompt Enhancement},
enhanced,
breakable, 
colback=black!2,
colframe=black!60,
boxrule=0.4pt,
listing only,
width=\textwidth,
listing engine=minted,
minted language=text,
minted options={
    breaklines=true,
    breakanywhere=true,
    fontsize=\footnotesize,
    tabsize=2,
    obeytabs=true,
    autogobble=false,
    showspaces=false,
    showtabs=false,
    fontfamily=phv
}
}

\definecolor{cvprblue}{rgb}{0.21,0.49,0.74}
\definecolor{imapred}{rgb}{0.90,0.10,0.10}
\definecolor{deepcvprblue}{rgb}{0.05,0.39,0.79}
\definecolor{imapblue}{rgb}{0.04,0.19,0.78}
\usepackage[pagebackref,breaklinks,colorlinks,allcolors=cvprblue]{hyperref}

\usepackage{multirow}
\usepackage{multicol}
\usepackage{booktabs}
\usepackage{xcolor}
\usepackage[x11names]{xcolor}
\usepackage{colortbl}
\usepackage{kotex}
\usepackage{setspace}
\usepackage{caption}

\newcommand{\ourparagraph}[1]{\vspace{3pt}\noindent\textbf{#1}}
\newcommand*{\tran}{^{\mkern-1.5mu\mathsf{T}}}
\usepackage{algorithm}
\usepackage{algpseudocode}

\usepackage{amsmath,amsfonts,bm}

\def\eqref#1{equation~\ref{#1}}

\def\1{\bm{1}}

\def\vc{{\bm{c}}}

\def\vh{{\bm{h}}}

\def\vk{{\bm{k}}}

\def\vq{{\bm{q}}}

\def\vv{{\bm{v}}}

\def\mA{{\bm{A}}}

\def\mG{{\bm{G}}}

\def\mK{{\bm{K}}}

\def\mQ{{\bm{Q}}}

\def\mV{{\bm{V}}}

\DeclareMathAlphabet{\mathsfit}{\encodingdefault}{\sfdefault}{m}{sl}
\SetMathAlphabet{\mathsfit}{bold}{\encodingdefault}{\sfdefault}{bx}{n}

\usepackage{tikz}
\newcommand{\xmark}{%
\tikz[scale=0.23] {
    \draw[line width=0.7,line cap=round] (0,0) to [bend left=6] (1,1);
    \draw[line width=0.7,line cap=round] (0.2,0.95) to [bend right=3] (0.8,0.05);
}}

\newcommand{\ourmethod}[1]{%
  \ifnum#1=1
    GramCol%
  \else\ifnum#1=2
    IMAP%
  \fi\fi
}

\def\paperID{} 
\def\confName{CVPR}
\def\confYear{2026}

\title{\textit{I'm a Map!} Interpretable Motion-Attentive Maps: \\ Spatio-Temporally Localizing Concepts in Video Diffusion Transformers}

\author{Youngjun Jun \quad
Seil Kang \quad
Woojung Han \quad
Seong Jae Hwang\\
Yonsei University\\
{\tt\small  \{youngjun, seil, dnwjddl, seongjae\}@yonsei.ac.kr}
}

\begin{document}

\twocolumn[{%
    \renewcommand\twocolumn[1][]{#1}%
    \maketitle
    
    \centering
    \includegraphics[width=\textwidth]{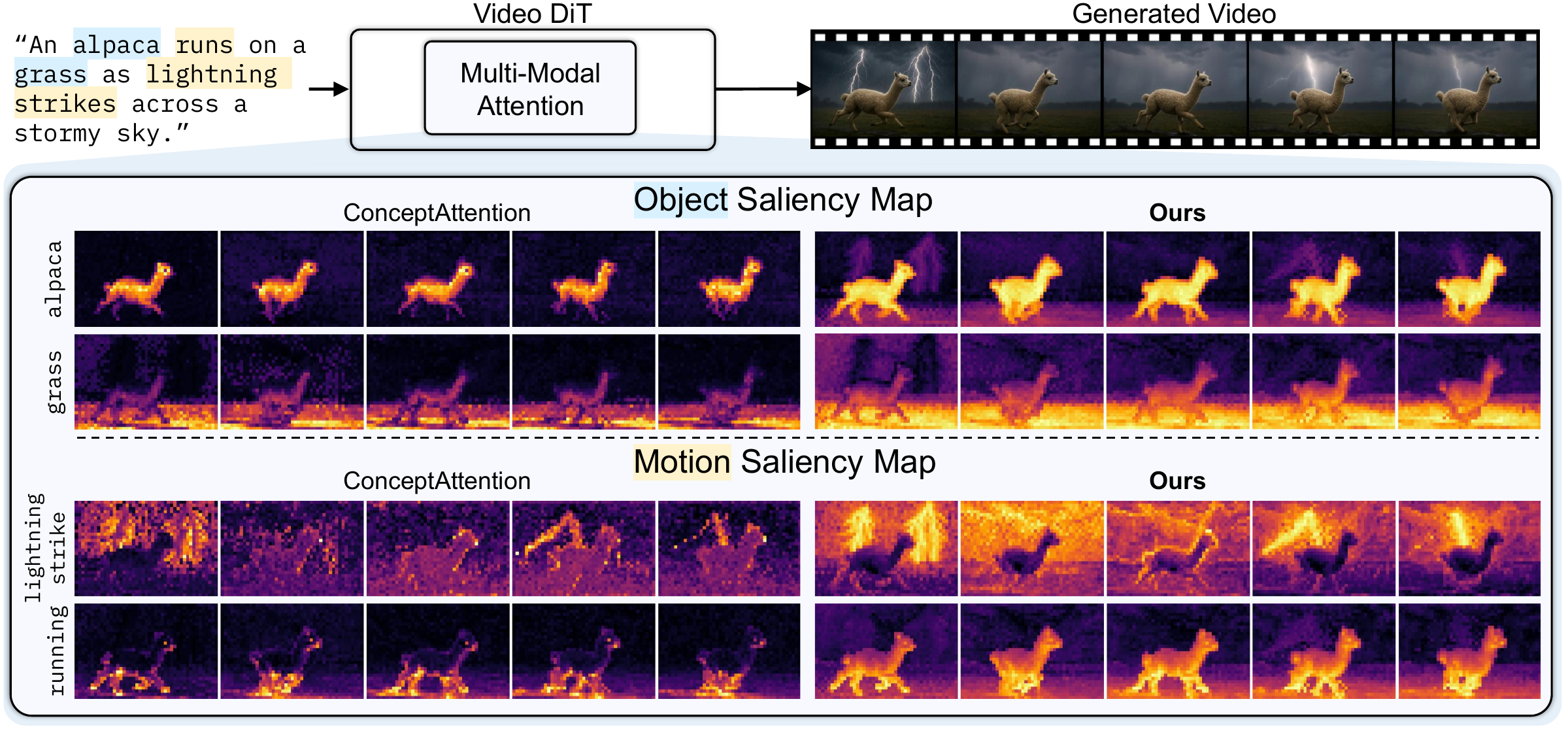}
    
    \captionsetup{hypcap=false}
    \vspace{-6pt}
    \captionof{figure}{
        Interpretable Motion-Attentive Maps (IMAP).
        IMAP is an interpretable map that spatially and temporally localizes \textit{any} motion concept (e.g., lightning strike). Without additional training, IMAP is obtained from the features of video diffusion transformers at motion-specific attention heads.
        Our code is available at \href{https://github.com/youngjun-jun/IMAP}{https://github.com/youngjun-jun/IMAP}.
    }
    \vspace{10pt}
    \label{fig:fig1}
  }]

\begin{abstract}
Video Diffusion Transformers (DiTs) have been synthesizing high-quality video with high fidelity from given text descriptions involving motion. However, understanding how Video DiTs convert motion words into video remains insufficient. Furthermore, while prior studies on interpretable saliency maps primarily target objects, motion-related behavior in Video DiTs remains largely unexplored. In this paper, we investigate concrete motion features that specify when and which object moves for a given motion concept. First, to spatially localize, we introduce \ourmethod{1}, which adaptively produces per-frame saliency maps for any text concept, including both motion and non-motion. Second, we propose a motion-feature selection algorithm to obtain an Interpretable Motion-Attentive Map (IMAP) that localizes motion spatially and temporally. Our method discovers concept saliency maps without the need for any gradient calculation or parameter update. Experimentally, our method shows outstanding localization capability on the motion localization task and zero-shot video semantic segmentation, providing interpretable and clearer saliency maps for both motion and non-motion concepts.

\end{abstract}
\vspace{-10pt}    
\section{Introduction}
\label{sec:intro}

Diffusion transformers (DiTs)~\cite{DiT} have become the foundation of vision generation due to their exceptional visual fidelity and scalability \cite{SD3,SANA-video,CogVideoX,CosmosPredict}. Specifically, given text descriptions, multi-modal diffusion transformers (MM-DiTs) synthesize high-quality visual contents with high fidelity \cite{Wan,Flux,Hunyuanvideo}. Despite such advancements, understanding the generative mechanisms inside the black-box MM-DiTs is still insufficient. A deeper understanding of how MM-DiTs work is essential for substantiating their predictions, identifying failures, and ultimately improving their alignment with human intent and expectations \cite{parktext,basu2024mechanistic,tumanyan2023plug}.

Most studies on interpretability of existing MM-DiTs have been limited to the image domain \cite{ConceptAttention, toker2024diffusion, basu2024mechanistic}. Unlike a single image, a video is a sequence of images with temporal movement, conveying motion information. Accordingly, to emulate such temporal consistency, MM-DiTs for video (i.e., Video DiTs) should be able to comprehend videos. Consequently, analyzing Video DiTs facilitates the extraction of interpretable temporal features, thereby providing insight into the black-box video generation process.

Regarding temporal features in Video DiTs, existing studies have mainly focused on the inter-frame dynamics of visual tokens \cite{DiTFlow,DiffTrack}. DiTFlow~\cite{DiTFlow} attempts to find motion flows using cross-frame attention, and DiffTrack~\cite{DiffTrack} identifies temporal correspondences via query-key matching between frames. 
However, how Video DiTs digest a given text description and generate corresponding temporal movement remains largely unexplored. This raises the following question: \textit{Do Video DiTs truly understand and create motion?} 
In this regard, visualizing where Video DiTs focus in the video given text tokens (i.e., saliency map) provides insight into model behavior.
While ConceptAttention~\cite{ConceptAttention} provides an interpretable saliency map for a concept text token, it only offers spatial separation and does not provide interpretability for motion or temporal movement.

In this paper, we explore an interpretable saliency map for understanding how Video DiTs process motion.
This motion saliency map is able to reveal when and which object moves given a motion word (e.g., Fig.~\ref{fig:fig1}).
To construct such a map, we focus on temporal features of Video DiTs that localize motion.
Specifically, we analyze Video DiTs at the levels of timesteps, DiT blocks, and attention heads with consideration of their specialized roles~\cite{Seg4Diff,ConceptAttention,DiffTrack}.
From this analysis, we find that attention query-key matching exhibits strong spatial localizability and that the separation of frame embeddings is related to motion localizability.

Based on these observations, we first introduce \textit{\ourmethod{1}}, which produces saliency maps by leveraging attention query-key matching. \ourmethod{1} computes a similarity map using the Gram matrix of video tokens and a query-key-matched text-surrogate token, adaptively selecting text-surrogate tokens for each frame.
By using a surrogate token rather than a text token directly, 
we mitigate unintended artifacts caused by cross-modal feature similarity calculations.
Then, to temporally localize motion, 
we identify motion-related attention heads through a frame-wise separation score and present an \textit{Interpretable Motion-Attentive Map (IMAP)}, which is a spatiotemporal saliency map for the given motion concept.
Our IMAP aggregates features across motion-related attention heads in a lightweight, training-free, and gradient-free manner. Furthermore, IMAP is applicable regardless of the attention type of DiT (joint attention \cite{CogVideoX,Hunyuanvideo} or cross attention~\cite{Wan,SANA-video}). 
Moreover, as shown in Fig.~\ref{fig:fig1}, IMAP works with arbitrary prompts and can be applied to any existing video to obtain distinct saliency maps in a zero-shot manner via re-noising and denoising.

Experimentally, IMAP produces clear motion saliency maps that indicate when and which object moves, outperforming existing baselines. 
Additionally, our method is easily applicable to zero-shot video semantic segmentation.

\ourparagraph{Contributions.} 
The main contributions of this work are
\begin{itemize}[noitemsep, nolistsep, leftmargin=*]
    \item We propose \ourmethod{1} that clearly visualizes any text concept feature in Video DiTs, using a text-surrogate token and Gram matrix of video tokens.
    \item We present Interpretable Motion-Attentive Maps (IMAP) that spatiotemporally localize motion concepts by leveraging motion attention heads.
    \item As an interpretable saliency map, IMAP improves our understanding of how Video DiTs process videos and is applicable to perception tasks such as segmentation.
\end{itemize}

\section{Related Work}
\label{sec:rel}

\subsection{Diffusion Model Interpretability}
\label{ssec:interp}

Diffusion Models (DMs)~\cite{NCSN,YangSongSDE,EDM} generate data through iterative denoising, standing out in various fields~\cite{SD3,Flux,CogVideoX,Wan,SDS,MaskedDLM}.
To understand the rationale behind the remarkable success of DMs, studies have attempted to interpret them analytically \cite{ManifoldDM,CFGpp,Autoguidance} and visually \cite{WhatTheDAAM,ConceptAttention,DiffusionExplorer}.

\ourparagraph{Analytic Aspects.}
Because pure (unconditional) DMs~\cite{NCSN,YangSongSDE,EDM} are grounded in rigorous mathematics, much of the interpretability literature has focused on how DMs understand and process conditioning. 
Chung et al.~\cite{ManifoldDM,CFGpp} propose a geometric interpretation on the manifold for the diffusion process.
AutoGuidance~\cite{Autoguidance} and Guidance Interval~\cite{GuidanceInterval} analyze how classifier-free guidance~\cite{CFG} biases diffusion sampling toward high-density regions.
PCG~\cite{PCG} identifies the stochastic differences of CFG in SDE- and ODE-based formulations and interprets CFG as a predictor-corrector scheme.
From a different perspective, some studies suggests that DMs memorize and duplicate their training samples rather than create novel content \cite{UnderstandCopy,DigitalForgery,DetExpMit}.

\ourparagraph{Visual Aspects.}
There have also been attempts to visually analyze how the designed denoising network leverages the conditioning information during generation.
DAAM~\cite{WhatTheDAAM} leverages cross-attention maps and obtains attribution maps for nouns, verbs, adjectives, and adverbs. ConceptAttention~\cite{ConceptAttention} finds a saliency map using the concept feature and validates outstanding performance in zero-shot image segmentation. TokenRank~\cite{TokenRank} presents a viewpoint that interprets the attention map as the transition matrix of a discrete-time Markov chain (DTMC). Additionally, TokenRank acquires image segmentation maps exploiting the attention transition, weighting DTMC states by the second-largest eigenvalue $\lambda_2$ of the transition matrix. 
Regarding the temporal feature, there are attempts to use cross-frame attention. DiTFlow~\cite{DiTFlow} finds a displacement matrix using cross-frame attention and achieves motion transfer from the reference video to the target video. DiffTrack~\cite{DiffTrack} discovers temporal correspondence in cross-frame query-key matching of a specific layer.
Those studies focus on spatial segmentation or temporal continuity, and we aim to find a motion map that localizes motion spatially and temporally.

\subsection{Interpretability in Attention Mechanism}
\label{ssec:mha}

As the core mechanism of transformers, multi-head attention is an attention mechanism composed of multiple heads~\cite{Transformer}. Because they undergo mutually independent matrix multiplication processes, the heads come to possess independent information flows. Therefore, there were findings that the roles of multiple heads are different in studies of large language models~\cite{BERTLook,olsson2022context,zheng2024attention,ferrando2024primer}, large vision-language models (LVLMs)~\cite{YourLVLM,UnveilingVisualPercep}, vision transformers (ViTs)~\cite{DINO,DINOv2,DINOv3} and diffusion transformers~\cite{SparseVideoGen, HeadHunter}. Several studies find that a specific head delivers key information and seek to evaluate the importance of heads~\cite{HeadKnockout,InterpretAttnHeads}. In the DINO series~\cite{DINO,DINOv2,DINOv3}, it has been found that some heads of ViT naturally form attention maps along object boundaries, and Kang et al.~\cite{YourLVLM} discovered that a few specific heads of LVLM provide features necessary for visual grounding. Especially, in Video DiTs, to reduce the attention computational cost between visual tokens, several methods find that spatial heads and temporal heads and sparsify them in different ways~\cite{SparseVideoGen,Sparse-vDiT,TrainFreeSparseAttn}.
Going beyond prior work in Video DiTs, we focus on identifying the internal mechanisms that attribute motion to generated videos. We further interpret these mechanisms through a human-interpretable saliency-map analysis.

\section{Preliminary}
\label{sec:prelim}

\subsection{Multi-Modal Diffusion Transformers}
\label{ssec:prelim-mmdit}

Video DiTs such as CogVideoX~\cite{CogVideoX} and HunyuanVideo \cite{Hunyuanvideo} are diffusion models~\cite{NCSN,YangSongSDE} that use DDIM scheduler~\cite{DDIM} or Rectified flow scheduler~\cite{RectifiedFlow}.
\footnote{The frameworks of diffusion models and flow models are equivalent \cite{Diff-Flow}.We collectively refer to them as diffusion models for convenience.} 
These Video DiTs are an architecture in which multi-modal diffusion transformer (MM-DiT) blocks~\cite{SD3} are stacked layer by layer.
In MM-DiT blocks, the text and visual token embeddings are processed in independent streams (i.e., double stream), and multi-modal attention (MM-Attn) is the key module that integrates these two embeddings. Concretely, MM-Attn jointly performs multi-head self-attention over visual and text embeddings, thereby enabling the two heterogeneous modalities to interplay in the latent space.
By letting $\mQ$, $\mK$, $\mV$ be the query/key/value projection matrices and the projected joint (visual/text) embeddings be $\vq_{xp} = \left[\vq_x, \vq_p\right]$, $\vk_{xp} = \left[\vk_x, \vk_p\right]$, $\vv_{xp} = \left[\vv_x, \vv_p\right]$, the self-attention operation is carried out as follows:
\begin{equation}
    \vh_{xp} = \left[\vh_{x},\text{ }\vh_{p}\right] = \mathrm{softmax}\left({\vq_{xp}\vk_{xp}\tran}/{\sqrt{d}}\right) \vv_{xp},
\end{equation}
where $\vh_{x}$, $\vh_{p}$ are the visual/text token embeddings and $d$ is the head dimension.
In particular, since Video DiTs concatenate patchified frames, $\vh_{x}$ takes the form of a series of per-frame embeddings. Therefore, in MM-Attn, the visual token embeddings of all frames respectively compute the attention score with the text token embeddings:
\begin{align}
    \vq_{x}&=\left[\vq_{f_{0}}, \vq_{f_{1}}, \vq_{f_{2}}, \ldots, \vq_{f_{F}}\right], \nonumber\\ 
    \mA_{f_i}&=\mathrm{softmax}\left({\vq_{f_{i}}\vk_{xp}\tran}/{\sqrt{d}}\right)\vv_{f_{i}p},
\end{align}
where $f_i$ refers $i$-th frame. In this paper, we call $\vq_{f_i}\vk_p\tran$ as Query-Key Matching, similar to prior work \cite{DiffTrack}.

\begin{figure*}[t]
  \centering
   \includegraphics[width=\textwidth]{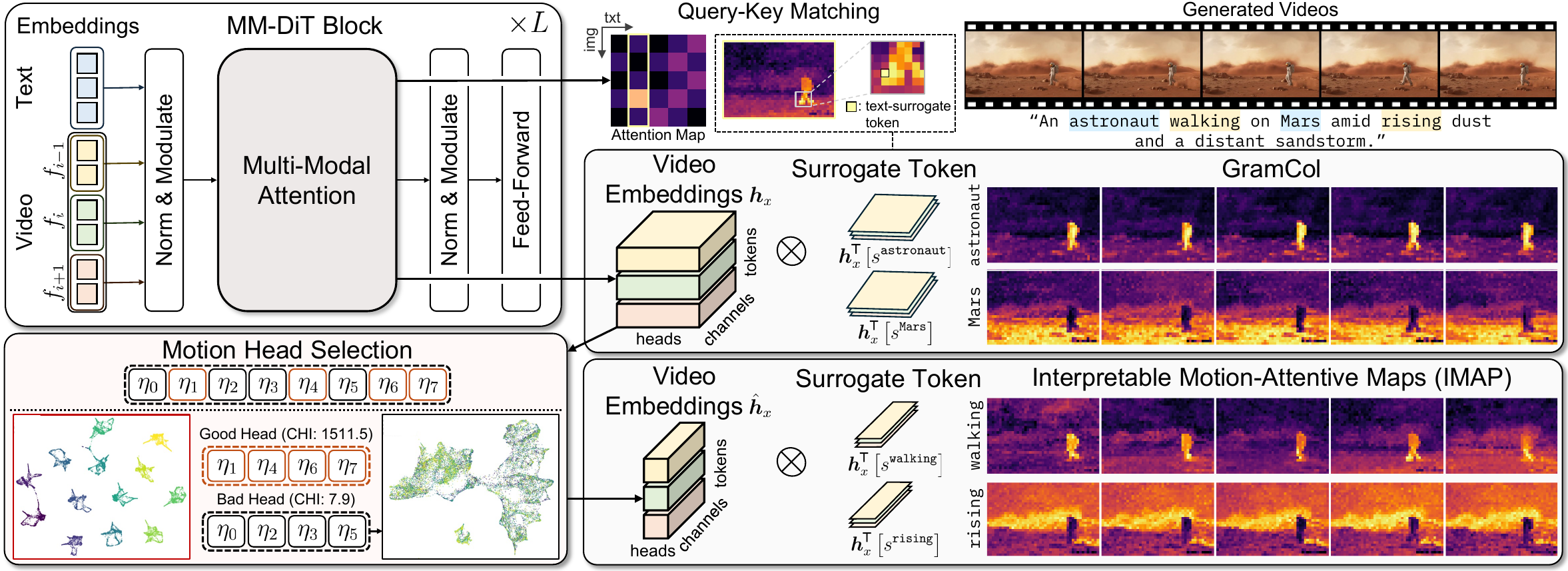}
   \vspace{-18pt}
   \caption{Spatiotemporal localization pipeline. This pipeline obtains a video saliency map for \textit{any} concept using Video DiTs. Given a concept, we first obtain a text-surrogate token via Query–Key Matching, and then compute the \ourmethod{1} to derive its spatial saliency map. For motion concepts, we additionally identify motion heads before computing \ourmethod{1}, thereby improving temporal localization. 
   }
   \vspace{-10pt}
   \label{fig:pipe}
\end{figure*}

\subsection{Concept Token Stream}
\label{ssec:prelim-conceptattn}

Pre-trained Video DiTs take a text prompt and a noisy latent as inputs. Accordingly, the internal MM-Attn handles only prompt tokens for the text embeddings. To obtain the concept-token embeddings provided by the user, prior work~\cite{ConceptAttention} updates the concept token along the frozen text stream.
Following this pre-trained text stream, at each MM-DiT block $l$, we update the concept-token embeddings $\vc^{l}$ using the pre-trained adaptive layer normalization ($\mathrm{AdaLN}$) and the pre-trained feed-forward ($\mathrm{FF}$) layer as follows:
\begin{align}
    \vh_c &= \mathrm{softmax}\left({\vq_{c}\vk_{xc}\tran}/{\sqrt{d}}\right) \vv_{xc},\\
    \vc^{l+1} &\xleftarrow{} \vc^l + \alpha_1\mathrm{AdaLN}(\vh_c), \nonumber\\
    \vc^{l+1} &\xleftarrow{} \vc^{l+1} + \alpha_2\mathrm{FF}(\mathrm{AdaLN}(\vc^{l+1})). \nonumber
    \label{eq:concept-stream}
\end{align}


\section{Methods}
\label{sec:method}

\textit{Motion} is the temporal movement of an object, and a \textit{motion concept} refers to verbal words that represent this movement.
Our goal is to spatially and temporally localize the visual region for a given motion concept within Video DiTs.
From an information-flow perspective~\cite{elhage2021mathematical}, we view the Query-Key (QK) circuit as determining where tokens attend, thereby providing a form of spatial localization. Also, the Output-Value (OV) circuit mediates what information is transferred, shaping the temporal localization of features within the attention mechanism.

\ourparagraph{Spatiotemporal Localization Pipeline.}
The overall pipeline for obtaining saliency maps is shown in Fig.~\ref{fig:pipe}. This pipeline operates on the MM-DiT blocks that are effective for localization (Sec.~\ref{ssec:time-layer}). Our localization generates a saliency map for every concept token using \textit{\ourmethod{1}} (Sec.~\ref{ssec:spatial}) by default, while motion-related concepts are further processed with an additional \textit{motion head selection} step (Sec.~\ref{ssec:temporal}) before applying \ourmethod{1}.
\ourmethod{1} uses a single visual token as a text surrogate to obtain a spatial localization map, whereas the \textit{interpretable motion-attentive map (IMAP)} (Sec.~\ref{ssec:temporal}) obtained via motion head selection enables spatiotemporal localization of motion.

\subsection{Subject of Analysis}
\label{ssec:time-layer}

Video DiTs consist of $L$ layers (i.e., MM-DiT blocks) at each of the $T$ diffusion timesteps, and each layer performs a specialized role~\cite{Seg4Diff,ConceptAttention,DiffTrack}. Accordingly, aggregating all features across the $L\times T$ layers makes it difficult to capture distinct features, and is both redundant and computationally inefficient.
To lens the concepts, including motion, we observe the concept features sequentially at the diffusion timestep and DiT layer levels.

\ourparagraph{Timesteps.}
First, we tighten the analysis scope by discarding early timesteps that are close to pure noise. The first reason is that, due to the highly noisy latent, features extracted at early timesteps are semantically intractable, making it difficult to obtain clear boundaries. Second, we empirically observe that many memorization-related features (e.g., watermarks) emerge at these early steps, making the features blurry.
This aligns with prior findings on memorization in diffusion models, showing that updating only early timestep latents mitigates memorization \cite{DetExpMit,AMG,AdjustInitNoise}. 
Further discussion of these phenomena is provided in the supplement.

\begin{figure}[t]
    \centering
    \vspace{-3pt}
   \includegraphics[width=\linewidth]{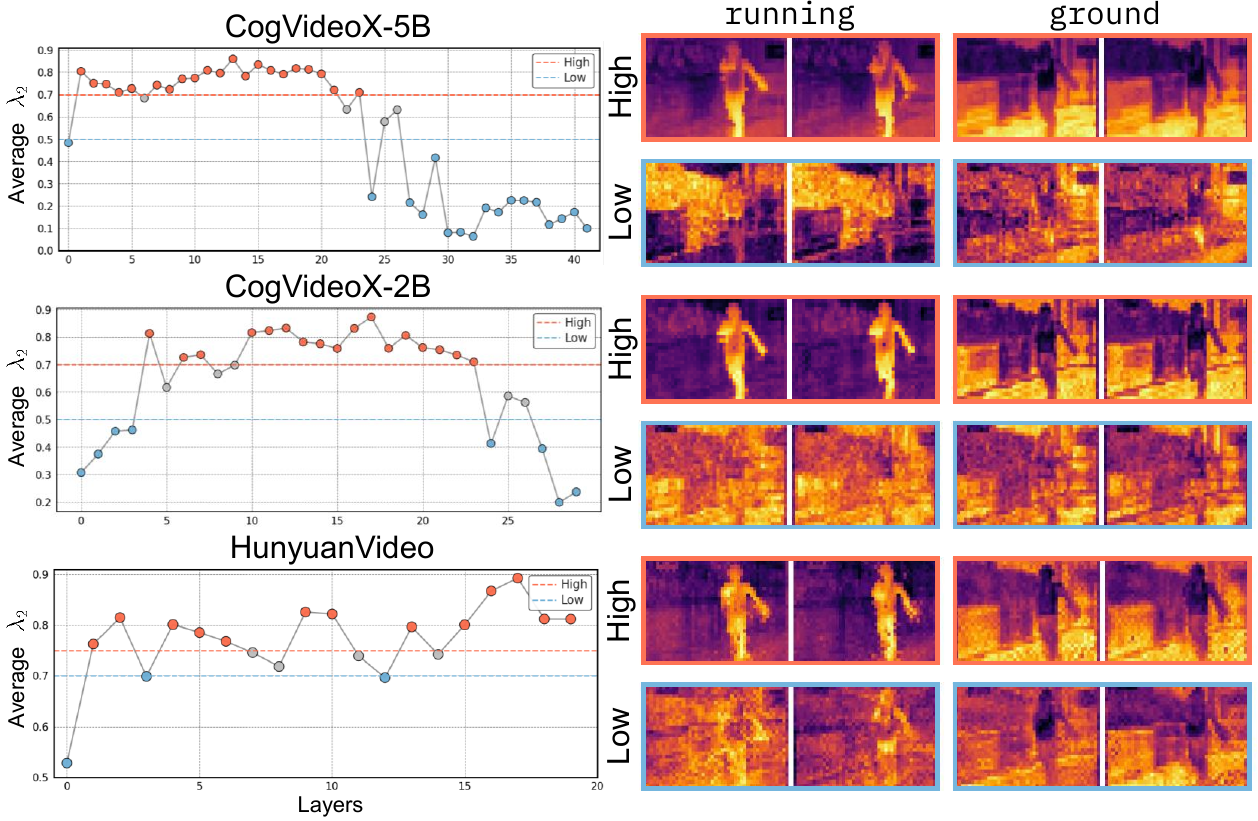}

    \vspace{-5pt}
   \caption{Layer-wise average of $\lambda_2$ and the corresponding feature maps. As the average $\lambda_2$ increases, the extracted features become sharper and more interpretable.
   }
   \vspace{-14pt}
   \label{fig:layer}
\end{figure}

\ourparagraph{Layers.}
Now, we select the layer that is informative for the spatio-temporal semantic feature. Following prior work using the framework of a discrete-time Markov chain (DTMC)~\cite{DTMC,TokenRank}, we interpret each token as a state, and the attention weight matrix as the state transition probability matrix. 
At this point, TokenRank~\cite{TokenRank} hypothesizes that the second-largest eigenvalue $\lambda_2$ of the attention weight matrix indicates the importance of transition matrix.
Following these assumptions, we select the layer with a larger $\lambda_2$ of the attention matrix.
Concretely, we use the average $\lambda_2$ over attention heads of MM-Attn as a layer-select criterion.
Fig.~\ref{fig:layer} shows the per-layer average $\lambda_2$ in CogVideoX (2B/5B)~\cite{CogVideoX} and HunyuanVideo~\cite{Hunyuanvideo}.
Across these models, layers with a higher average $\lambda_2$ exhibit richer semantic features, whereas layers with a lower average $\lambda_2$ tend to produce somewhat blurry features. We provide more details in the supplement.

\subsection{\ourmethod{1}: Spatial Localization}
\label{ssec:spatial}
Spatial localization is the process of making concept features visually emergent. We perform spatial localization by taking the product of two embeddings representing the concept and the video.
We exploit Query-Key matching to obtain a \textit{text-surrogate token}  for the concept and \textit{\ourmethod{1}} converts it into a saliency map form.

\begin{figure}[t]
  \centering
   \includegraphics[width=\linewidth]{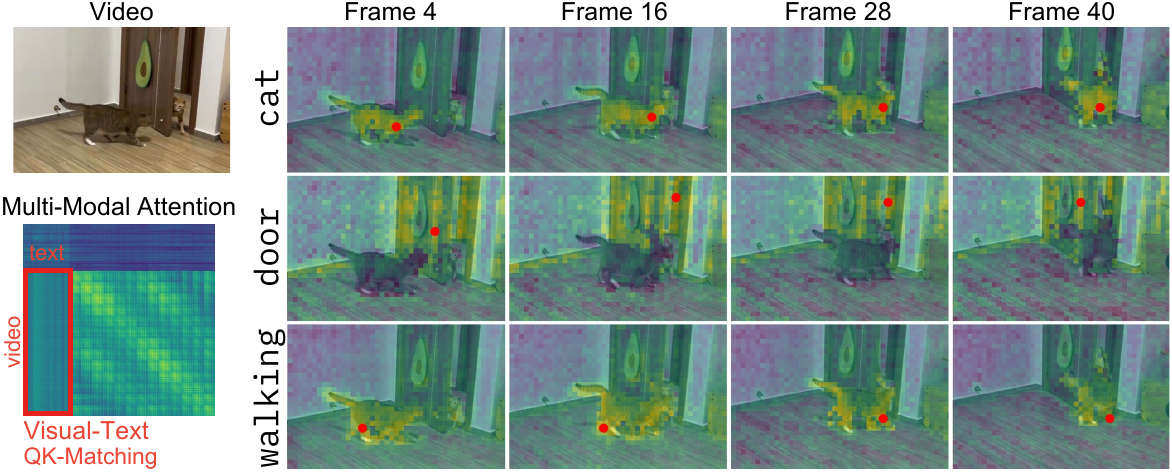}

    \vspace{-5pt}
   \caption{QK-Matching visualizations per text token. While QK-Matching yields somewhat unclear spatial localization, its peak (red dot) still accurately pinpoints the target concept.}
   \vspace{-12pt}
   \label{fig:qk-match}
\end{figure}

\ourparagraph{Query-Key Matching (QK-Matching).}
The attention score formed by QK-Matching ($\vq_{f_i}\vk_c\tran$) has been extensively studied~\cite{DiffTrack,OVAM}. In particular, it has been widely used to localize the image token that matches the text token and has been used as an interpretable visualization tool \cite{WhatTheDAAM,ConceptAttention}. In Fig.~\ref{fig:qk-match}, we examine whether QK-Matching can localize concept words (e.g., cat, walking).
Although QK-Matching produces activation maps by measuring concept–token similarity, it does not guarantee reliable localization across all cases.
However, if we look at the visual token with the largest attention scores (red dot in Fig.~\ref{fig:qk-match}), it becomes a tool that pinpoints the highly relevant visual location.
These conclusions are consistent with prior work~\cite{xu2015show, chen2020uniter, chenspatial}.
Furthermore, when evaluating point-location accuracy on a video segmentation dataset~\cite{VSPW} for foreground/background objects, this peak location identifies the object with an accuracy of 0.9544. See the supplement for more details.

\ourparagraph{\ourmethod{1}.}
Since QK-Matching accurately locates the highly relevant patch for a given text token, we employ QK-Matching to select the visual token that most reliably represents the given text token.
We refer to this selected visual token as \textit{text-surrogate token} with index $s_{f_i}^c$ for frame $i$:
\begin{equation}
    s_{f_i}^c = \underset{p\in\{1,2,3,\dots,P\}}{\mathrm{argmax}}(\mathrm{row}_p(\vq_{f_i})\vk_{c}\tran), \text{ for concept } c, 
    \label{eq:text-surrogate-token}
\end{equation}
where $P$ is the number of visual tokens.
 We compute \ourmethod{1} as the $s_{f_i}$-th column of the Gram matrix $\mG$ of the visual token embeddings $\vh_{x,1}\tran, \vh_{x,2}\tran, \dots, \vh_{x,P}\tran\in\mathbb{R}^{d\times 1}$: 
\begin{align}
    \mG = \vh_x\vh_x\tran \in \mathbb{R}^{P\times P}, \\
    \mathrm{\ourmethod{1}}(\mG, c) = \mG[s_{f_i}^c] \in \mathbb{R}^{P}.
    \label{eq:autoprojection}
\end{align}
In Eq.~(\ref{eq:autoprojection}), the $(i,j)$-th entry of the Gram matrix $\mG$ corresponds to the similarity between the $i$-th and $j$-th visual token embeddings. Thus, it can be interpreted as a similarity map with respect to the text-surrogate visual token obtained via QK-matching. In this case, the resulting \ourmethod{1} is the product of two embeddings lying on the same manifold. In particular, when these embeddings represent the same semantics, their token embeddings become similar, yielding a positive highlight over semantically related regions.

Since we automatically select the text surrogate tokens for every attention head and frame, \ourmethod{1} addresses head-wise discrepancies and adaptively adjusts these tokens according to temporal movement. 
Moreover, because it is a similarity-based map of the visual token, the more similar the visual tokens are, the larger the positive value, resulting in positive highlighted regions. This property is important for interpretability as it prevents the emphasis on large negative values.
Consequently, we obtain the final \ourmethod{1} map by averaging \ourmethod{1} over the selected timesteps $t\in\mathcal{T}$, layers $l\in\mathcal{L}$, and heads $\eta\in\mathcal{H}$:
\begin{equation}
    \mathrm{\ourmethod{1}Map}(c)=\frac{1}{\vert\mathcal{T}\vert\vert\mathcal{L}\vert\vert\mathcal{H}\vert}\sum_{t,l,\eta} \mathrm{\ourmethod{1}}(\mG, c).
    \label{eq:auto_map}
\end{equation}

\ourparagraph{Remark.}
ConceptAttention~\cite{ConceptAttention} is a methodology that multiplies the image latent and the concept text embedding:
\begin{equation}
    \mathrm{ConceptAttention}=\underset{c}{\mathrm{softmax}}\left(\vh_x \vh_c\tran \right).
    \label{eq:softmax}
\end{equation}
ConceptAttention also exploits products of embeddings, but it differs from \ourmethod{1} in two important aspects. \textbf{First}, \ourmethod{1} uses a Gram matrix of visual token embeddings, where each entry encodes their pairwise similarity. This construction yields saliency maps with \textit{positive highlightability} across all heads and frames. This means that areas similar to the text-surrogate token will have a large positive value.
In contrast, ConceptAttention leverages embeddings from different modalities, which results in the absence of this interpretable property. In particular, this phenomenon can hinder the acquisition of distinct features due to heterogeneous behaviors across attention heads (see supplements). 
Furthermore, since we leverage visual-token similarities, \ourmethod{1} becomes naturally suitable for handling frame-varying patterns.
\textbf{Second}, \ourmethod{1} does not depend on a softmax over a user-defined concept list. Applying a softmax over concepts like ConceptAttention (Eq.~(\ref{eq:softmax})) makes the saliency map incomplete when a patch represents more than one concept (e.g., leg, walking). In \ourmethod{1}, by contrast, a saliency map can be obtained even when only a single concept is provided, without such competition.

\subsection{Motion Heads: Temporal Localization}
\label{ssec:temporal}

We now investigate temporal localization specialized for motion concepts (e.g., walking). 
Previous works on Video DiTs have argued that MM-Attn heads are categorized into spatial and temporal heads~\cite{SparseVideoGen,Sparse-vDiT,TrainFreeSparseAttn}. Inspired by these studies, we aim to identify motion-related temporal heads responsible for motion localization. To this end, we analyze the token-wise properties of the visual token embeddings.

\textit{Motion} involves temporal movement, which causes differences between video frames. 
Accordingly, 
we assume that a head exhibiting large differences among per-frame visual tokens is rich in temporal motion features.
For each head, the visual token embedding $\vh_x\in\mathbb{R}^{P\times d}$ is formed by concatenating the per-frame visual tokens and flattening them: $P=FHW$ for height $H$ and width $W$ of the diffusion latent.
At this point, we treat each frame as a separate cluster and measure the \textit{separation score} of the visual tokens (e.g., Calinski-Harabasz index (CHI)~\cite{CHI}, Davies-Bouldin index~\cite{DBI}, Fisher ratio~\cite{Fisher-ratio}, etc.).
In other words, we regard $\vh_x$ as a collection of $P$ vectors $\vh_{x,j}\tran \in \mathbb{R}^d$ and, for each frame, treat the $HW$ vectors $\vh_{x,j}\tran$ as datapoints.

\begin{figure}[t]
  \centering
   \includegraphics[width=\linewidth]{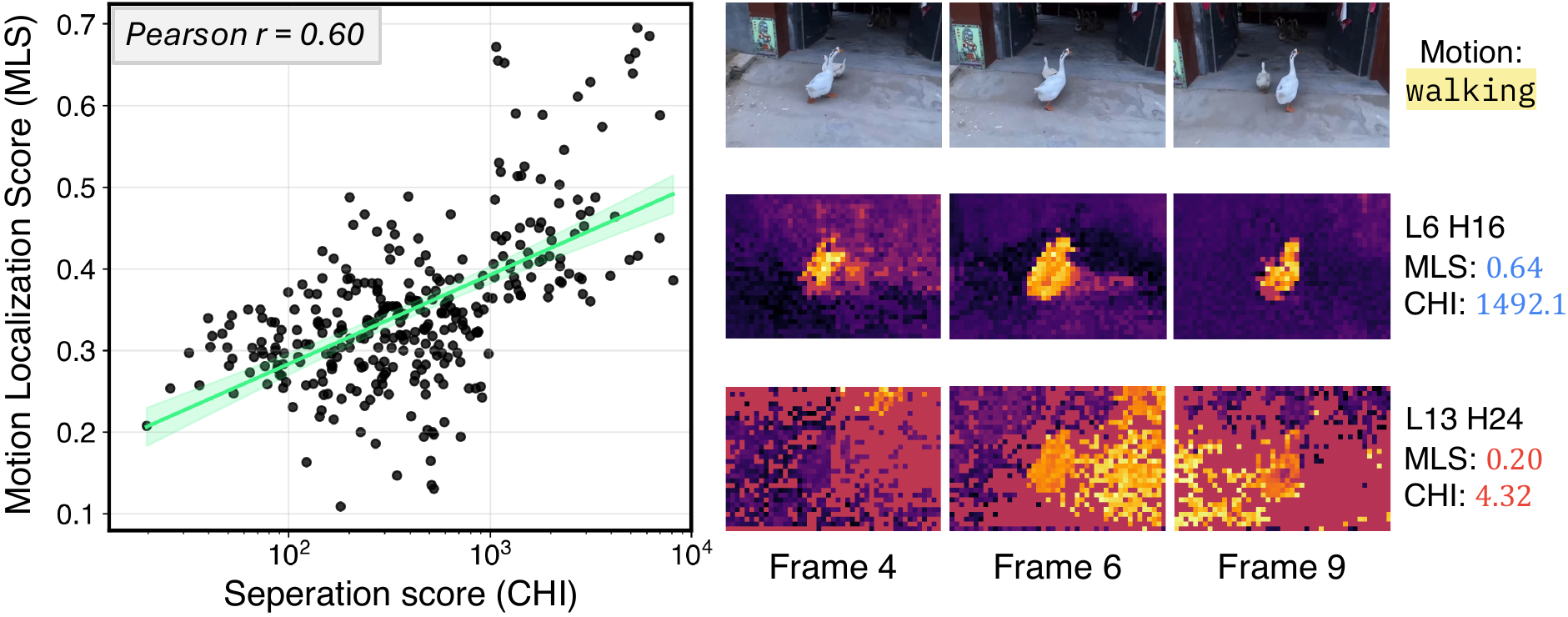}

    \vspace{-5pt}
   \caption{Visualization of Motion Localization Score (MLS) versus the separation score (Calinski-Harabasz index, CHI). MLS measured from \ourmethod{1} extracted across attention heads in layers $\mathcal{L}$ of CogVideoX-2B. Heads with higher CHI scores tend to exhibit higher MLS, with a Pearson correlation coefficient of 0.60.
   }
   \vspace{-9pt}
   \label{fig:head}
\end{figure}

In this case, if the separation score (e.g., CHI) is large, the differences in visual token features between frames are large, implying strong temporal variation. 
To validate our hypothesis that a high separation score corresponds to better motion features, we extract \ourmethod{1} from the heads in layers $\mathcal{L}$, and evaluate them using the Motion Localization Score (MLS) (in Sec.~\ref{ssec:stml}). To this end, we use the random subset of MeViS~\cite{MeViS} validation set and randomly sample $1,500$ \ourmethod{1} from attention heads. 
Fig.~\ref{fig:head} presents the MLS with respect to CHI, the corresponding regression line, and several representative \ourmethod{1} samples. Since a higher separation score tends to correspond to a higher MLS, we conclude that highly separated heads with large temporal discrepancy are effective at localizing motion.
Additionally, as shown in the bottom-left of Fig.~\ref{fig:pipe}, the selected heads exhibit clear inter-frame feature differences when visualized via a dimensionality reduction method \cite{PaCMAP}.

Therefore, we use only heads with high separation scores to obtain the saliency maps for the motion concepts. 
In this situation, a na\"ive strategy would be to (1) compute separation scores for all heads and then return to the first layers to select the desired ones, or to (2) store all \ourmethod{1} maps and later use only those with high separation scores. However, both approaches are computationally inefficient and particularly ill-suited to iterative diffusion models.
To circumvent this issue, for each layer $l$ we select the top-$k$ heads with the highest separation scores and retain only their visual token embeddings $\hat{\vh}_x$. We then compute the \ourmethod{1} map for motion concept $c_m$ according to Eq.~(\ref{eq:autoprojection}, \ref{eq:auto_map}):
\begin{align}
    \hat{\mG}=&\hat{\vh}_x\hat{\vh}_x\tran, \\
    \mathrm{IMAP}(c_m)=\frac{1}{\vert\mathcal{T}\vert\vert\mathcal{L}\vert\vert\hat{\mathcal{H}}\vert}&\sum_{t,l,\hat{\eta}} \mathrm{\ourmethod{1}}(\hat{\mG}, c_m).
    \label{eq:imam}
\end{align}
We refer this spatiotemporal motion localization map as the \textit{Interpretable Motion-Attentive Map (IMAP)}.

\ourparagraph{Remark.}
Since \ourmethod{1} uses only one column of the Gram matrix, the computation overhead is negligible, and the CHI requires very lightweight operations (see our supplement). Furthermore, our pipeline is fully automatic, including the selection of the text-surrogate token and motion head.

\section{Experiments}
\label{sec:exp}

We experimentally evaluate and analyze how effective the \textit{interpretable motion-attentive map (IMAP)} is for motion localization. We use the CogVideoX (2B/5B)~\cite{CogVideoX} and HunyuanVideo~\cite{Hunyuanvideo} text-to-video diffusion transformers, and for a given injected concept, we leverage the concept-token stream (Sec.~\ref{ssec:prelim-conceptattn}) of the double-stream MM-DiT blocks to obtain concept saliency maps in a training-free manner. 
Cross-attention–based models~\cite{Wan,SANA-video} that cannot leverage the concept-token stream are discussed in the supplement.

\ourparagraph{Benchmark.}
For consistent evaluation, we use the real video dataset. We add noise to these videos according to the diffusion schedule and extract saliency maps along the denoising trajectory. For fair evaluation, we explicitly exclude analysis samples from our protocol. Specifically, we chunk videos from the MeViS~\cite{MeViS} train set into 49-frame clips and use the Qwen3-VL~\cite{Qwen3-VL} to caption each clip and drop motionless videos. Our final benchmark consists of 504 videos covering 150 distinct motion types.

\begin{figure*}[t]
   \includegraphics[width=\linewidth]{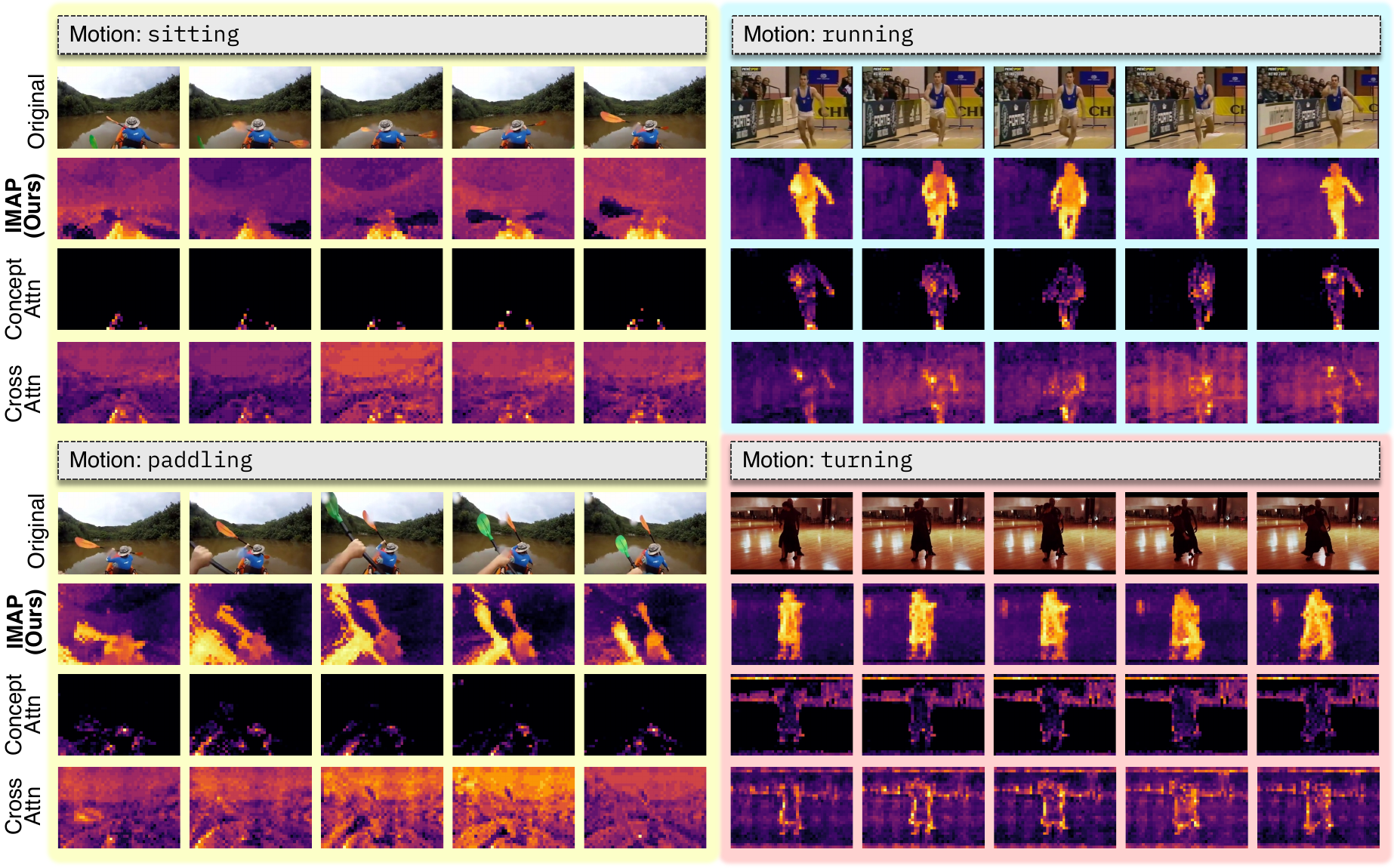}
   \vspace{-15pt}
   \caption{Qualitative comparisons of motion localization results using CogVideoX-5B on the MeViS dataset
   }
   \vspace{-10pt}
   \label{fig:motion}
\end{figure*}

\ourparagraph{Metric: Motion Localization Score (MLS).}
Evaluating motion localization maps is a nontrivial task. To manage this, we employ the state-of-the-art frontier large language model (LLM) OpenAI o3-pro~\cite{o3-pro}, similar to prior work~\cite{WhatTheDAAM,PhysicsIQ}. This LLM assesses the saliency maps along five metrics: Spatial Localization (SL), Temporal Localization (TL), Prompt Relevance (PR), Specificity/Sparsity (SS), and Objectness/Boundary quality (OBJ).
We adopt a rubric-based judgement protocol; the detailed criteria, scoring prompts, and examples are provided in the supplement.

\ourparagraph{Implementation Details.}
Following our analysis in Fig.~\ref{fig:layer}, we use layers with average $\lambda_2$ greater than $0.7$ for CogVideoX and $0.75$ for HunyuanVideo, respectively.
And, we use only the double-stream MM-DiT blocks for HunyuanVideo. For motion head selection, we adopt the Calinski-Harabasz index (CHI)~\cite{CHI} as the separation score and use only the top-5 heads in all experiments. For more details, please refer to the supplementary material.

\subsection{Spatiotemporal Motion Localization Results}
\label{ssec:stml}

\noindent\textbf{Baselines.}
We adopt existing methods with localization capability as baselines, covering diverse backbones: the ViT-based Video CLIP (ViCLIP)~\cite{ViCLIP}, the U-Net-based video generation model VideoCrafter2~\cite{VideoCrafter2} with DAAM~\cite{WhatTheDAAM}, and DiT-based video generation models (CogVideoX 2B/5B~\cite{CogVideoX}, HunyuanVideo~\cite{Hunyuanvideo}) with cross-attention aggregation and ConceptAttention~\cite{ConceptAttention}.

\ourparagraph{Results.}
As shown in Table~\ref{tab:motion}, IMAP achieves the best performance across all five metrics among non-diffusion feature maps, diffusion U-Net feature maps, and diffusion transformer feature maps. Moreover, for all three DiT-based models (CogVideoX-2B/5B, and HunyuanVideo), IMAP consistently outperforms alternative methods. Fig.~\ref{fig:motion} provides a qualitative comparison of the DiT-based approaches. IMAP exhibits superior spatiotemporal localization across all frames and a wide variety of motion categories (e.g., paddling, turning) compared to the baselines. See the supplements for more qualitative comparisons.

\subsection{Ablation Study}
\label{ssec:ablation}

We evaluate the effect of (i) replacing cross-attention aggregation or ConceptAttention with our proposed \ourmethod{1} (Exp.~\#1), and (ii) selecting layers and motion heads (Exp.~\#2, 3). We additionally examine the performance when (iii) applying a softmax operation (Eq.~(\ref{eq:softmax})) over concepts, as in ConceptAttention (Exp.~\#5).
Furthermore, we empirically validate the faithfulness of our motion head selection via random head experiments in the supplement.

\begin{table}[t]
\setlength{\tabcolsep}{6pt}
\renewcommand{\arraystretch}{1.35}
\caption{Motion localization results on including five metrics.
}
\vspace{-5pt}
\resizebox{\linewidth}{!}{%
    \centering
    \begin{tabular}{lccccccc}
    \specialrule{1pt}{0pt}{3pt}
    Method & Backbone & SL & TL & PR & SS & OBJ & Avg. \\
    \specialrule{1.2pt}{2pt}{2pt}
    ViCLIP & ViT-H & 0.33 & 0.17 & 0.35 & 0.29 & 0.28 & 0.28 \\
    DAAM & VideoCrafter2 & 0.36 & 0.17 & 0.38 & 0.32 & 0.35 & 0.32 \\
    \specialrule{0.5pt}{0pt}{0pt}
    Cross Attention & HunyuanVideo & 0.39 & 0.25 & 0.41 & 0.36 & 0.34 & 0.35 \\
    ConceptAttention & HunyuanVideo & 0.42 & 0.26 & 0.44 & 0.35 & 0.34 & 0.36 \\
    \rowcolor{cyan!10} IMAP & HunyuanVideo & 0.60 & 0.41 & 0.62 & 0.50 & 0.62 & 0.55 \\
    \specialrule{0.5pt}{0pt}{0pt}
    Cross Attention & CogVideoX-2B & 0.29 & 0.56 & 0.35 & 0.23 & 0.28 & 0.34 \\
    ConceptAttention & CogVideoX-2B & 0.42 & 0.26 & 0.44 & 0.35 & 0.34 & 0.36 \\
    \rowcolor{cyan!10} IMAP & CogVideoX-2B & 0.49 & \textbf{0.62} & 0.56 & 0.48 & 0.55 & 0.54 \\
    \specialrule{0.5pt}{0pt}{0pt}
    Cross Attention & CogVideoX-5B & 0.41 & 0.27 & 0.43 & 0.34 & 0.33 & 0.36 \\
    ConceptAttention & CogVideoX-5B & 0.50 & 0.32 & 0.51 & 0.47 & 0.47 & 0.45 \\
    \rowcolor{cyan!10} IMAP & CogVideoX-5B & \textbf{0.68} & 0.48 & \textbf{0.69} & \textbf{0.61} & \textbf{0.64} & \textbf{0.62} \\
    \specialrule{1pt}{0pt}{0pt}
    \end{tabular}
}
\vspace{-9pt}
\label{tab:motion}
\end{table}

\begin{figure}[t]
  \centering
    \vspace{-9pt}
   \includegraphics[width=\linewidth]{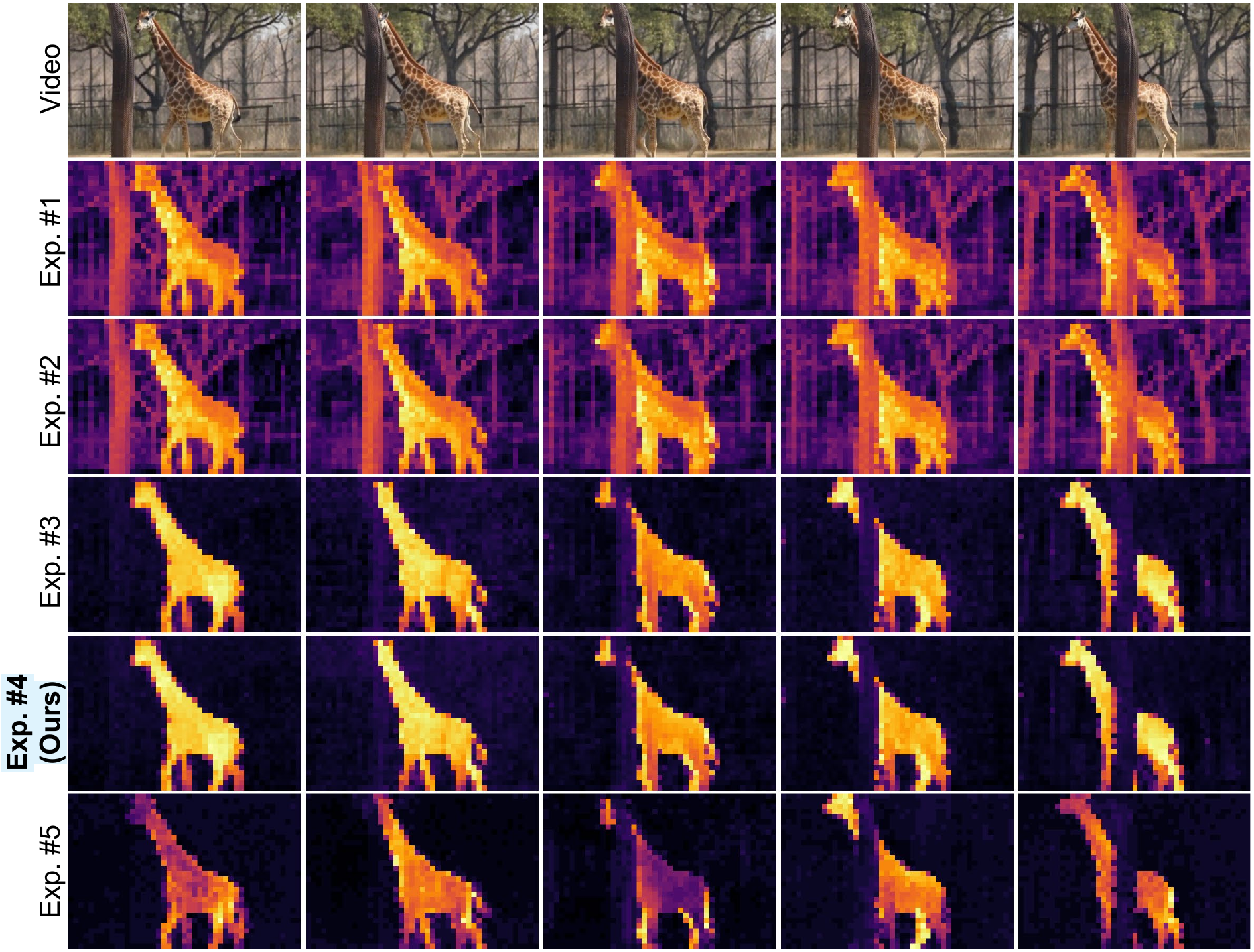}
    \vspace{-17pt}
   \caption{Qualitative ablation results exhibiting progressive improvement in motion localizability (concept: walking).
   }
    \vspace{-10pt}
   \label{fig:ablation}
\end{figure}

As shown in Table~\ref{tab:ablation}, \ourmethod{1} without any layer or head selection still strictly outperforms all baselines without softmax postprocessing. Furthermore, both layer selection (Sec.~\ref{ssec:time-layer}) and motion head selection (Sec.~\ref{ssec:temporal}) yield clear performance gains. Interestingly, applying a softmax over concepts to IMAP leads to additional improvements on several metrics.
Fig.~\ref{fig:ablation} shows related qualitative results. The motion-head selection used in Exp.~\#3-4 accurately highlights the agent of motion. Introducing a softmax operation in Exp.~\#5 yields more distinct features but somewhat reduces frame-wise consistency.
For more experimental analysis, please refer to the supplementary material.

\subsection{Zero-Shot Video Semantic Segmentation}
\label{ssec:vss}

\vspace{-3pt}

To thoroughly evaluate the spatial localization capability of \ourmethod{1}, we conduct a zero-shot video semantic segmentation (VSS) task. In zero-shot VSS, segmentation is performed at inference time without any labels, and the predicted segments are evaluated by comparing them with the ground-truth (GT) labels of the first frame during evaluation. However, methods such as cross-attention, ConceptAttention, and IMAP operate on Video DiTs for given concept words, and thus cannot be applied under an identical, fully label-free inference setting. Therefore, analogous to using the first-frame GT labels, our default setting assumes that Video DiTs are given a concept list at inference time and make predictions conditioned on these concepts. In this setting, there is no first-frame GT matching, and we use only one diffusion timestep to obtain saliency maps.

\begin{table}[t]
\setlength{\tabcolsep}{5pt}
\renewcommand{\arraystretch}{1.3}
\caption{Ablation results showing the impact of our components on the performance over five metrics.
}
\vspace{-7pt}
\resizebox{\linewidth}{!}{%
    \centering
    \begin{tabular}{clcccccc}
    \specialrule{1pt}{0pt}{3pt}
    Exp. \# & Method & SL & TL & PR & SS & OBJ & Avg. \\
    \specialrule{1.2pt}{1pt}{2pt}
    & Cross Attention & 0.41 & 0.27 & 0.43 & 0.34 & 0.33 & 0.36 \\
    & ConceptAttention w/o softmax & 0.42 & 0.24 & 0.44 & 0.37 & 0.38 & 0.37 \\
    & ConceptAttention w/ softmax & 0.50 & 0.32 & 0.51 & 0.47 & 0.47 & 0.45 \\
    \specialrule{0.8pt}{2pt}{2pt}
    1 & \ourmethod{1} (all layers) & 0.45 & 0.30 & 0.47 & 0.41 & 0.42 & 0.41 \\
    2 & + layer selection & 0.47 & 0.34 & 0.48 & 0.48 & 0.50 & 0.46 \\
    3 & + motion head selection & 0.53 & 0.34 & 0.55 & 0.46 & 0.48 & 0.47 \\
    \rowcolor{cyan!10} 4 & + both (IMAP) & 0.68 & 0.48 & 0.69 & 0.61 & 0.64 & 0.62 \\
    5 & IMAP w/ softmax & 0.61 & 0.55 & 0.62 & 0.58 & 0.66 & 0.60 \\
    \specialrule{1pt}{0pt}{0pt}
    \end{tabular}
}
\label{tab:ablation}
\end{table}

\ourparagraph{Evaluation Dataset \& Baselines.}
We use the VSPW~\cite{VSPW} validation set to compare \ourmethod{1} maps against baselines. This VSPW dataset contains 343 videos and 124 object classes. As baselines, we consider specialized segmentation models (UniVS~\cite{UniVS}, DVIS++~\cite{DVISpp}, CLIPpy~\cite{CLIPpy}, EmerDiff~\cite{EmerDiff}, VidSegDiff~\cite{VidSegDiff}) and interpretable saliency maps (Cross Attention, ConceptAttention~\cite{CrossAttnHeadPosition}).

\begin{figure}[t]
  \centering
    \vspace{-9pt}
   \includegraphics[width=\linewidth]{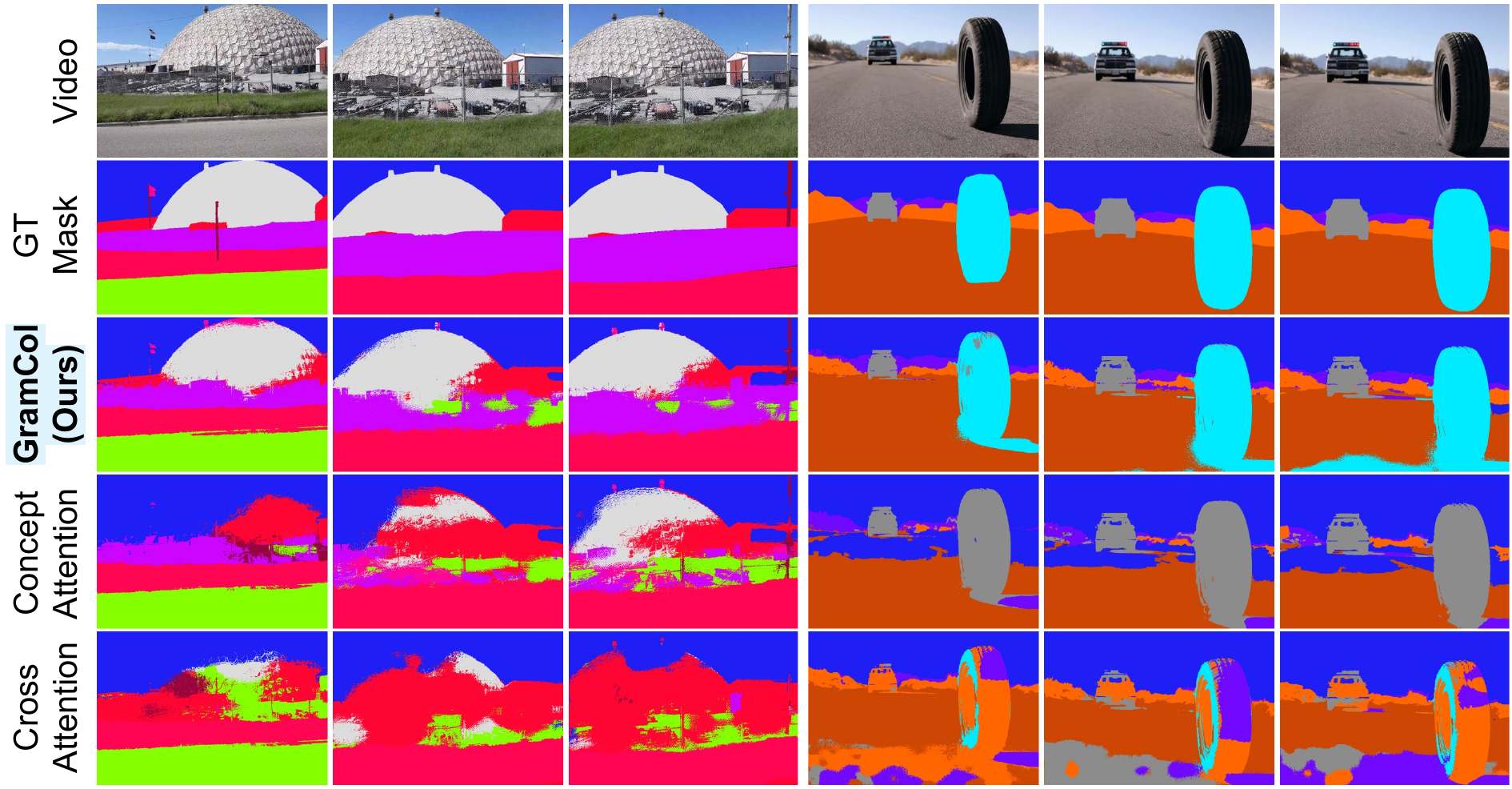}

    \vspace{-5pt}
   \caption{Qualitative results of zero-shot video semantic segmentation using interpretable map methods for Video DiTs. All these saliency maps are upsampled by AnyUp~\cite{AnyUp}.
   }
    \vspace{-15pt}
   \label{fig:vss}
\end{figure}

\ourparagraph{Results.}
As shown in Table~\ref{tab:vss}, \ourmethod{1} attains the highest mIoU among the saliency maps of Video DiTs, achieving high accuracy with moderate video consistency (VC). 
As shown in Fig.~\ref{fig:vss}, the saliency maps exhibit segmentation ability even without any clustering methods such as a KNN classifier. Among the Video DiT saliency methods, \ourmethod{1} produces the most accurate segmentation predictions. While its performance still lags behind specialized segmentation models, these results highlight the potential of leveraging Video DiT feature maps for future research.

\begin{table}[t]
\setlength{\tabcolsep}{4pt}
\renewcommand{\arraystretch}{1.2}
\caption{Quantitative results of zero-shot video semantic segmentation on the VSPW dataset.
}
\vspace{-7pt}
\resizebox{\linewidth}{!}{%
    \centering
    \begin{tabular}{llccccc}
    \specialrule{1pt}{0pt}{3pt}
    & Method & & Backbone & mIoU & mVC$_8$ & mVC$_{16}$ \\
    \specialrule{1.2pt}{1pt}{2pt}
    \multirow{2}{*}{\shortstack{Supervised\\Segmentation}} & UniVS & \textcolor{gray}{\scriptsize{CVPR'24}} & Swin-T & 59.8 & 92.3 & -\\
    & DVIS++ & \textcolor{gray}{\scriptsize{ICCV'23}} & VIT-L & 63.8 & 95.7 & 95.1 \\
    \specialrule{0.5pt}{2pt}{2pt}
    \multirow{5}{*}{\shortstack{Zero-Shot\\Segmentation}} & CLIPpy & \textcolor{gray}{\scriptsize{ICCV'23}} & T5 + DINO & 17.7 & 72.4 & 68.4 \\
    & EmerDiff & \textcolor{gray}{\scriptsize{ICLR'24}} & SD 2.1 & 43.4 & 68.9 & 64.3 \\
    & VidSegDiff & \textcolor{gray}{\scriptsize{arXiv'24}} & SD 2.1 & 60.6 & 90.7 & 89.6 \\
    & EmerDiff & \textcolor{gray}{\scriptsize{ICLR'24}} & SVD & 39.7 & 82.1 & 78.5 \\
    & VidSegDiff & \textcolor{gray}{\scriptsize{arXiv'24}} &  SVD & 53.2 & 89.3 & 88.0 \\
    \specialrule{1.5pt}{2pt}{2pt}
    \multirow{3}{*}{\shortstack{Interpretable\\Saliency Map}} & Cross Attention & & CogVideoX-5B & 16.8 & 71.5 & 59.1 \\
    & ConceptAtttention & \textcolor{gray}{\scriptsize{ICML'25}} & CogVideoX-5B & 25.0 & 80.4 & 72.1 \\
    & \cellcolor{cyan!10}\ourmethod{1} (Ours)
    & \cellcolor{cyan!10}
    & \cellcolor{cyan!10}CogVideoX-5B
    & \cellcolor{cyan!10}28.9
    & \cellcolor{cyan!10}75.2
    & \cellcolor{cyan!10}66.0 \\
    \specialrule{0.5pt}{0pt}{2pt}
    \multirow{3}{*}{+ AnyUp} & Cross Attention & & CogVideoX-5B & 17.5 & 74.5 & 64.0 \\
    & ConceptAtttention & \textcolor{gray}{\scriptsize{ICML'25}} & CogVideoX-5B & 25.7 & \textbf{82.4} & \textbf{75.5} \\
    & \cellcolor{cyan!10}\ourmethod{1} (Ours)
    & \cellcolor{cyan!10}
    & \cellcolor{cyan!10}CogVideoX-5B
    & \cellcolor{cyan!10}\textbf{30.1}
    & \cellcolor{cyan!10}77.9
    & \cellcolor{cyan!10}70.1 \\
    \specialrule{1pt}{0pt}{0pt}
    \end{tabular}
}
\label{tab:vss}
\end{table}

\vspace{-3pt}
\section{Conclusion}
\label{sec:conc}

\vspace{-5pt}

In this paper, we present IMAP, interpretable motion-attentive maps for Video DiTs that spatially and temporally localize motion concepts.
In particular, we investigate whether Video DiT can spatially localize concepts via text-surrogate tokens and whether any of its thousands of heads drive temporal localization, which we call motion-heads.
We show that our \ourmethod{1} and \ourmethod{2} can support to obtatin interpretable motion maps at a human-level with full automation and without any training or optimization.
Lastly, we hope our findings clarify how Video DiTs generate motion concepts and what further insights these mechanisms provide.
We provide more discussion and failure cases in our supplementary materials.

\clearpage
{
    \small
    \bibliographystyle{ieeenat_fullname}
    \bibliography{bib}
}

\clearpage
\onecolumn

\section*{A. Algorithm \& Computational Overhead}
\label{sec:sup-alg}
\addcontentsline{toc}{section}{A. Algorithm \& Computational Overhead}

Our Interpretable Motion-Attentive Map (IMAP) is derived during the denoising process of Video DiTs. Therefore, IMAP can be obtained either by generating from noise or by re-noising an existing video and then denoising. In both cases, we extract IMAPs at each MM-DiT block and aggregate them. The per-block computation is detailed in Algorithm~\ref{alg:imap}. The additional computing time introduced by IMAP is reported in Table~\ref{tab:supp-layer}.

\begin{center}
\begin{minipage}{0.65\linewidth}
\begin{algorithm}[H]\small
\caption{IMAP in an MM-DiT block}
\centering
\begin{spacing}{1.1}
\begin{algorithmic}
\State \textbf{Input:} block index $l\in \{1,2,\dots,L\}$, pre-trained query/key/value projection matrices $\mQ, \mK, \mV$, the number of heads $\mathcal{H}$, visual token embedding $\vh_x^l$, text (prompt) token embedding $\vh_p^l$, concept token embedding $\vc^l$, selected layer set $\mathcal{L}$, selected timestep set $\mathcal{T}$, total latent space frame, height, width: $F$, $H$, $W$, and total patches $P=FHW$.
\State \textbf{Norm \& Modulate}: $\vh_x^l, \vh_p^l \xleftarrow{} \mathrm{AdaLN}(\vh_x^l, \vh_p^l)$
\State \textbf{Concept Stream}: $\vc^l \xleftarrow{} \mathrm{AdaLN}(\vc^l)$
\State \textbf{Joint Attention}: 
\State\hspace{1em} \textbf{Projection}
\State\hspace{2em} $\vq_{x}=\mQ\vh_x^l$, $\vk_{x}=\mK\vh_x^l$, $\vv_{x}=\mV\vh_x^l$
\State\hspace{2em} $\vq_{p}=\mQ\vh_p^l$, $\vk_{p}=\mK\vh_p^l$, $\vv_{p}=\mV\vh_p^l$
\State\hspace{2em} $\vq_{c}=\mQ\vc^l$, $\vk_{c}=\mK\vc^l$, $\vv_{c}=\mV\vc^l$
\State\hspace{1em} \textbf{MM-Attn}
\State\hspace{2em} $\vh_{x}^l, \vh_{p}^l \xleftarrow{} \mathrm{softmax}\left({\vq_{xp}\vk_{xp}\tran}/\sqrt{d}\right)\vv_{xp}$
\State\hspace{1em} \textcolor{cvprblue}{\textbf{Motion Head Seleciton}}
\State\hspace{2em} $\hat{\vh}_{x}^l \xleftarrow{} \mathrm{reshape}(\vh_{x}^l, \{\mathcal{H}, P, d \xrightarrow{} \mathcal{H}, F, (HW), d\})$
\State\hspace{2em} $\text{SepScores} = \underset{\eta\in\{1,2,\dots,\mathcal{H}\}}{\mathrm{CHI}}(\hat{\vh}_{x,\eta}^l)$
\State\hspace{2em} $\hat{\vh}_{x}^l \xleftarrow{} \mathrm{IndexSelect}(\hat{\vh}_{x}^l, \mathrm{topk\_args}(\mathrm{SepScores}))$
\State\hspace{1em} \textcolor{cvprblue}{\textbf{Query-Key Matching}}
\State\hspace{2em} $\vq_{f_{i\in F}}\vk_{c}\tran=\mathrm{reshape}(\vq_{x}\vk_{c}\tran, \{P \xrightarrow{} F, (HW)\})_{i\in F}$
\State\hspace{1em} \textcolor{cvprblue}{\textbf{Text-Surrogate Token}}
\State\hspace{2em} $s_{f_i}^c = \underset{p\in\{1,2,3,\dots,P\}}{\mathrm{argmax}}(\mathrm{row}_p(\vq_{f_i})\vk_{c}\tran)$
\State\hspace{1em} \textcolor{deepcvprblue}{\textbf{\ourmethod{1}}}
\State\hspace{2em} $\mathrm{\ourmethod{1}}(c) = \vh_{x}^l\vh_{x}^{l\;\mkern-1.5mu\mathsf{T}}[s_{f_i}^c]$
\State\hspace{1em} \textcolor{deepcvprblue}{\textbf{IMAP}}
\State\hspace{2em} $\mathrm{IMAP}(c) = \hat{\vh}_{x}^l\hat{\vh}_{x}^{l\;\mkern-1.5mu\mathsf{T}}[s_{f_i}^c]$
\State \textbf{Norm \& Modulate}
\State\hspace{1em} $\vh_{xp}^{l+1} \xleftarrow{} \vh_{xp}^{l} + \alpha_{1,xp}\mathrm{AdaLN}(\vh_{xp})$
\State\hspace{1em} $\vh_{xp}^{l+1} \xleftarrow{} \vh_{xp}^{l+1} + \alpha_{2,xp}\mathrm{FF}(\mathrm{AdaLN}(\vh_xp^{l+1}))$
\State \textbf{Concept Stream}
\State\hspace{1em} $\vh_c = \mathrm{softmax}\left({\vq_{c}\vk_{xc}\tran}/{\sqrt{d}}\right) \vv_{xc}$
\State\hspace{1em} $\vc^{l+1} \xleftarrow{} \vc^l + \alpha_{1,p}\mathrm{AdaLN}(\vh_c)$
\State\hspace{1em} $\vc^{l+1} \xleftarrow{} \vc^{l+1} + \alpha_{2,p}\mathrm{FF}(\mathrm{AdaLN}(\vc^{l+1}))$
\State \textbf{Output:} $\vh_x^{l+1}$, $\vh_p^{l+1}$, $\vc^{l+1}$, $\mathrm{\ourmethod{1}}$, $\mathrm{IMAP}$.

\end{algorithmic}
\end{spacing}
\label{alg:imap}
\end{algorithm}
\end{minipage}
\end{center}

\begin{table}[h]
\setlength{\tabcolsep}{9pt} 
\renewcommand{\arraystretch}{0.9}
\caption{Stage-wise inference time. All timings are measured on an A100~80GB GPU using CogVideoX-5B~\cite{CogVideoX}, with re-noising followed by a single denoising timestep.
(Video Encode: VAE encoding stage of the real video; Map Save: visualization process of IMAPs.)
}
\centering
\scriptsize
\resizebox{\linewidth}{!}{%
    \begin{tabular}{lccccc}
    \specialrule{0.8pt}{0pt}{3pt}
    \multirow{2}{*}{Stage} & \multirow{2}{*}{Video Encode} & \multicolumn{3}{c}{Diffusion Inference} & \multirow{2}{*}{Map Save} \\
    \cmidrule{3-5}
    & & Diffusion Denoising & Query-Key Matching & Motion Head Selection &  \\
    \specialrule{0.5pt}{2pt}{3pt}    
    Time (s) & 10.79 & 58.67 & 10.35 & 0.083 & 7.17 \\
    \specialrule{0.8pt}{2pt}{0pt}
    \end{tabular}
}
\label{tab:compute}
\end{table}

\clearpage
\section*{B. More Analysis}
\label{sec:sup-analysis}
\addcontentsline{toc}{section}{B. More Analysis}

We provide a detailed analysis of IMAP. In Sec.~\hyperref[ssec:sup-timestep]{B.1}, we analyze the characteristics of IMAP across timesteps. Sec.~\hyperref[ssec:sup-layer]{B.2} qualitatively and quantitatively evaluates IMAP for layers with different average $\lambda_2$ values. In Sec.~\hyperref[ssec:sup-head]{B.3}, we discuss the importance of mitigating negative highlights, and in Sec.~\hyperref[ssec:sup-softmax]{B.4}, we examine the impact of the softmax operation on the resulting saliency maps. Finally, Sec.~\hyperref[ssec:sup-separation]{B.5} investigates different types of separation scores.

\subsection*{B.1. Timesteps}
\label{ssec:sup-timestep}
\addcontentsline{toc}{subsection}{B.1. Timesteps}

Diffusion models~\cite{NCSN,YangSongSDE,EDM} generate data by iteratively denoising from pure noise. As is well known, early timesteps primarily capture coarse-grained structure, while later timesteps refine fine-grained details \cite{DPM-Slim,LatentSpace,MAS}, meaning that the denoising network plays different roles at different timesteps. Motivated by this, we also extract IMAP at each timestep and examine how it varies across the denoising trajectory.

\begin{figure*}[b]
  \centering
   \includegraphics[width=\linewidth]{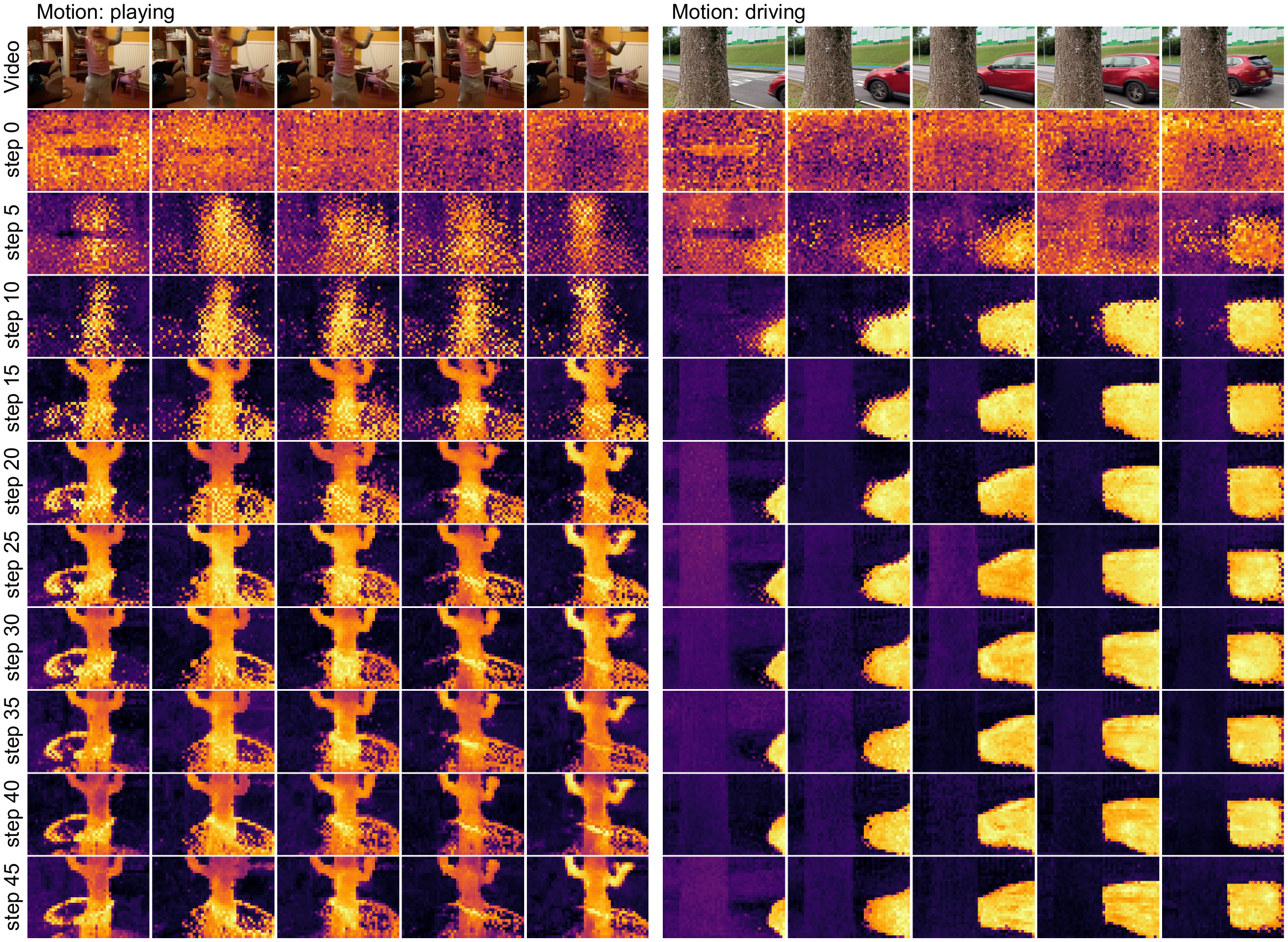}
   \caption{Visualization of IMAP over various timesteps
   }
   \label{fig:supp-timestep}
\end{figure*}

Fig.~\ref{fig:supp-timestep} shows IMAPs extracted from CogVideoX-5B~\cite{CogVideoX} with $50$ inference steps, taken at timesteps $0$, $5$, $10$, $\dots$, $45$. At timesteps $0$ and $5$, the IMAPs exhibit watermark-like features that are not present in the actual video and gradually disappear as denoising proceeds. Up to around timestep $15$, noise induced by the noisy latent reduces the clarity of the maps. Notably, except for the early timesteps, even a single timestep is sufficient to yield discriminative features. Overall, except for these early steps, IMAPs remain largely consistent across timesteps. Therefore, we find that no additional processing is necessary beyond excluding the early timesteps.

In particular, this observation of watermark-like features aligns with insights from prior work on memorization. 
These studies report that even updating the latent only at denoising step~$0$ is sufficient to remove memorization~\cite{DetExpMit,AMG,AdjustInitNoise}.

\subsection*{B.2. Layers}
\label{ssec:sup-layer}
\addcontentsline{toc}{subsection}{B.2. Layers}

Following prior work based on discrete-time Markov chains~\cite{DTMC,TokenRank}, we use the average second-largest eigenvalue $\lambda_2$ of the head-wise MM-Attn attention matrices as a criterion for layer selection.
To validate this criterion, in Table~\ref{tab:supp-layer} we compare using all layers against several subsets: layers with average $\lambda_2$ below 0.5, above 0.5, above 0.7, above 0.75, and above 0.8.
As we increase the average $\lambda_2$ threshold for selecting layers, nearly all metrics consistently improve. Moreover, we observe higher performance despite using fewer layers. While using 0.8 as the threshold yields the best results, we prioritize stable layer selection over marginal gains in performance and therefore adopt thresholds of $0.7$ for CogVideoX (2B/5B)~\cite{CogVideoX} and $0.75$ for HunyuanVideo~\cite{Hunyuanvideo} as our default settings.
Qualitative comparisons of IMAPs for layers with average $\lambda_2$ below 0.5 and above 0.7 are shown in Fig.~\ref{fig:supp-highlow-1} and Fig.~\ref{fig:supp-highlow-2}.

\begin{table}[h]
\setlength{\tabcolsep}{6pt} 
\renewcommand{\arraystretch}{1.3}
\caption{Quantitative analysis of the layer thresholds in CogVideoX-5B
}
\centering
\resizebox{0.7\linewidth}{!}{%
    \begin{tabular}{cccccccc}
    \specialrule{0.8pt}{0pt}{3pt}
    Layers & \# layers & SL & TL & PR & SS & OBJ & Avg. \\
    \specialrule{1pt}{1pt}{2pt}
    All layers & 42 & 0.53 & 0.34 & 0.55 & 0.46 & 0.48 & 0.47 \\
    \specialrule{0.3pt}{2pt}{2pt}
    $<0.50$ & 17 & 0.34 \small{\textcolor{imapred}{(-0.19)}} & 0.22 \small{\textcolor{imapred}{(-0.12)}} & 0.37 \small{\textcolor{imapred}{(-0.18)}} & 0.29 \small{\textcolor{imapred}{(-0.17)}} & 0.28 \small{\textcolor{imapred}{(-0.20)}} & 0.30 \small{\textcolor{imapred}{(-0.17)}} \\
    $>0.50$ & 25 & 0.57 \small{\textcolor{cvprblue}{(+0.04)}} & 0.44 \small{\textcolor{imapblue}{(+0.10)}} & 0.58 \small{\textcolor{cvprblue}{(+0.03)}} & 0.53 \small{\textcolor{cvprblue}{(+0.07)}} & 0.53 \small{\textcolor{cvprblue}{(+0.05)}} & 0.53 \small{\textcolor{cvprblue}{(+0.06)}} \\
    \rowcolor{cyan!10} $>0.70$ & 21 & 0.68 \small{\textcolor{imapblue}{(+0.15)}} & 0.48 \small{\textcolor{imapblue}{(+0.14)}} & 0.69 \small{\textcolor{imapblue}{(+0.14)}} & 0.61 \small{\textcolor{imapblue}{(+0.15)}} & 0.64 \small{\textcolor{imapblue}{(+0.16)}} & 0.62 \small{\textcolor{imapblue}{(+0.15)}} \\
    $>0.75$ & 14 & 0.66 \small{\textcolor{imapblue}{(+0.13)}} & 0.51 \small{\textcolor{imapblue}{(+0.17)}} & 0.68 \small{\textcolor{imapblue}{(+0.13)}} & 0.61 \small{\textcolor{imapblue}{(+0.15)}} & 0.64 \small{\textcolor{imapblue}{(+0.16)}} & 0.62 \small{\textcolor{imapblue}{(+0.15)}} \\
    $>0.80$ & 7 & 0.69 \small{\textcolor{imapblue}{(+0.16)}} & 0.55 \small{\textcolor{imapblue}{(+0.21)}} & 0.70 \small{\textcolor{imapblue}{(+0.15)}} & 0.65 \small{\textcolor{imapblue}{(+0.19)}} & 0.65 \small{\textcolor{imapblue}{(+0.17)}} & 0.65 \small{\textcolor{imapblue}{(+0.18)}} \\
    \specialrule{0.8pt}{0pt}{0pt}
    \end{tabular}
}
\label{tab:supp-layer}
\end{table}

\begin{figure*}[h]
  \centering
   \includegraphics[width=0.7\linewidth]{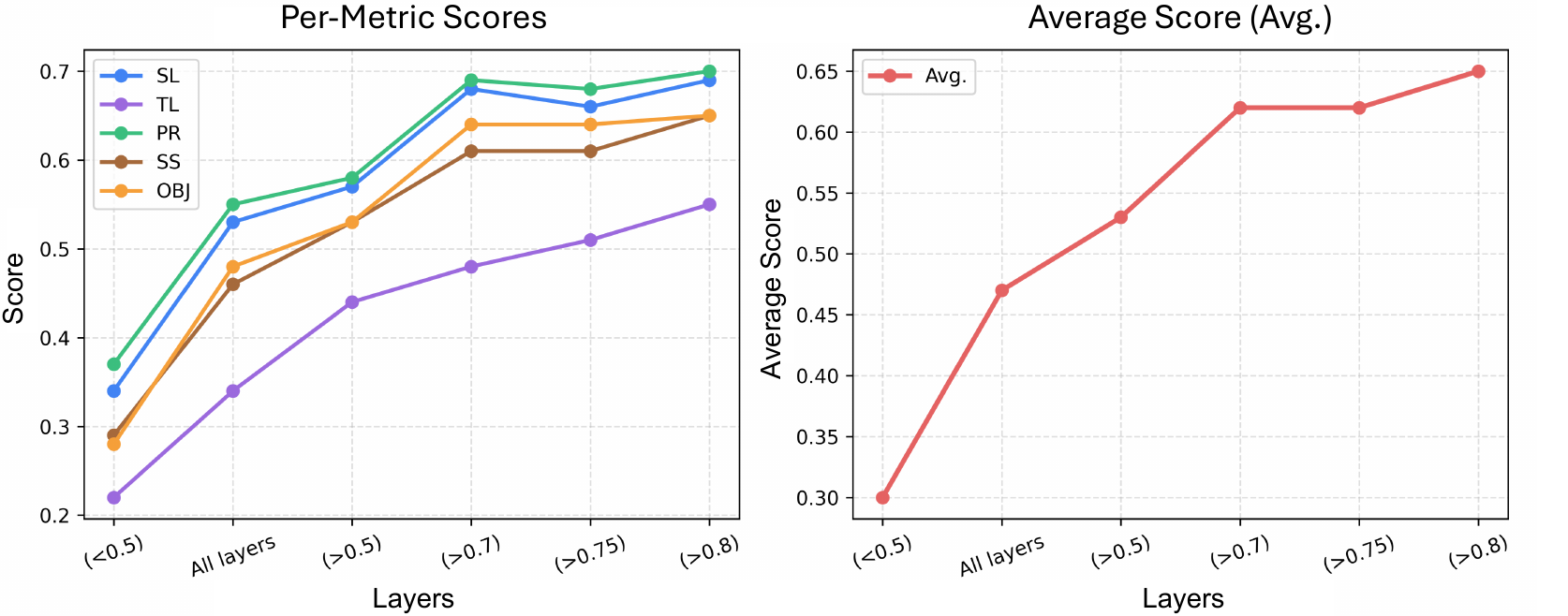}
   \caption{Quantitative analysis of the layer thresholds in CogVideoX-5B
   }
   \label{fig:supp-tabfig}
\end{figure*}

\subsection*{B.3. Negative Heads}
\label{ssec:sup-head}
\addcontentsline{toc}{subsection}{B.3. Negative Heads}

Our \ourmethod{1} selects the column of the Gram matrix of visual token embeddings $\vh_x$ corresponding to the text-surrogate token index. Since each entry of the Gram matrix is an inner product, i.e., a similarity measure, pairs of similar token embeddings yield positive values, which increase as their similarity grows.
Therefore, IMAPs extracted under the \ourmethod{1} operation exhibit \textit{positive highlightability}, meaning that highlighted regions take positive values.

However, as shown in Fig.~\ref{fig:supp-negative-heads}, ConceptAttention~\cite{ConceptAttention}, which measures similarity between concept-token embeddings $\vh_c$ and visual-token embeddings $\vh_x$, can produce negatively highlighted regions in certain attention heads (e.g., L15H9, L15H24, L15H29; L: layer, H: head) because the two modalities differ. Although these negatively highlighted heads do not substantially affect the overall saliency map, they are a factor that should be considered when aiming for sharper and more interpretable saliency maps.

\subsection*{B.4. Softmax Operation over Concepts}
\label{ssec:sup-softmax}
\addcontentsline{toc}{subsection}{B.4. Softmax Operation over Concepts}

\begin{figure}[t]
\centering
\begin{minipage}[t!]{0.58\linewidth}
\centering
\setlength{\tabcolsep}{6pt}
\renewcommand{\arraystretch}{1.1}
\captionof{table}{Quantitative analysis of softmax operation
}
\centering
\resizebox{\linewidth}{!}{
    \begin{tabular}{clccccccc}
    \specialrule{0.9pt}{0pt}{3.5pt}
    Backbone & Method & Softmax & SL & TL & PR & SS & OBJ & Avg. \\
    \specialrule{1.1pt}{1.5pt}{2pt}
    \multirow{6}{*}{\rotatebox{90}{HunyuanVideo}} & \multirow{2}{*}{Cross Attention} & \xmark & 0.39 & 0.25 & 0.41 & 0.36 & 0.34 & 0.35 \\
    & & \checkmark & 0.42 & 0.26 & 0.44 & 0.36 & 0.34 & 0.36 \\
    \cmidrule{2-9}
    & \multirow{2}{*}{ConceptAttention} & \xmark & 0.42 & 0.24 & 0.44 & 0.35 & 0.35 & 0.36 \\
    & & \checkmark & 0.42 & 0.26 & 0.44 & 0.35 & 0.34 & 0.36 \\
    \cmidrule{2-9}
    & \multirow{2}{*}{IMAP} & \cellcolor{cyan!10}\xmark & \cellcolor{cyan!10}0.60 & \cellcolor{cyan!10}0.41 & \cellcolor{cyan!10}0.62 & \cellcolor{cyan!10}0.50 & \cellcolor{cyan!10}0.62 & \cellcolor{cyan!10}0.55 \\
    & & \checkmark & 0.60 & 0.47 & 0.63 & 0.53 & 0.52 & 0.55 \\
    \specialrule{0.9pt}{0pt}{2pt}
    \multirow{6}{*}{\rotatebox{90}{CogVideoX-2B}} & \multirow{2}{*}{Cross Attention} & \xmark & 0.29 & 0.56 & 0.36 & 0.27 & 0.31 & 0.36 \\
    & & \checkmark & 0.42 & 0.26 & 0.45 & 0.35 & 0.33 & 0.36 \\
    \cmidrule{2-9}
    & \multirow{2}{*}{ConceptAttention} & \xmark & 0.26 & 0.56 & 0.35 & 0.23 & 0.28 & 0.34 \\
    & & \checkmark & 0.42 & 0.26 & 0.44 & 0.35 & 0.34 & 0.36 \\
    \cmidrule{2-9}
    & \multirow{2}{*}{IMAP} & \cellcolor{cyan!10}\xmark & \cellcolor{cyan!10}0.49 & \cellcolor{cyan!10}0.62 & \cellcolor{cyan!10}0.56 & \cellcolor{cyan!10}0.48 & \cellcolor{cyan!10}0.55 & \cellcolor{cyan!10}0.54 \\
    & & \checkmark & 0.58 & 0.43 & 0.61 & 0.54 & 0.54 & 0.54 \\
    \specialrule{0.9pt}{0pt}{2pt}
    \multirow{6}{*}{\rotatebox{90}{CogVideoX-5B}} & \multirow{2}{*}{Cross Attention} & \xmark & 0.41 & 0.27 & 0.43 & 0.34 & 0.33 & 0.36 \\
    & & \checkmark & 0.46 & 0.32 & 0.48 & 0.40 & 0.39 & 0.41 \\
    \cmidrule{2-9}
    & \multirow{2}{*}{ConceptAttention} & \xmark & 0.42 & 0.24 & 0.44 & 0.37 & 0.38 & 0.37 \\
    & & \checkmark & 0.50 & 0.32 & 0.51 & 0.47 & 0.47 & 0.45 \\
    \cmidrule{2-9}
    & \multirow{2}{*}{IMAP} & \cellcolor{cyan!10}\xmark & \cellcolor{cyan!10}0.68 & \cellcolor{cyan!10}0.48 & \cellcolor{cyan!10}0.69 & \cellcolor{cyan!10}0.61 & \cellcolor{cyan!10}0.64 & \cellcolor{cyan!10}0.62 \\
    & & \checkmark & 0.61 & 0.55 & 0.62 & 0.58 & 0.66 & 0.60 \\
    \specialrule{0.9pt}{0pt}{0pt}
    \end{tabular}
}
\label{tab:supp-softmax}
\end{minipage}
\hfill
\begin{minipage}[t!]{0.38\linewidth}
\centering
\includegraphics[width=\linewidth]{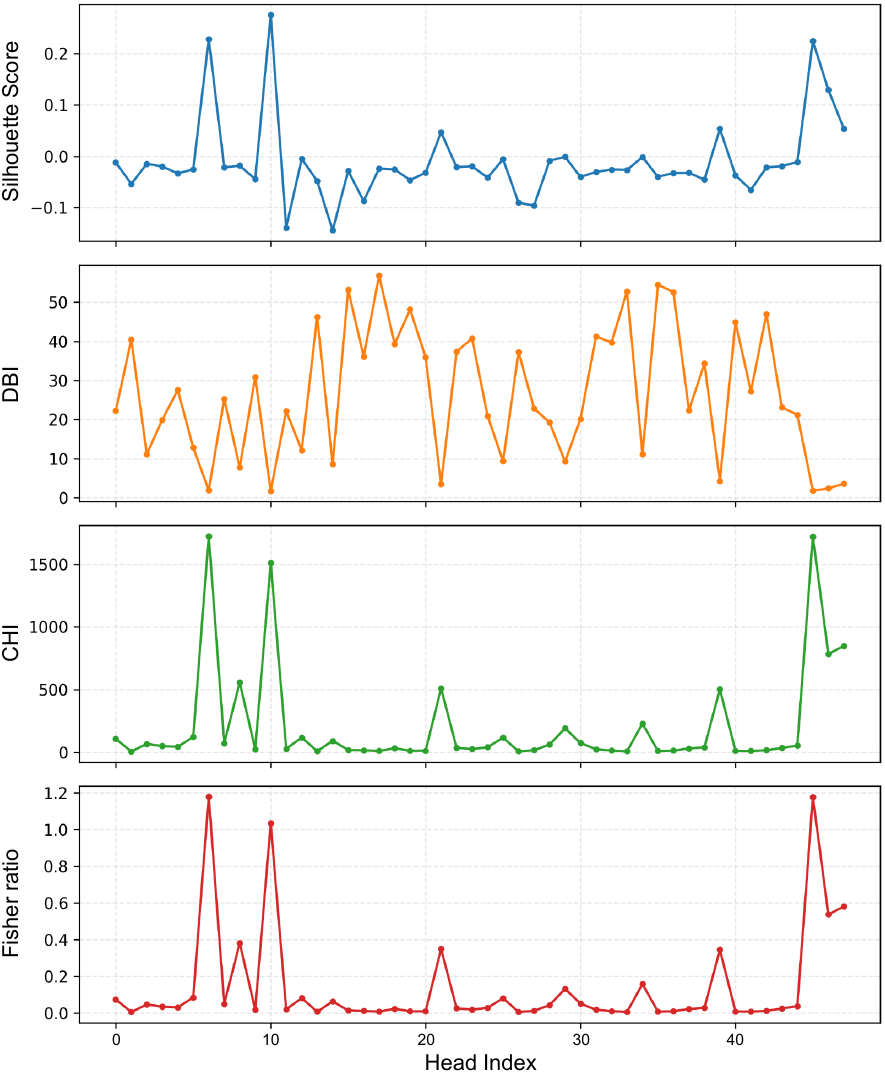}
\caption{Comparison of different separation scores
}
\label{fig:supp-sepscore}
\end{minipage}

\end{figure}

\begin{figure*}[h]
  \centering
   \includegraphics[width=0.9\linewidth]{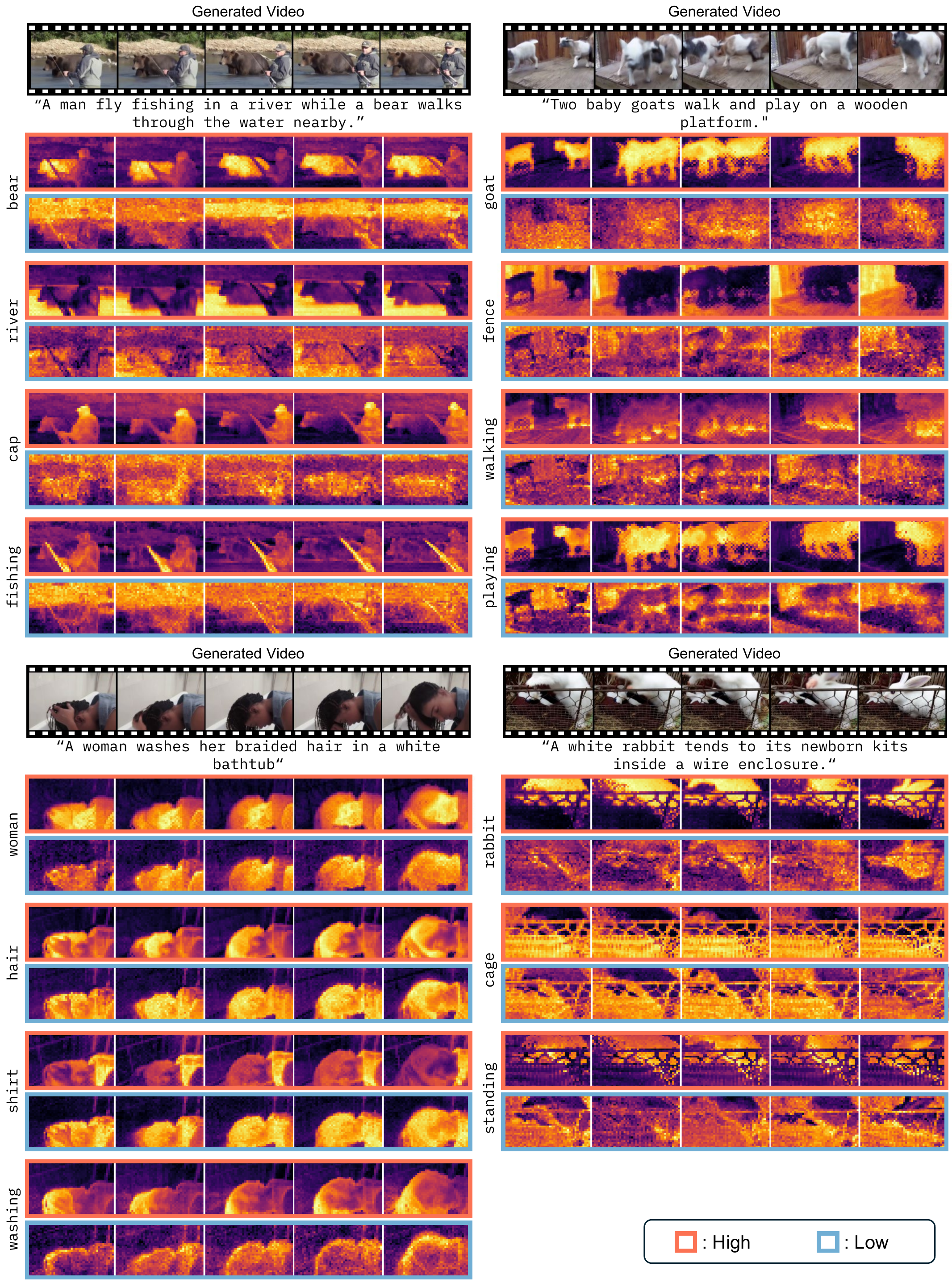}
   \caption{Comparison of IMAP under high ($>0.70$) vs.\ low ($<0.50$) average $\lambda_2$ layers in CogVidoX-5B
   }
   \label{fig:supp-highlow-1}
\end{figure*}

\begin{figure*}[h]
  \centering
   \includegraphics[width=\linewidth]{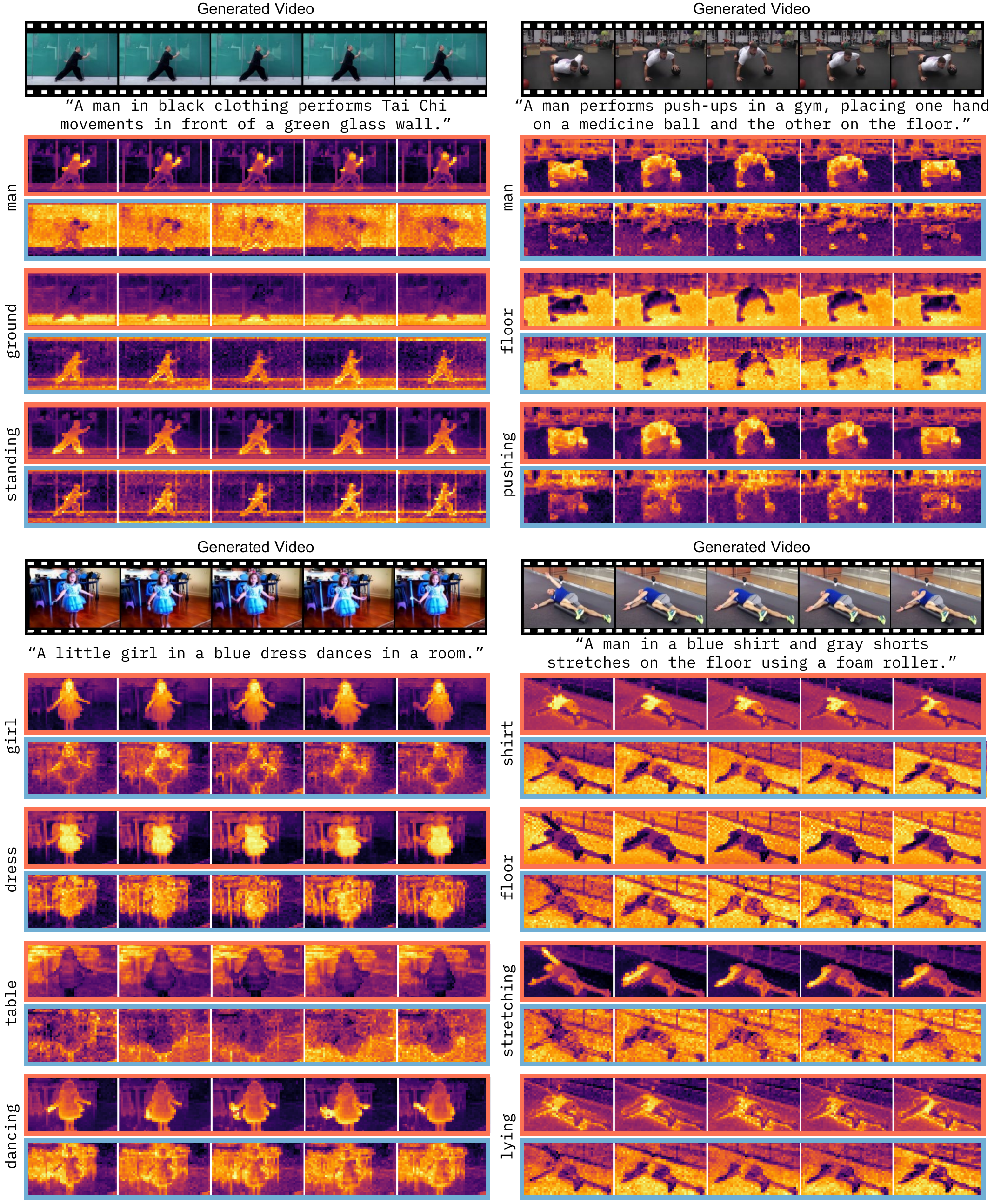}
   \caption{Comparison of IMAP under high ($>0.70$) vs.\ low ($<0.50$) average $\lambda_2$ layers in CogVidoX-5B
   }
   \label{fig:supp-highlow-2}
\end{figure*}

\begin{figure*}[h]
  \centering
   \includegraphics[width=\linewidth]{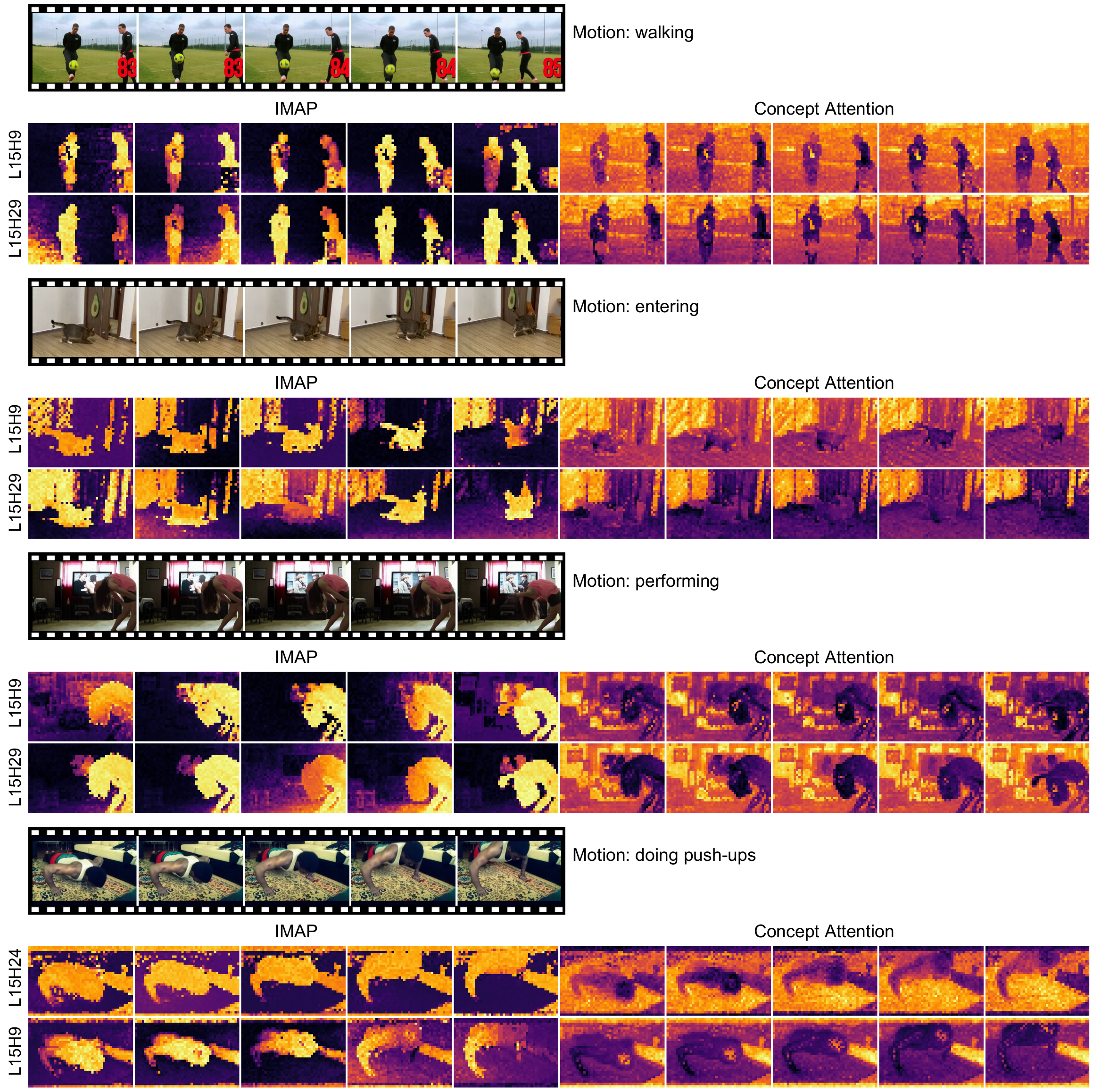}
   \caption{Examples of negative heads in CogVideoX-5B (L: layer, H: head)
   }
   \label{fig:supp-negative-heads}
\end{figure*}

Applying a softmax operation over concepts yields, for each voxel in the $F\times H\times W$ grid, a probability indicating which concept it represents best:
\begin{equation}
    \underset{c}{\mathrm{softmax}}(\mathrm{SaliencyMap}_c).
\end{equation}
In this process, the amplification and contraction caused by the exponential function in softmax function alter the resulting saliency map.
We quantitatively and qualitatively analyze how the saliency maps change under the softmax operation. 
Table~\ref{tab:supp-softmax} reports quantitative results for cross-attention, ConceptAttention, and IMAP extracted from HunyuanVideo and CogVideoX (2B/5B). 
For cross attention and ConceptAttention, applying the softmax operation improves almost all metrics. This indicates that these methods somewhat rely on this strategy and suggests that they may be vulnerable when irrelevant concepts are included in the concept list.
For example, if both \texttt{person} and \texttt{man} are present in the concept list, the highlights can be diluted.
In contrast, for IMAP, applying the softmax operation increases some metrics but decreases others, resulting in only marginal changes on average. In particular, the trade-off between SL and TL for CogVideoX-2B suggests that incorporating the softmax operation is not optimal for IMAP.

\begin{figure*}[h]
  \centering
   \includegraphics[width=0.9\linewidth]{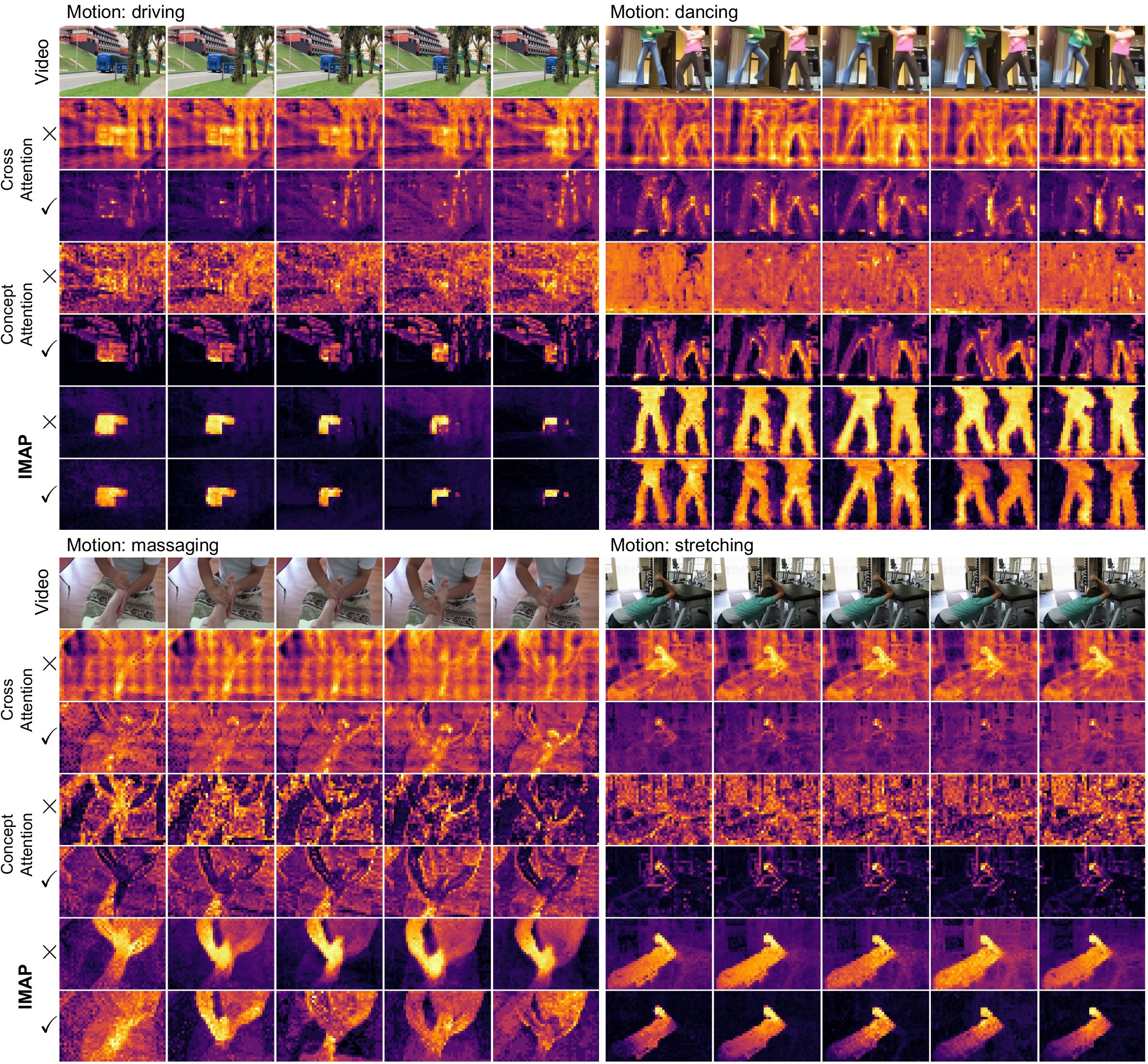}
   \caption{Qualitative comparison of the effect of applying the softmax operation ($\times$: without softmax, \checkmark: with softmax)
   }
   \label{fig:supp-softmax-1}
\end{figure*}

\begin{figure*}[h]
  \centering
   \includegraphics[width=0.9\linewidth]{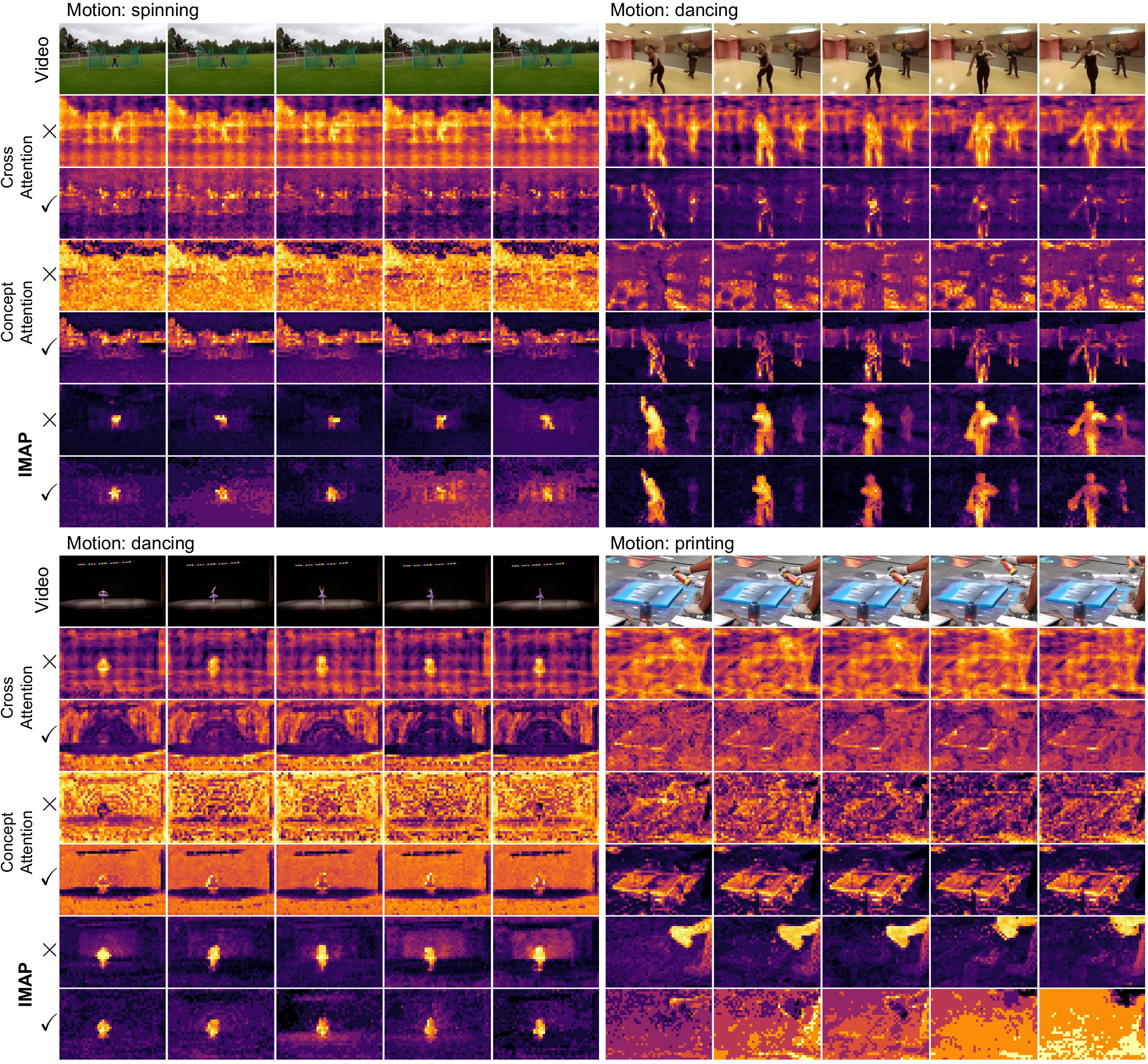}
   \caption{Qualitative comparison of the effect of applying the softmax operation ($\times$: without softmax, \checkmark: with softmax)
   }
   \label{fig:supp-softmax-2}
\end{figure*}

Related qualitative results for CogVideoX-5B are shown in Fig.~\ref{fig:supp-softmax-1} and Fig.~\ref{fig:supp-softmax-2}. 
In these figures, for cross-attention and ConceptAttention, the softmax operation generally makes semantic features more clearly emergent, although it occasionally distorts already meaningful structures. 
By contrast, for IMAP, softmax either has little effect or tends to distort the saliency maps. 
This demonstrates that our IMAP does not rely on the softmax operation; instead, we regard it as an unnecessary post-processing step.

\subsection*{B.5. Separation Scores}
\label{ssec:sup-separation}
\addcontentsline{toc}{subsection}{B.5. Separation Scores}

In Sec.~\ref{ssec:temporal}, we use the separation score of frame-wise clusters of visual-token embeddings as a criterion for motion-head selection. 
Candidate separation metrics include the Silhouette score~\cite{Silhouette}, Davies-Bouldin index (DBI)~\cite{DBI}, Calinski-Harabasz index (CHI)~\cite{CHI}, and the Fisher ratio~\cite{Fisher-ratio}. 
For DBI, lower values indicate higher separability, whereas for the other metrics, higher values indicate better separation.

We adopt CHI as the separation metric for computing the separation score. 
First, it is computationally efficient: according to Jain et al.~\cite{jain1999data}, when the number of datapoints is $P = FHW$, the number of clusters is $F$, and the data dimension is $d$, the time complexity is 
Silhouette score: $O(P^2 d)$, DBI: $O(P d + F^2 d)$, CHI: $O(P d)$, and Fisher ratio: $O(P d)$. 
Second, as shown in Fig.~\ref{fig:supp-sepscore}, the heads identified as highly separated are largely consistent across metrics (low-scoring heads for DBI and high-scoring heads for the others).

\begin{figure*}[h]
  \centering
   \includegraphics[width=\linewidth]{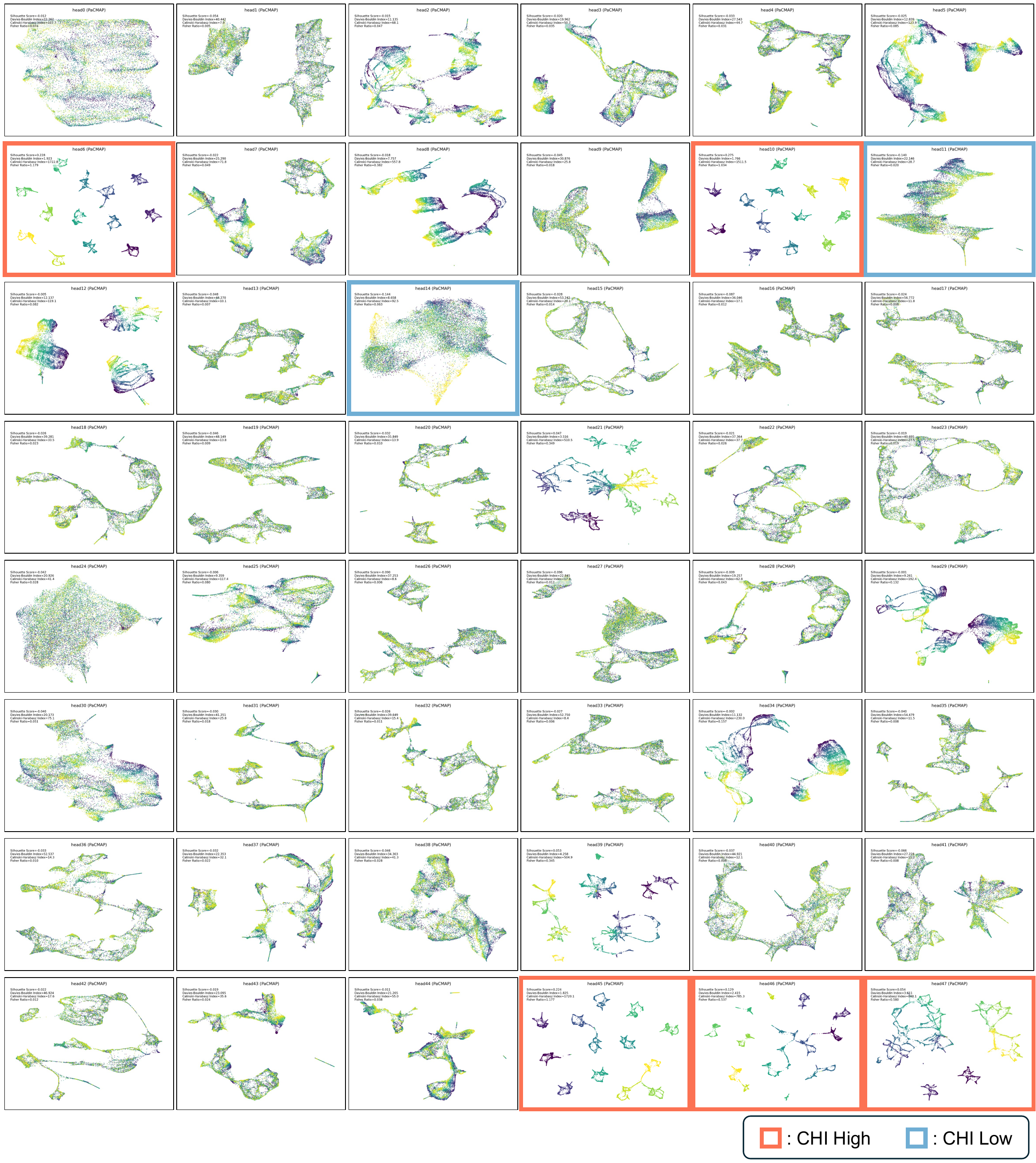}
   \caption{Visualization of visual token embeddings via dimensionality reduction~\cite{PaCMAP} (colors indicate frames)
   }
   \label{fig:supp-cluster}
\end{figure*}

Fig.~\ref{fig:supp-cluster} shows head-wise visualizations of visual token embeddings under different separation scores. Each grid corresponds to an attention head, and each color denotes a frame. We use PaCMAP~\cite{PaCMAP}, a dimensionality reduction method that preserves both local and global structure, and indeed observe that heads with clearly separated clusters exhibit higher CHI values.

\clearpage

\clearpage

\clearpage
\vspace*{\fill}
\vspace*{\fill}

\clearpage
\vspace*{\fill}
\vspace*{\fill}

\clearpage
\vspace*{\fill}
\vspace*{\fill}

\clearpage
\vspace*{\fill}
\vspace*{\fill}

\clearpage

\section*{C. Implementation Details}
\addcontentsline{toc}{section}{C. Implementation Details}

\subsection*{C.1. Evaluation Dataset: MeViS}
\label{ssec:sup-mevis}
\addcontentsline{toc}{subsection}{C.1. Evaluation Dataset: MeViS}

We use the MeViS~\cite{MeViS} train dataset as our dataset for evaluating motion localization. MeViS train dataset is a referring video object segmentation benchmark consisting of $1,662$ videos of varying lengths, each accompanied by expressions that describe a referred target object. To adapt the dataset to Video DiTs, we chunk each video into clips of $49$ frames and discard videos shorter than $49$ frames, resulting in $1,168$ clips with $49$ frames.

Next, we employ the large vision-language model Qwen3-VL~\cite{Qwen3-VL} to caption each video and annotate the concepts present in it. During annotation, we explicitly instruct the model to separate \textit{objects} that appear in the video from \textit{motions} associated with those objects. In parallel, we categorize camera movement into $17$ types and label each video accordingly, then filter out all clips with non-static cameras, since camera motion is not relevant for evaluating object-centric motion localization. 
After this filtering, we obtain a final set of $504$ videos with captions and concept lists. The prompts used for preprocessing, as well as the 504 captions and concept annotations, are provided in \boxlabel{Preprocess Prompt} and \boxlabel{MeViS Caption-Concept Pairs}. 

\begin{figure*}[h]
  \centering
   \includegraphics[width=0.54\linewidth]{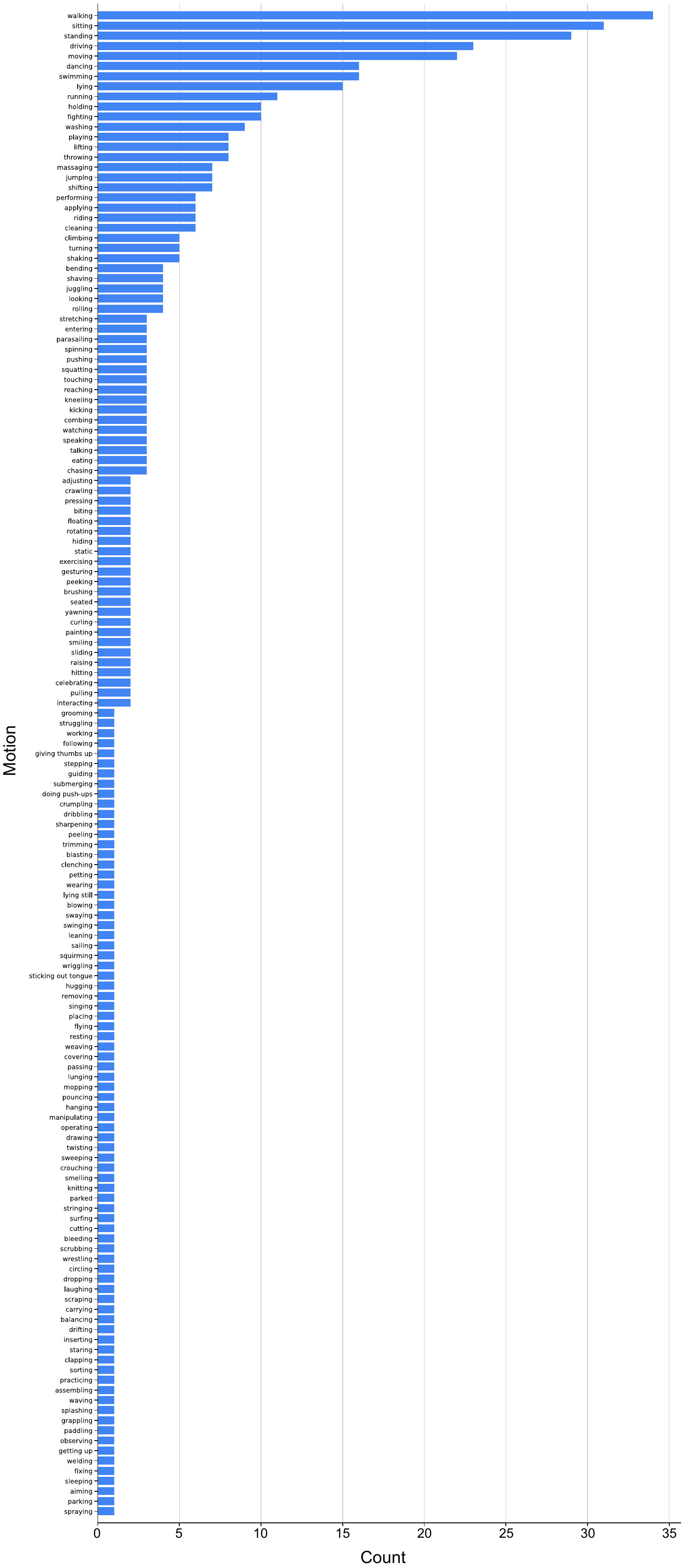}
   \caption{Counts of $150$ motion words in the preprocessed MeViS dataset
   }
   \label{fig:supp-dataset-motion}
\end{figure*}

\vspace{2pt}
For evaluation, we use exactly one motion word per video from the 504 videos, resulting in a total of 150 distinct motion words, as shown in Fig.~\ref{fig:supp-dataset-motion}.

\begin{preprocessprompt}
<SYSTEM>
  <ROLE>You are a precise video analyst.</ROLE>
  <OUTPUT_ONLY>
    Return ONLY a single XML block EXACTLY between these sentinels:
    <<<BEGIN_XML>>>
    <result>
      <caption></caption>
      <concepts></concepts>
      <camera></camera>
      <motion></motion>
    </result>
    <<<END_XML>>>
    No markdown or text outside. Any extra words outside XML are forbidden.
  </OUTPUT_ONLY>

  <TASK>
    <INSTRUCTIONS>
      Watch the video and read any provided "expressions" (if any).
      - <caption>: one concise English sentence, descriptive.
      - <concepts>: all distinct objects visible; comma-separated, lower-case, singular nouns only; no duplicates (e.g., "person, ball, tree").
      - <camera>: one or more labels from &lt;ALLOWED_CAMERA_LABELS&gt; (comma-separated, most prominent first); if none, "static".
      - <motion>: comma-separated UNIQUE single-word, specific verbs (lower-case) for visible actions AND non-actions (e.g., "standing", "sitting"). Only include if clearly visible; do not infer from context/audio/text; no directions/modifiers; never "moving"/"going"/"doing"; deduplicate; use "static" ONLY if no visible action/posture is identifiable.
    </INSTRUCTIONS>
  </TASK>

  <ALLOWED_CAMERA_LABELS>
    pan left, pan right, tilt up, tilt down, roll left, roll right, dolly in, dolly out,
    truck left, truck right, pedestal up, pedestal down, arc left, arc right, zoom in, zoom out, static
  </ALLOWED_CAMERA_LABELS>

  <STRICT_RULES>
    - Do NOT invent objects that are not visible.
    - concepts: nouns only, lower-case, singular; comma-separated; no duplicates.
    - camera: ONLY labels above; otherwise "static".
    - motion: visible-only single-word verbs; include actions/non-actions only when clearly seen; no directions/modifiers; never "moving"/"going"/"doing"; deduplicate; "static" only if none identifiable.
    - Ignore micro-jitter; treat as "static" unless clear camera motion exists.
    - Output must START with "<<<BEGIN_XML>>>" and END with "<<<END_XML>>>", with the <result> block in between. No extra text.
  </STRICT_RULES>
</SYSTEM>
\end{preprocessprompt}

\begin{captionconcepts}
"caption": Two people are dancing hand in hand in a room., "concepts": person, mirror, bed, shirt, pants, dress, dancing
"caption": Two people are dancing on a stage with red and white drapes., "concepts": person, stage, curtain, dancing
"caption": A boy stands on a sidewalk, preparing to jump., "concepts": boy, car, sidewalk, standing
"caption": A man performs martial arts movements in a wide stance on a blue mat., "concepts": man, mat, screen, flower, clothing, standing, performing
"caption": A giraffe walks toward a mesh-covered pole in an enclosure., "concepts": giraffe, pole, mesh, fence, tree, walking
"caption": man standing in front of the washbasin bending over to brush his teeth, "concepts": man, washbasin, toothbrush, toothpaste, watch, bottle, mirror, standing, bending, washing
"caption": A person washes their hands under a running faucet in a metal sink., "concepts": person, sink, faucet, bottle, hand, washing
"caption": A person jumps to dunk a basketball into the hoop., "concepts": person, basketball, hoop, tree, fence, hammock, pot, ground, jumping, dunking, holding
"caption": A pigeon walks on a paved surface., "concepts": pigeon, pavement, leaf, building, walking
"caption": A little girl in a pink dress rides a pink scooter while pushing a toy off the floor., "concepts": girl, scooter, toy, floor, door, rug, chair, couch, table, riding, pushing, holding, bending
"caption": A man kneels on the floor in a gym, adjusting a chain attached to a kettlebell., "concepts": man, kettlebell, chain, gym, tank top, shorts, floor, barbell, wall, kneeling, adjusting
"caption": Two mongooses groom each other on a sandy patch near rocks and vegetation., "concepts": mongoose, sand, rock, plant, grooming
"caption": A man juggles a football on a grass field while another man walks around him., "concepts": man, football, field, fence, light pole, juggling, walking, standing
"caption": A young boy in a yellow shirt smiles while using a head massager on his head., "concepts": boy, shirt, head massager, wall, smiling, massaging
"caption": Two dogs are fighting on a dirt ground., "concepts": dog, ground, fighting
"caption": Two dogs are fighting on a dirt ground., "concepts": dog, ground, fighting, pinning, struggling
"caption": Two dogs are fighting on a dirt ground., "concepts": dog, ground, fighting
"caption": A man hammers a wooden panel onto a shed frame while another man stands on the roof in the background., "concepts": man, shed, ladder, roof, brick wall, tree, sky, hammer, panel, hammering, standing, working
"caption": A woman is kneeling on the ground and stretching her arm on a gym bench., "concepts": woman, bench, gym equipment, door, floor, wall, stretching
"caption": A woman dances in a studio while a man stands in the background holding a phone., "concepts": woman, man, phone, floor, mirror, light, wall, dancing, standing
"caption": Two men are fighting with sticks in a parking lot., "concepts": man, stick, car, building, tree, fighting, standing
"caption": A woman with purple hair applies dye to her hair while wearing gloves., "concepts": woman, hair, glove, door, towel, tank top, applying
"caption": Two tigers fight in front of a rocky enclosure wall., "concepts": tiger, rock, wall, ground, pole, fighting, standing, walking
"caption": Two monitor lizards move through a rocky enclosure with logs and a cave-like structure., "concepts": monitor lizard, log, rock, cave, dirt, moving, crawling
"caption": A ballet dancer in a purple dress spins and dances on a stage under colored lights., "concepts": ballerina, dress, stage, lights, curtain, spinning, dancing
"caption": A young boy in a red cap sits on the ground and then stands up while holding a large green broom., "concepts": boy, cap, broom, cup, floor, motorcycle, standing
"caption": A person in a white shirt rides a bicycle on a paved path with another person on a bicycle behind them, set against a backdrop of green hills and a village., "concepts": person, bicycle, path, grass, hill, house, tree, vineyard, riding, following
"caption": A hand presses illuminated red and blue buttons on a music production device in a dark environment., "concepts": hand, device, button, screen, pressing
"caption": Motorboat and rider moving across choppy waves under a cloudy sky., "concepts": motorboat, rider, sea, wave, sky, riding
"caption": A tabby cat walks toward a doorway where an orange cat is peeking, then both cats enter the room together., "concepts": cat, door, floor, wall, avocado decoration, walking, entering
"caption": Two cats enter a room through a doorway., "concepts": cat, door, floor, wall, cat tree, running
"caption": A person is shaving a man's face with a straight razor., "concepts": man, person, razor, shaving cream, cape, towel, chair, shaving
"caption": A woman applies a substance to a man's leg while he sits on the edge of a bathtub., "concepts": man, woman, leg, bathtub, towel, container, brush, shirt, jeans, shelf, bottle, sitting, applying
"caption": A man stands in a backyard juggling flaming torches., "concepts": man, torch, trampoline, fence, tree, grass, pergola, standing
"caption": A girl performs a backbend in front of a television displaying a drama scene., "concepts": girl, television, person, couch, speaker, picture, curtain, carpet, ottoman, backbending, performing
"caption": A woman in a cow-print coat holds a helmet and gives a thumbs up., "concepts": person, coat, helmet, thumbs up, wooden wall, standing, holding, giving thumbs up
"caption": A baby lies on a colorful mat and bites its arm., "concepts": baby, mat, blanket, toy, clothing, lying, biting
"caption": A yellow pool cleaning machine floats on the water's surface., "concepts": machine, water, floating
"caption": A child in a dress is standing and moving slightly to the right., "concepts": child, dress, floor, standing
"caption": Two white geese walk across a concrete yard toward an open doorway, with more birds visible inside., "concepts": goose, doorway, concrete, poster, bird, walking
"caption": Monkeys interact on a dirt ground scattered with leaves., "concepts": monkey, leaf, ground, playing, sitting, moving
"caption": A man and a woman in formal attire dance on a reflective stage with a starry backdrop., "concepts": man, woman, dress, suit, bow tie, stage, lights, dancing
"caption": A man in a black tank top and shorts stands on a concrete circle in a grassy field, swinging his arms and kicking his leg forward before stepping back., "concepts": man, concrete circle, grass, tree, building, silo, ball, standing, kicking, swinging, stepping
"caption": A man trains two black bears in an arena with a stone wall backdrop, guiding one bear to stand on its hind legs and walk toward a metal frame while the other sits on a chair., "concepts": man, bear, chair, metal frame, stone wall, standing, walking, sitting, feeding, guiding
"caption": Two girls dance indoors in a room with a door and a water dispenser., "concepts": girl, door, water dispenser, curtain, floor, dancing
"caption": A duck submerges its head in shallow water while another duck stands nearby in a grassy wetland., "concepts": duck, water, grass, hill, submerging, standing
"caption": Two people are parasailing under a colorful parachute against a blue sky with scattered clouds., "concepts": person, parachute, rope, sky, cloud, parasailing
"caption": A person massages another person's foot., "concepts": person, foot, towel, floor, chair, massaging
"caption": Two dogs are moving around on a wooden floor., "concepts": dog, floor, moving, standing
"caption": A little girl climbs off a sofa., "concepts": girl, sofa, pillow, headband, climbing
"caption": a man doing push-ups on a patterned rug, "concepts": man, rug, couch, shirt, shorts, doing push-ups
"caption": A person operates a floor sander on a wooden floor, revealing a clean surface., "concepts": person, floor sander, floor, shoe, cord, standing
"caption": A child opens his mouth as an adult holds a yellow toothbrush near his teeth., "concepts": child, adult, toothbrush, opening, brushing
"caption": A man wearing a patterned shirt, headphones, and sunglasses speaks to the camera., "concepts": man, shirt, headphones, sunglasses, wall, poster, speaking, wearing
"caption": A young boy stands on a sidewalk, pouting with his lips pursed., "concepts": boy, sidewalk, tree, house, lamp post, car, standing
"caption": A person is windsurfing on water, holding the sail and standing on the board., "concepts": person, sail, board, water, glove, harness, standing, holding
"caption": A person rides a bicycle on a paved road, viewed from a first-person perspective., "concepts": bicycle, wheel, road, shoe, handlebar, riding
"caption": A person in a wheelchair is sitting while another person stands nearby., "concepts": person, wheelchair, shoe, seated, standing
"caption": A man and a woman dance together in a room with wooden floors., "concepts": man, woman, floor, window, table, speaker, hat, dress, pants, drum, dancing
"caption": A barge moves along a river near a construction site with buildings and a tower in the background., "concepts": barge, river, building, tower, construction site, dirt pile, boat, bridge, sky, moving
"caption": A barge moves forward along a river, passing a construction site with buildings and a water tower in the background., "concepts": barge, river, building, water tower, construction site, tire, boat, dirt pile, bridge, moving
"caption": A barge moves forward on a river past a construction site and buildings under a clear sky., "concepts": barge, river, building, construction site, chimney, bridge, dirt pile, tire, sky, moving
"caption": A baby yawns widely while being held., "concepts": baby, person, door, wall, blanket, holding, yawning
"caption": A human hand points at lizards moving on sand near a window., "concepts": lizard, hand, sand, window, pointing, crawling, lying
"caption": Lizards move around in a sandy enclosure near a glass window., "concepts": lizard, sand, glass, window, screen, crawling, lying still
"caption": A man blows up a red balloon while holding it with both hands., "concepts": man, balloon, television, bookshelf, blowing
"caption": A man performs a barbell squat in a gym while another man stands in the background., "concepts": man, barbell, gym, weight machine, floor, bag, ceiling, light, person, lifting, standing, squatting
"caption": A road intersection with vehicles moving through under a hazy sky., "concepts": truck, car, road, crosswalk, traffic light, tree, pole, building, driving, turning, moving
"caption": A hand touches and cleans a brown wall edge near a gutter., "concepts": hand, wall, gutter, watch, ring, brick, touching, cleaning
"caption": man stretching on the floor, "concepts": man, floor, wall, painting, gym equipment, dumbbell, barbell, radiator, light, ladder, stretching
"caption": A person swims freestyle in a pool lane., "concepts": person, pool, lane line, swimming
"caption": A colorful fish-shaped object hangs and sways on a pole in a grassy area near a track., "concepts": fish, pole, grass, track, chair, banner, swaying
"caption": A woman swings a stick in a dark outdoor setting while another person stands still nearby., "concepts": person, woman, man, stick, grass, cap, swinging, standing
"caption": A person uses scissors to cut a woven basket., "concepts": person, scissors, basket, hand, cutting, holding
"caption": Two cats are fighting on a tiled floor., "concepts": cat, floor, fighting, lying, standing
"caption": A street scene with a grassy area, trees, a sidewalk, a road, a green fence, and a person riding a motorcycle in the distance., "concepts": person, motorcycle, car, tree, grass, sidewalk, road, fence, sign, building, sky, riding, driving
"caption": a woman performs a backbend in a hallway, "concepts": woman, hallway, light, picture, door, backbending, performing
"caption": Two tigers interact on a grassy area near a pool, with one standing on the back of the other., "concepts": tiger, grass, pool, tree, rock, blanket, standing, climbing, leaning
"caption": A man sails on a board with a red and yellow sail across the ocean., "concepts": man, sail, board, ocean, sailing
"caption": A person juggles flaming objects at night in front of a tree., "concepts": person, fire, tree, shirt, juggling
"caption": Giraffes walk forward in a tiled corridor., "concepts": giraffe, corridor, wall, walking
"caption": A mother dog lies in a wooden box with her newborn puppies, then moves away, leaving the puppies to squirm and nurse on a blanket., "concepts": dog, puppy, blanket, wooden box, lying, moving, nursing, squirming
"caption": A mother dog enters a wooden box to nurse her newborn puppies, then leaves and returns multiple times., "concepts": dog, puppy, box, blanket, entering, nursing, leaving, returning, lying
"caption": Several newborn puppies, including black and white ones, are huddled together in a wooden box lined with fabric., "concepts": puppy, box, blanket, cloth, suckling, lying, wriggling
"caption": A hand massages a foot wearing a fuzzy sock., "concepts": hand, foot, sock, bracelet, table, couch, pillow, massaging
"caption": A woman performs a long jump, landing in the sand pit., "concepts": woman, sand pit, track, person, chair, flag, stadium, spectator, jumping, landing, sitting, kneeling
"caption": Two tigers walk along a dirt path behind a chain-link fence., "concepts": tiger, fence, path, tree, walking, standing
"caption": A man and a woman walk away from the camera on a grassy slope beside a road, with a tree trunk in the foreground., "concepts": man, woman, tree, road, grass, building, phone, walking
"caption": A man in a white shirt walks along a sidewalk while looking at his phone, partially obscured by a tree trunk in the foreground., "concepts": man, tree, sidewalk, road, grass, building, phone, walking, looking
"caption": A man and a young boy stick out their tongues while laughing., "concepts": man, boy, tongue, sticking out tongue
"caption": A tiger stands close to the camera while another tiger walks away in the background., "concepts": tiger, ground, building, tree, standing, walking
"caption": a standing man shaves a seated man's hair with a hair clipper, "concepts": man, hair clipper, wall, shaving, sitting, standing
"caption": A man walks across a paved area while looking at his phone., "concepts": man, phone, pavement, pillar, building, tree, grass, gazebo, car, walking, looking
"caption": A man in a pink helmet holds a Segway while interviewing two people on steps in front of a large building., "concepts": person, helmet, segway, microphone, building, steps, woman, man, jacket, sunglasses, text overlay, standing, interviewing, holding, turning
"caption": A person receives a foot massage on a red mat while lying down., "concepts": person, foot, mat, hand, leg, massage table, floor, metal shelf, massaging, lying
"caption": A black car drives on a road past a tree trunk., "concepts": car, tree, road, building, grass, sidewalk, sign, driving
"caption": A man dribbles two basketballs simultaneously on a gym court., "concepts": man, basketball, court, wall, driving
"caption": A man lies on a sofa hugging a baby., "concepts": man, baby, sofa, pillow, lying, hugging
"caption": A woman stands in a park holding a hula hoop, with a bicycle and backpack nearby., "concepts": woman, hula hoop, bicycle, backpack, tree, house, car, grass, bench, standing
"caption": A girl pushes a pink bicycle away from a blue bicycle on a paved surface., "concepts": bicycle, girl, pavement, bush, building, pushing, walking, standing
"caption": A girl pushes a pink bicycle past a parked blue bicycle., "concepts": bicycle, girl, pavement, bush, building, pushing, standing
"caption": A girl pushes a pink bicycle past a blue bicycle on a paved path., "concepts": girl, bicycle, pavement, bush, building, pushing, walking, standing
"caption": A little girl plays with a hula hoop in a living room., "concepts": girl, hula hoop, couch, chair, table, radiator, bookshelf, pillow, ball, blanket, playing
"caption": A red pole extends across a lake with a floating object and distant hills under a cloudy sky., "concepts": lake, hill, cloud, pole, object, floating
"caption": A baby lies on its back while hands stroke its legs., "concepts": baby, hand, bed, leg, nail polish, bracelet, lying, touching
"caption": A yellow fish swims near the bottom of an aquarium while a white fish swims in the background among driftwood and gravel., "concepts": fish, driftwood, gravel, aquarium, swimming, static
"caption": A yellow fish swims near the bottom of an aquarium while a large white fish swims in the background among driftwood and gravel., "concepts": fish, gravel, driftwood, swimming, standing
"caption": A white fish and a yellow fish swim in an aquarium with rocky structures and gravel bottom., "concepts": fish, rock, gravel, aquarium, swimming
"caption": a woman is removing hair from a man's back while he lies prone on a table, "concepts": man, woman, table, hair, back, gloves, lying, removing
"caption": A person walks on a sidewalk while a double-decker bus drives around a roundabout., "concepts": person, bus, bicycle, roundabout, sidewalk, tree, road, sign, construction vehicle, walking, driving, riding, turning
"caption": A man sings or talks while standing indoors., "concepts": man, curtain, wall, blind, singing
"caption": A woman throws a shot put inside a netted throwing circle., "concepts": woman, shot put, net, throwing circle, tree, ground, throwing, standing
"caption": A person swims underwater in a pool., "concepts": person, water, swimming
"caption": A young boy in a blue and white striped shirt kicks a soccer ball on a grassy field., "concepts": boy, soccer ball, grass, kicking
"caption": A girl uses a pink curling iron to style her hair., "concepts": girl, hair, curling iron, shirt, curtain, curling
"caption": A person in a yellow shirt kicks a soccer ball toward a goal on a grassy field., "concepts": person, soccer ball, goal, field, building, tree, kicking, standing
"caption": Two people are playfully fighting with swords in a grassy outdoor area., "concepts": person, sword, tree, ground, fighting, standing
"caption": A man juggling balls on a rooftop terrace., "concepts": man, ball, rooftop, potted plant, table, sky, clouds, juggling
"caption": Two birds walking through tall grass., "concepts": bird, grass, walking
"caption": A man in a gym prepares to lift a barbell loaded with weights., "concepts": man, barbell, weight, gym floor, sneaker, tank top, shorts, lifting
"caption": A man in a blue coat stands and places a blue ball into a transparent tube while another man squats nearby holding a similar tube., "concepts": man, ball, tube, coat, chair, standing, squatting, placing
"caption": A person paints a small figurine with a fine brush., "concepts": hand, figurine, brush, container, paint, table, painting
"caption": A girl with long dark hair is sitting and gesturing while smiling, with another person seated beside her., "concepts": girl, person, chair, table, cup, shirt, sitting, gesturing, smiling
"caption": Two geese fly over a patchwork of green fields and forests., "concepts": goose, field, forest, flying
"caption": View from inside a vehicle driving on a wet road during rain, with a truck ahead., "concepts": vehicle, truck, road, tree, raindrop, windshield, driving
"caption": A baby in pajamas crawls on a carpeted floor while wearing oversized shoes., "concepts": baby, pajamas, shoe, carpet, furniture, crawling, sitting
"caption": Two rabbits are sleeping together on wood shavings, with one brown rabbit resting on top of a white rabbit., "concepts": rabbit, wood shavings, sleeping, resting
"caption": Two young goats walk and jump on a wooden platform., "concepts": goat, platform, jumping, walking
"caption": Two young goats walk and play on a wooden platform., "concepts": goat, platform, walking, playing, jumping
"caption": A tiger cub walks around a large rock while another tiger stands on the rock., "concepts": tiger, rock, ground, walking, standing
"caption": A red car drives past on a road, moving from the foreground toward the background., "concepts": car, tree, road, grass, building, sign, driving
"caption": a woman in a red shirt and black pants is standing in a room with blinds and a door, performing arm movements., "concepts": woman, shirt, pants, window, blinds, door, carpet, standing, stretching
"caption": Two elephants are near a pool in an enclosure, with one drinking water and the other walking into the pool., "concepts": elephant, pool, tree, fence, ground, drinking, walking, standing
"caption": Two elephants walk into a pool of water in an enclosure., "concepts": elephant, pool, tree, fence, rock, ground, walking, entering
"caption": A young boy wearing a yellow shirt slides down a water slide., "concepts": boy, water slide, shirt, sliding
"caption": A male athlete performs a vault, spinning in the air above the vaulting horse in a crowded arena., "concepts": athlete, vaulting horse, crowd, arena, light, spinning, jumping
"caption": A person's hands weave bamboo strips into a circular frame., "concepts": hand, bamboo, frame, weaving
"caption": Two girls sit at a table, one drinking from a cup while the other shakes slightly in her seat., "concepts": girl, cup, table, t-shirt, straw, chair, counter, plant, window, light, wall, blanket, picture, drinking, sitting, shaking
"caption": A man wearing sunglasses covers his mouth with his hand against a red background., "concepts": man, sunglasses, hand, shirt, wall, covering
"caption": A young woman hands a red rose to an older woman sitting at a table., "concepts": rose, woman, old lady, table, cup, chair, flower, holding, sitting, passing
"caption": A woman combs her long dark hair with a brush., "concepts": woman, hair, brush, door, sweater, combing
"caption": A man shines his black boots using an automatic shoe polisher., "concepts": man, boot, shoe polisher, floor, standing
"caption": A man in a white shirt performs lunges with a barbell on his shoulders in a gym., "concepts": man, barbell, gym equipment, floor, shirt, shorts, shoes, lunging, standing
"caption": a person washes dishes in a kitchen sink using a yellow sponge, "concepts": person, dish, sink, sponge, cup, plate, faucet, dish rack, tube, washing
"caption": Two dogs, one white and one black and white, stand on a tiled floor with dappled sunlight., "concepts": dog, floor, standing
"caption": A man with a prosthetic leg stands on a circular platform on a grassy field, preparing to throw a shot put., "concepts": man, prosthetic leg, platform, grass, bottle, building, mountain, sky, standing
"caption": A little girl in a pink hoodie mops the tiled floor with a long-handled mop., "concepts": girl, mop, floor, door, hoodie, shoe, bucket, mopping
"caption": A man sits on a couch and shakes his head and arms vigorously., "concepts": man, couch, pillow, television, table, remote, cup, speaker, box, shaking
"caption": A white car drives past a directional sign on a grassy campus area., "concepts": car, sign, grass, road, tree, sidewalk, bus, crane, driving
"caption": A cat peeks out from a blue PlayStation 5 box on a countertop., "concepts": cat, box, countertop, litter box, peeking
"caption": A cat is inside a blue PlayStation 5 box, occasionally peeking out with its head and paws., "concepts": cat, box, table, litter box, peeking, hiding
"caption": A cat emerges from a blue PlayStation 5 box on a table., "concepts": cat, box, hand, table, emerging, reaching
"caption": Two fish swim among rocks in an aquarium., "concepts": fish, rock, swimming, standing
"caption": Two fish swim near rocks on the bottom of an aquarium., "concepts": fish, rock, swimming, standing
"caption": A fish with circular markings swims near rocks in an aquarium., "concepts": fish, rock, swimming
"caption": A fish swims over rocks in an aquarium., "concepts": fish, rock, swimming
"caption": Fish swim around rocks in an aquarium., "concepts": fish, rock, swimming, standing
"caption": A man wearing glasses looks upward while holding a plastic bottle, his head tilted back and mouth open., "concepts": man, glasses, bottle, book, shirt, looking up, holding, shaking
"caption": A person is cleaning a brown leather shoe with a brush while another person holds the shoe steady., "concepts": person, shoe, brush, stool, floor, cleaning, holding, sitting
"caption": Two people jump on a trampoline in a backyard., "concepts": person, trampoline, house, wall, tree, grass, shed, jumping
"caption": A white cat jumps onto a surface and pounces on an orange ball., "concepts": cat, ball, surface, bag, floor, jumping, pouncing
"caption": A white cat jumps onto a blue sofa while a red ball on a stick is visible., "concepts": cat, sofa, ball, stick, jumping
"caption": Two lizards fight on a paved surface., "concepts": lizard, pavement, fighting
"caption": A man in a green shirt plays a boxing arcade game while a woman stands nearby drinking., "concepts": man, woman, arcade game, drink, t-shirt, wall, playing, drinking, standing, shifting
"caption": A man sits on a purple exercise ball at a desk and then lies down., "concepts": man, exercise ball, desk, computer monitor, chair, seated, lying
"caption": Two pigeons are inside a metal cage, with one perched on a wooden pole and the other moving around., "concepts": pigeon, cage, pole, bowl, concrete block, standing, moving
"caption": Two pigeons move around inside a metal cage., "concepts": pigeon, cage, bowl, pole, concrete block, moving, standing
"caption": Two pigeons move around inside a metal cage., "concepts": pigeon, cage, pole, bowl, brick, moving
"caption": Two pigeons are inside a metal cage, one drinking from a bowl while the other moves around., "concepts": pigeon, cage, bowl, concrete block, drinking, moving
"caption": Two pigeons move around inside a wire cage., "concepts": pigeon, cage, pole, bowl, concrete block, jumping, standing
"caption": A man performs sit-ups on a gym machine., "concepts": man, machine, gym, floor, window, wall, towel, exercising
"caption": man throwing an axe at a wooden plank, "concepts": man, axe, wooden plank, tree, bag, throwing
"caption": A man spins and throws a discus in a stadium filled with spectators., "concepts": man, discus, stadium, crowd, net, track, bench, spinning, throwing, sitting
"caption": The two cats fight and wrestle on the grass., "concepts": cat, grass, pole, fighting, wrestling, rolling
"caption": a little girl in a pink shirt and diaper performs a somersault on a carpeted floor, "concepts": girl, diaper, shirt, carpet, pillow, flip flop, couch, doing somersault, performing
"caption": A man performs pull-ups on a wall-mounted bar in a gym., "concepts": man, pull-up bar, door, window, bench, weight plate, wall, exercising, turning, walking
"caption": A dog walks around a parked scooter on a concrete surface., "concepts": dog, scooter, concrete, walking, standing
"caption": A man stands on a self-balancing scooter in a room, extending his arms for balance., "concepts": man, self-balancing scooter, table, chair, clothing, curtain, wall, standing
"caption": A person braids a girl's hair in a room with fluorescent lighting., "concepts": person, girl, hair, hand, sweater, ceiling, light, braiding, sitting
"caption": A giant panda stands over a smaller panda lying on the ground, both surrounded by green foliage., "concepts": giant panda, panda, foliage, wall, standing, lying
"caption": A man lies face down on a table while another person applies a white strip to his back., "concepts": man, person, table, strip, room, lying, applying
"caption": Two bearded dragons interact on a carpeted surface, with one climbing on top of the other as a human hand occasionally touches them., "concepts": bearded dragon, carpet, hand, climbing, sitting, moving, touching
"caption": A woman in a red costume dances on stage in front of a banner and red backdrop., "concepts": woman, costume, banner, backdrop, stage, poster, dancing
"caption": A gray cat stands in a cage, moving its head and body slightly while another cat is partially visible in the background., "concepts": cat, cage, toy, bowl, standing, moving
"caption": person performing a backflip on a gymnastics mat in an indoor gymnasium., "concepts": person, mat, wall, window, floor, woman, gymnastics equipment, jumping, flipping, standing, sitting
"caption": A blue bus drives past a tree on a road., "concepts": bus, tree, road, building, fence, grass, driving
"caption": A bird with grey and blue feathers is perched on a person's hand near a pink cage., "concepts": bird, hand, cage, standing
"caption": A blue and grey parakeet perches on a person's hand near a pink cage., "concepts": bird, hand, cage, standing
"caption": A blue and grey parakeet perches on a person's hand near a pink cage., "concepts": bird, hand, cage, standing
"caption": A blue and grey bird is perched on a person's hand near a pink cage., "concepts": bird, hand, cage, pen, standing
"caption": A man rolls a yellow ball on his arm while standing in front of a closet., "concepts": man, ball, closet, hanger, t-shirt, wall, rolling
"caption": A blue bus drives away on a road near a grassy area and buildings., "concepts": bus, road, building, tree, sign, grass, sidewalk, driving
"caption": A blue bus drives away on a road near a building and trees., "concepts": bus, building, road, tree, sign, grass, sidewalk, driving
"caption": A blue bus drives past a road sign on a campus with buildings and trees in the background., "concepts": bus, road, sign, building, tree, grass, sidewalk, driving
"caption": A blue bus drives away on a road beside a grassy slope and buildings., "concepts": bus, road, building, tree, sign, grass, sidewalk, driving
"caption": A person braids the hair of a woman with long, wavy hair., "concepts": person, woman, hair, braid, headband, shirt, braiding, sitting
"caption": A toddler stands on a soccer ball on a grassy field while an adult stands nearby., "concepts": child, adult, soccer ball, grass, standing
"caption": A person in a white uniform with a green belt performs a high kick toward a wooden board held by another person in a white uniform with a black belt., "concepts": person, board, uniform, belt, mat, mirror, martial arts equipment, floor, wall, standing, kicking, holding, sitting
"caption": A man washes his hair with his hands., "concepts": man, hair, hand, foam, washing, standing
"caption": A baby in a polka-dotted shirt holds a tissue to their face and then lowers it., "concepts": baby, tissue, shirt, playpen, rubbing, sitting
"caption": A man and a woman dance together on a stage in front of an audience., "concepts": man, woman, audience, stage, banner, microphone, speaker, camera, chair, floor, dancing, sitting, standing, watching
"caption": A panda lies on a swing while another panda climbs a wooden structure in a grassy enclosure., "concepts": panda, swing, wooden structure, grass, tree, tire, bench, building, fence, lying, climbing
"caption": A panda sits on a swing in a grassy enclosure with wooden structures and tires, while another panda is on a log in the background., "concepts": panda, swing, log, tire, grass, tree, bench, building, wooden structure, sitting, hanging
"caption": A panda hangs upside down from a wooden structure while another panda lies on a swing in a grassy enclosure with trees and wooden play equipment., "concepts": panda, swing, wooden structure, grass, tree, tire, bench, building, person, hanging, lying, sitting
"caption": a person shaving their leg with a razor, "concepts": person, razor, leg, bracelet, tile, shaving
"caption": A young child pushes a shopping cart forward., "concepts": child, shopping cart, floor, pushing
"caption": A person dressed as Captain America cleans a window while two children watch and salute from inside., "concepts": person, child, window, costume, bucket, mask, building, rope, table, standing, cleaning, saluting, watching
"caption": A little girl in a blue dress dances in a room., "concepts": girl, dress, crown, wand, floor, chair, table, kitchen, stool, carpet, dancing
"caption": Two giraffes are fighting by entwining their necks., "concepts": giraffe, fighting
"caption": A person is cleaning a toilet with a blue toilet brush., "concepts": person, toilet, toilet brush, cleaning
"caption": A man wearing sunglasses sits at a table with hookahs and a beer bottle, talking and nodding., "concepts": man, sunglasses, table, hookah, beer bottle, cup, bowl, wallet, tattoo, shirt, nodding, talking, sitting, holding
"caption": A person's hands manipulate a metal bracelet against a blue background., "concepts": hand, bracelet, tool, watch, surface, holding, bending, manipulating
"caption": Two brown dogs interact on a dirt path scattered with debris., "concepts": dog, dirt, debris, plastic bag, stick, plant, moving, standing
"caption": Two brown dogs play on a dirt and gravel ground scattered with debris., "concepts": dog, ground, debris, dirt, gravel, plant, plastic bag, log, playing, standing, moving
"caption": A man walks behind two cows on a snowy path, leading them forward., "concepts": man, cow, path, snow, building, pole, cement mixer, walking
"caption": A little girl rides a scooter on a residential street., "concepts": girl, scooter, car, house, street, riding
"caption": A young boy drives a riding lawnmower through a wooded area while a man walks toward him., "concepts": boy, man, lawnmower, tree, fence, driving, walking, operating
"caption": A silver car drives past on a paved road., "concepts": car, road, pillar, building, tree, grass, sign, driving
"caption": A white car drives away on a paved road, partially obscured by a pillar., "concepts": car, pillar, road, tree, grass, sign, building, driving
"caption": A white car drives away on a paved road, partially obscured by a large white pillar., "concepts": car, pillar, road, tree, grass, sign, building, fire hydrant, driving
"caption": A white car drives past a paved area with a large pillar in the foreground., "concepts": car, pillar, road, tree, sign, fire hydrant, grass, building, driving
"caption": A white car drives along a road near a grassy area with trees and a building., "concepts": car, road, tree, building, grass, fire hydrant, sign, pavement, pillar, driving
"caption": A hand with glittery nail polish draws a bridge with a blue pencil on paper., "concepts": hand, pencil, paper, drawing
"caption": Blurred close-up of a person's face and shoulder, with indistinct background., "concepts": person, lying
"caption": Two monkeys interact on a rock, with one lying down and the other sitting and then walking away., "concepts": monkey, rock, grass, lying, sitting, walking
"caption": A person gives a massage to a woman lying on her stomach., "concepts": person, woman, massage table, hand, back, lying, massaging
"caption": person sitting and using a mobile phone, "concepts": person, mobile phone, sofa, gate, wall, using, sitting
"caption": A man in a red vest uses a long-handled tool to stir a container of material., "concepts": man, tool, container, vest, shirt, tie, scissors, box, cloth, pole, shaking, stirring, holding
"caption": an infant wearing a yellow hat lies on a patterned blanket, "concepts": infant, hat, blanket, lying
"caption": A woman with her hair in a ponytail raises her arms upward against a black background., "concepts": woman, arm, hand, hair, tank top, wristband, raising
"caption": Two people dance and wave their hands in a radio studio while a man talks on a television screen., "concepts": person, television, microphone, headset, table, computer, chair, plant, dancing, waving, sitting, watching
"caption": Two black birds are inside a red cage, with one standing on a perch and the other moving around on the bottom., "concepts": bird, cage, perch, heart-shaped decoration, green container, standing, moving
"caption": A woman lies on a bed while a man massages her back., "concepts": woman, man, bed, towel, floor, pillow, lying, massaging
"caption": A woman mops the floor while a black dog bites the mop head., "concepts": person, dog, mop, door, table, bottle, candle, slipper, floor, wall, plant, bag, mopping, biting, standing
"caption": Two people are dancing together in a large room with carpeted floors and windows along the back wall., "concepts": person, floor, window, light, partition, dancing
"caption": A woman in a belly dance costume stands on muddy ground, twisting her body., "concepts": woman, costume, ground, plant, twisting
"caption": A man sweeps the floor with a broom in a dimly lit indoor space., "concepts": man, broom, floor, chair, light, sweeping
"caption": A person gives a massage to another person lying face down on a bed., "concepts": person, bed, massage table, towel, floor, wall, window, lying, massaging, standing
"caption": A man lifts a barbell from his shoulders to above his head in a gym while others watch., "concepts": man, barbell, gym, weight, bench, box, mat, person, lifting, standing, sitting
"caption": A person's hand and arm are visible underwater, moving through green-tinted water., "concepts": person, hand, arm, water, swimming
"caption": A person crouches and shakes hands with a dog on a tiled floor., "concepts": person, dog, floor, cabinet, container, shaking, sitting, crouching
"caption": A man in a blue plaid shirt bumps his head on a hanging basket while another man stands behind him., "concepts": man, basket, wall, shirt, standing, hitting
"caption": A man lies on a bench pressing a barbell while another man stands nearby., "concepts": man, barbell, bench, gym equipment, person, lying, pressing, standing
"caption": A man shaves his beard with an electric razor in a bathroom., "concepts": man, beard, electric razor, shower curtain, shaving
"caption": A sea turtle swims gracefully through clear blue water., "concepts": sea turtle, water, swimming
"caption": A fluffy dog runs back and forth across a living room., "concepts": dog, cabinet, plant, cage, chair, curtain, painting, table, robot vacuum, running
"caption": A dog runs out from behind a curtain and across a room., "concepts": dog, cage, chair, curtain, plant, cabinet, pet carrier, table, running
"caption": person lying on the grass, "concepts": person, grass, lying
"caption": A dog swims in a stream and then runs onto the shore., "concepts": dog, stream, rock, grass, leaf, swimming, running
"caption": A dog runs along a rocky shore next to a body of water., "concepts": dog, rock, water, plant, running
"caption": Two people are parasailing under a red parachute with a smiling face design over the ocean., "concepts": person, parachute, ocean, sky, parasailing
"caption": A person in a blue shirt and hat paddles a kayak on a river while another person paddles alongside, surrounded by lush greenery., "concepts": person, kayak, paddle, river, tree, hat, shirt, paddling, sitting
"caption": Two people dance together on a dimly lit stage with musical instruments in the background., "concepts": man, woman, stage, musical instrument, microphone, drum, speaker, dancing
"caption": A shirtless man shaves his face with a razor in a bathroom., "concepts": man, razor, towel, door, hook, shaving cream, wristband, shaving, standing
"caption": A man in a red shirt holds a wooden box., "concepts": man, box, shirt, tool, floor, holding
"caption": A black tabby cat and an orange and white cat stand near a wooden bench outside a building on a sunny day., "concepts": cat, bench, building, bicycle, pavement, tree, standing, smelling
"caption": Two cats, one black and one orange and white, stand near a wooden bench outside a building while a bicycle remains parked in the background., "concepts": cat, bench, bicycle, building, pavement, hose, bucket, palm tree, standing, smelling, shifting
"caption": A woman in a red shirt and white gloves sits at a table and handles small metal sheets while speaking., "concepts": woman, table, glove, metal sheet, cabinet, tube, microscope, paper, folder, sitting, handling, speaking
"caption": A football player in a red uniform with the number 7 stands on the field while another player in a white uniform moves behind him., "concepts": player, helmet, uniform, field, standing
"caption": A woman in a white coat is smoking and talking against a dark background., "concepts": woman, coat, cigarette, smoking, talking
"caption": Hands are knitting with a crochet hook and red yarn., "concepts": hands, crochet hook, yarn, knitting
"caption": Two rabbits, one white and one black, are running around in circles inside a cage., "concepts": rabbit, cage, tray, grid, running
"caption": A dog sniffs boxes on a stage while a man guides it, with another man standing in the background., "concepts": dog, man, box, stage, person, glasses, leash, floor, sniffing, guiding, standing, looking
"caption": Two giraffes are fighting in a grassy field, swinging their necks at each other., "concepts": giraffe, grass, tree, fighting
"caption": Two giraffes fight in a grassy field using their necks to strike each other., "concepts": giraffe, grass, tree, fighting, standing
"caption": A person stands on the right side of a grassy field, pointing into the distance and then celebrating with their head up., "concepts": person, field, tree, mountain, ground, pointing, celebrating
"caption": Giraffes stand and walk within an enclosure at a zoo, with trees and a building in the background., "concepts": giraffe, building, tree, fence, grass, hand, standing, walking, bending
"caption": A man stands in a discus throwing ring, preparing to throw the discus., "concepts": man, discus, net, field, goal, tree, building, person, standing, spinning
"caption": A man juggles three balls on a gravel path surrounded by greenery., "concepts": man, ball, path, tree, mountain, power line, juggling
"caption": A man in a blue shirt and gray shorts stretches on the floor using a foam roller., "concepts": man, foam roller, shirt, shorts, shoe, floor, mirror, stretching, lying
"caption": A man lies on a bed while a woman applies a substance to his leg, with another woman standing nearby., "concepts": man, woman, bed, chair, curtain, bottle, tray, sock, shoe, lying, standing, applying
"caption": a child kicks a football across a grassy field, "concepts": child, football, field, tree, building, kicking
"caption": A man in a dark shirt massages the neck and shoulder of a man in a white shirt., "concepts": man, watch, ring, shirt, hand, neck, shoulder, massaging
"caption": A man in an orange shirt stands up from the floor while a woman dances nearby in a living room., "concepts": man, woman, couch, television, table, floor, wall, doorway, standing, dancing, kneeling
"caption": Two airplanes are parked on an airport tarmac under an overcast sky., "concepts": airplane, tarmac, sky, parked
"caption": Person cleaning the floor with a pole-like tool., "concepts": person, pole, floor, wall, cleaning
"caption": Two hands stringing pearls onto a thread with a beaded ornament., "concepts": hand, pearl, thread, ring, beaded ornament, stringing
"caption": Two people are in a room, one sitting with their legs extended on a chair while the other adjusts their leg., "concepts": person, chair, leg, sock, shorts, table, floor, seated, shifting
"caption": A dog climbs a set of stone stairs and exits the frame., "concepts": dog, stairs, wall, door, climbing
"caption": A baby sits on a striped pillow and shakes slightly., "concepts": baby, pillow, blanket, seated, shaking
"caption": Two women are dancing in a hallway., "concepts": woman, hallway, door, floor, wall, dancing
"caption": A young boy lies on a weight bench in a garage, lifting a wooden stick above his head., "concepts": boy, bench, stick, basketball, rug, bottle, garbage can, weight rack, sandal, door, lying, lifting
"caption": A hand combs through long brown hair with a black comb., "concepts": hair, comb, hand, combing
"caption": Two rabbits are near a hole in a sandy wall, with one entering the hole., "concepts": rabbit, hole, wall, sand, entering, standing
"caption": A shirtless man sits while a woman applies a hair removal pad to his chest., "concepts": man, woman, tape, chest, hand, wall, lying, applying
"caption": A man stands on a driveway, simulating a shot put throw with one hand on his neck., "concepts": man, driveway, house, fence, trash can, tree, sky, throwing, standing
"caption": A young boy slides down a colorful water slide into a pool., "concepts": boy, slide, pool, water, playground, sliding
"caption": Two cats eat from bowls in a cluttered outdoor area., "concepts": cat, bowl, cage, bucket, cardboard box, door, branch, blanket, eating, standing
"caption": A young boy wearing boxing gloves punches a red punching bag while a man watches., "concepts": boy, man, punching bag, boxing glove, ring, floor, stand, hitting, standing
"caption": a man in a blue shirt faces the camera, "concepts": man, shirt, wall, standing
"caption": A boy in a red shirt is waving his arm and speaking., "concepts": boy, shirt, hand, ceiling, window, waving, speaking
"caption": A person cleans a pottery piece in a metal bowl of water., "concepts": person, bowl, water, pottery, cleaning
"caption": A man sits on a decline bench in a gym, demonstrating a sit-up exercise., "concepts": man, bench, gym equipment, shoe, floor, wall, ball, machine, sitting
"caption": A car drives along a road while two people walk on a grassy slope in the background., "concepts": car, person, tree, road, sidewalk, grass, building, driving, walking
"caption": man lifting dumbbells in a gym, "concepts": man, dumbbell, mirror, window, kettlebell, tablet, poster, air conditioner, bench, person, lifting
"caption": A person in a red shirt runs down a paved road surrounded by greenery., "concepts": person, road, tree, grass, running
"caption": A person is being towed on a board by a paraglider over the ocean near a sandy beach., "concepts": person, paraglider, ocean, beach, sky, cloud, surfing
"caption": A person wearing a red barber cape stands in a kitchen while another person cuts their hair., "concepts": person, barber cape, kitchen, cabinet, sink, counter, hair, scissors, standing, cutting
"caption": A girl in a school uniform braids her hair while sitting on a patterned rug., "concepts": girl, hair, uniform, rug, shelf, watch, tie, wall, blackboard, braiding, sitting
"caption": A person's mouth is open with gloved hands holding the tongue and a tool inside., "concepts": person, mouth, tongue, hand, glove, tool, bandage, blood, holding, cutting, bleeding
"caption": A person washes their foot in a white bathtub., "concepts": person, bathtub, foot, nail polish, ring, tile, washing, standing
"caption": A person scrubs a bright green shoe with a white brush on a towel., "concepts": person, shoe, brush, towel, container, bottle, scrubbing
"caption": A man performs sit-ups on a mat in a gym., "concepts": man, mat, gym equipment, dumbbell, floor, lying, sitting
"caption": Two men stand on blue exercise balls and throw small balls to each other., "concepts": man, exercise ball, ball, wall, poster, floor, shoe, standing, throwing
"caption": Two tiger cubs playfully wrestle on a bed of dry grass., "concepts": tiger cub, grass, playing, lying, wrestling
"caption": Two giraffes stand and move in a circular pattern in a dry, grassy landscape., "concepts": giraffe, grass, dirt, standing, circling
"caption": A woman in a red jacket walks along a dock toward a traditional wooden boat, then stands and looks at it., "concepts": boat, woman, dock, water, tree, building, scarf, jacket, shoe, rope, pole, walking, standing
"caption": A traditional wooden boat floats on a calm lake near a dock, with a person in a red jacket standing on the shore., "concepts": boat, lake, dock, person, jacket, tree, tower, bridge, standing
"caption": Two dogs playfully wrestle in a garden area near a building., "concepts": dog, plant, tree, building, wall, pavement, playing
"caption": Two dogs playfully wrestle and chase each other in a leaf-covered yard., "concepts": dog, tree, leaf, fence, plant, playing
"caption": A man lifts a barbell from his shoulders to above his head, then drops it., "concepts": man, barbell, chair, wall, lifting, standing, dropping
"caption": Hands wearing gloves comb through hair extensions laid out on a plastic sheet., "concepts": hair extension, glove, comb, plastic sheet, brush, combing, standing
"caption": A girl puts her finger on her nose, then smells it and laughs., "concepts": girl, finger, nose, jacket, shirt, earring, bench, putting, smelling, laughing
"caption": A car drives along a road past trees and a green and white building., "concepts": car, road, tree, building, sidewalk, sign, grass, driving
"caption": A man windsurfing on the water with a sail., "concepts": man, sail, board, water, hill, standing
"caption": A man takes a shower with water spraying from the showerhead., "concepts": man, showerhead, water, curtain, bathroom curtain, tile, towel rack, standing
"caption": A black car drives down a road while two people walk on the sidewalk., "concepts": car, person, road, tree, grass, building, fence, driving, walking
"caption": A white and grey rabbit eats a green leaf on a concrete floor., "concepts": rabbit, leaf, floor, eating
"caption": Two cats play on a mat, one lying on its back while the other climbs on top., "concepts": cat, mat, bowl, container, stool, lying, climbing, rolling
"caption": Two cats are playing and rolling around on a mat., "concepts": cat, mat, bowl, container, stool, playing, rolling, lying
"caption": Two cats playfully wrestle on a mat, with one lying on its back and the other climbing on top., "concepts": cat, mat, table, stool, bowl, lying, wrestling, climbing
"caption": Two pandas are fighting each other near a wall., "concepts": panda, wall, plant, fighting
"caption": A person uses a large knife to scrape mud from their hand while standing outdoors near a tree stump., "concepts": person, knife, hand, mud, tree stump, grass, scraping, standing
"caption": A silver car drives past on a road while a man runs on the sidewalk., "concepts": car, man, road, sidewalk, tree, building, grass, pavement, running, driving
"caption": A silver car drives along a road lined with trees and a sidewalk., "concepts": car, road, sidewalk, tree, grass, building, driving
"caption": A woman works at a table outside while a man walks by carrying a pink object., "concepts": woman, man, table, fence, house, door mat, tool, backpack, wood, wall, light, standing, walking, carrying
"caption": A woman in a green jacket performs exercise movements in front of a mirror in a fitness room., "concepts": person, mirror, clock, jacket, pants, wall, floor, bottle, exercise mat, door, dancing
"caption": A little boy throws a yellow ball with a smiley face in a living room., "concepts": boy, ball, television, fireplace, shirt, guitar, throwing, standing
"caption": A brown cat walks through a doorway while another cat runs ahead into a bathroom., "concepts": cat, doorway, bathroom, litter box, cat tree, floor, walking, running
"caption": A cat chases another cat through a doorway into a bathroom., "concepts": cat, doorway, bathroom, litter box, cat tree, floor, doorframe, running, chasing
"caption": Two birds are inside a white cage, one perched on a branch and the other moving around., "concepts": bird, cage, car, tree, jumping, standing
"caption": A woman lies on her back on a mat while another person stands on her feet, balancing upside down., "concepts": person, mat, television, table, door, wall, statue, lying, standing, balancing
"caption": A man juggles balls in a sunny backyard garden., "concepts": man, ball, bench, door, tree, plant, building, table, juggling
"caption": A yellow boat drifts down a fast-moving river with rocky banks and trees under a partly cloudy sky., "concepts": boat, river, tree, mountain, rock, sky, cloud, drifting
"caption": A young girl sits in a salon chair wearing a black cape while getting her hair cut., "concepts": girl, cape, hairdresser, chair, salon, sitting
"caption": A man inserts a pole with a glowing end into a furnace., "concepts": man, pole, furnace, flame, tool, workshop, inserting, standing
"caption": A mother rabbit and her baby rabbits are in a yellow plastic basket., "concepts": rabbit, baby rabbit, basket, moving
"caption": A mother rabbit and her newborn kits are inside a yellow plastic basket., "concepts": rabbit, kit, basket, grass, shifting
"caption": A person swims breaststroke in a pool, creating splashes., "concepts": person, pool, water, swimming
"caption": a woman in a decorated bikini performs a belly dance in a dimly lit room while people watch in the background, "concepts": woman, bikini, crowd, microphone, stage, standing, dancing
"caption": A man lies on a bench press bench and lifts a barbell while another man stands nearby., "concepts": man, barbell, bench press, gym equipment, bench, weight, lying, lifting, standing
"caption": Two antelopes walk through a grassy area with trees and logs in the background., "concepts": antelope, tree, log, grass, walking
"caption": A husky dog and a lamb interact closely in a grassy field, with the dog appearing to herd the lamb., "concepts": dog, lamb, grass, leash, person, herding, standing, walking, running
"caption": A ginger cat plays with a stuffed toy on a wooden floor while a gray cat walks past., "concepts": cat, stuffed toy, floor, playing, walking
"caption": A woman lifts a barbell in a gym while another person stands up in the background., "concepts": person, barbell, weight, gym equipment, bucket, floor mat, wall, light, lifting, squatting, standing
"caption": A baby lies on a man's chest while both rest on a sofa., "concepts": man, baby, sofa, blanket, lying, resting, staring
"caption": A baby sits on the floor surrounded by toys while an adult's hands clap in front of them., "concepts": baby, hand, toy, floor, sitting, clapping
"caption": A man and a woman dance together in a ballroom., "concepts": man, woman, floor, mirror, chair, light, standing, dancing, kicking, turning
"caption": a person spray painting a blue canvas on the ground, "concepts": person, canvas, spray paint can, glove, bracelet, ground, art, paint, spray painting, sitting, painting
"caption": A woman with curly hair touches her hair while talking., "concepts": woman, hair, shirt, screen, touching, talking
"caption": A child combs an adult's hair with a brush while the adult smiles., "concepts": child, adult, brush, hair, wall, decoration, glasses, combing, smiling, standing
"caption": A man in a green striped shirt bends over in a cluttered room while another person sits on the left., "concepts": person, man, shirt, television, cabinet, bottle, bookshelf, bending, sitting
"caption": Two zebras interact in a grassy field, appearing to play or fight., "concepts": zebra, grass, tree, playing
"caption": Two workers in protective gear stand in a dusty industrial space, each holding a hose., "concepts": worker, hose, helmet, vest, glove, floor, pipe, standing
"caption": Two young goats stand and move on a wooden platform., "concepts": goat, platform, standing, moving
"caption": Two baby goats walk and play on a wooden platform., "concepts": goat, platform, fence, walking, playing
"caption": A person sorts dishes in a kitchen sink area., "concepts": person, dish, container, cup, utensil, window, bottle, sorting
"caption": A man bends forward and gestures with his hands while speaking to a boy standing beside him in a room., "concepts": man, boy, car, curtain, window, radiator, floor, jacket, shirt, pants, ring, standing, bending, gesturing, looking, holding
"caption": person running on a treadmill, "concepts": person, treadmill, window, plant, blind, running
"caption": A man throws a frisbee toward a disc golf basket in a graffiti-covered skate park., "concepts": man, frisbee, disc golf basket, car, graffiti, tree, hill, sky, throwing, standing
"caption": A baby sits in a high chair and claps its hands., "concepts": baby, high chair, bib, clapping, sitting
"caption": Two men are fighting in a bedroom, with one man jumping out of bed and the other man attacking him., "concepts": man, bed, television, nightstand, blanket, pillow, fighting, jumping, standing, sitting
"caption": A young girl washes dishes at a kitchen sink while a boy stands behind her., "concepts": girl, boy, dish, sink, soap, bottle, window, countertop, cabinet, faucet, dish soap, dish brush, washing, standing
"caption": A person operates a metal instrument on a white cloth., "concepts": person, instrument, cloth, hand, stick, holding, pulling
"caption": A person performs several backflips on a blue mat in a gym., "concepts": person, mat, gym, wall, doing backflips, performing
"caption": Two men are practicing martial arts movements in a room with large windows., "concepts": man, room, window, floor, moving, practicing
"caption": A person holds a bow and arrow, aiming at targets on a grassy field from a wooden platform., "concepts": person, bow, arrow, target, grass, platform, standing
"caption": Two zebras are walking and turning around in an enclosure with a wooden fence and trees., "concepts": zebra, fence, tree, building, car, ground, walking, turning
"caption": A young boy lies down while someone washes his hair with soap., "concepts": boy, hand, hair, soap, lying, washing
"caption": Two people are working on constructing a wooden cabin in a wooded backyard., "concepts": person, cabin, tree, fence, tool, ladder, wood, saw, table, ground, building, walking, standing
"caption": A man in a red shirt and green apron is assembling white wooden panels while kneeling on the floor., "concepts": man, panel, floor, apron, shirt, kneeling, assembling
"caption": man walking and waving his arms in a park, "concepts": man, tree, road, shadow, walking, waving
"caption": A man picks up a long green object from the ground and stands upright on an indoor track., "concepts": man, object, track, bench, building, standing
"caption": A boy crawls through water on a pool floor while splashing., "concepts": boy, pool, water, wristband, swim trunks, wall, crawling, splashing
"caption": A blue bus drives past a bus stop on a road lined with trees and buildings., "concepts": bus, car, tree, building, road, grass, bus stop, sign, streetlight, driving
"caption": A woman with red hair uses a curling iron on a section of her hair., "concepts": woman, hair, curling iron, shirt, door, holding, curling
"caption": Two zebras fight in a dusty field, kicking and grappling with each other., "concepts": zebra, dust, grass, tree, fighting, kicking, grappling, standing
"caption": A yellow kayak moves down a river with a red kayak visible ahead, both being paddled by people., "concepts": person, kayak, paddle, river, tree, rowing, paddling
"caption": Two colorful parrots interact inside a blue cage, with one preening and the other observing., "concepts": parrot, cage, egg, preening, standing, observing
"caption": a woman cutting a child's hair while the child sits in a chair, "concepts": child, woman, chair, hair, comb, scissors, cape, cutting, sitting, standing
"caption": A man in a yellow shirt stands behind a woman on a self-balancing scooter on a brick walkway., "concepts": man, woman, self-balancing scooter, brick, chair, table, building, standing, riding
"caption": Two people are swimming underwater in clear blue water., "concepts": person, swim fin, snorkel, water, swimming
"caption": Two dogs are playfully fighting and wrestling on a tiled floor., "concepts": dog, floor, wall, box, fighting, playing
"caption": A little girl crawls on a carpet., "concepts": girl, carpet, crawling
"caption": A man wearing a glove applies a liquid from a bottle to a black shoe., "concepts": man, shoe, bottle, glove, watch, fan, holding, applying
"caption": A woman twirls two red-and-black sticks in a park-like setting while a dog walks in the background., "concepts": woman, stick, dog, building, tree, trash can, bench, twirling, walking, sitting
"caption": A person sprays black paint onto a piece of paper on a white tarp., "concepts": person, spray paint can, paper, tarp, shoe, spraying, kneeling
"caption": A high jumper in black lands on a blue mat and sits up., "concepts": athlete, mat, pole, stadium, person, track, flag, sitting, getting up
"caption": a person swimming in a pool, "concepts": person, pool, swimming
"caption": A woman holds a baby in an orange wrap against a white paneled wall., "concepts": woman, baby, wrap, wall, holding
"caption": Two ducks walk across a dry, leaf-covered ground near a blue exercise bar., "concepts": duck, ground, exercise bar, tree, wall, walking
"caption": Hands are washing in a sink with soap and water., "concepts": hand, sink, soap, faucet, washing
"caption": A man in a gym lifts a dumbbell while standing in front of a mirror., "concepts": man, dumbbell, mirror, gym equipment, cap, tank top, shorts, lifting, standing
"caption": A person wearing a welding helmet welds a metal piece, producing a bright flash of light., "concepts": person, welding helmet, metal, welding torch, light, welding
"caption": A woman stands in a living room, tossing and catching a bright yellow frisbee while speaking., "concepts": woman, frisbee, living room, plant, table, door, television, lamp, picture, necklace, shirt, skirt, throwing, catching, speaking, standing
"caption": A woman holds a baby while adjusting a blanket in front of a plain wall., "concepts": woman, baby, blanket, wall, holding, adjusting, standing
"caption": A man in a green shirt stands inside a netted enclosure, preparing to throw a hammer., "concepts": man, shirt, shorts, hammer, net, ground, tree, mountain, standing
"caption": a man sitting in a wheelchair, "concepts": man, wheelchair, t-shirt, jacket, shorts, mobile phone, seated
"caption": A lizard runs across a wooden floor in a room, chasing a small object., "concepts": lizard, floor, computer, bag, door, hand, toy, chair, running, chasing, lying, eating
"caption": a man throws a ball to the wall and catches it, "concepts": man, ball, wall, door, fire extinguisher, mat, rope, gym equipment, throwing, catching, shifting
"caption": A child drinks from a sippy cup and then opens their mouth wide., "concepts": child, sippy cup, drinking, yawning
"caption": A person is dancing in a room with a blue floor., "concepts": person, floor, pants, shoes, dancing
"caption": A woman with green hair fixes her hair while standing in front of a wall with a towel and a phone on a stand., "concepts": woman, hair, towel, phone, stand, wall, hook, tank top, fixing, standing
"caption": Hands are washing under a faucet with soap suds in a sink., "concepts": hand, sink, faucet, soap, countertop, bottle, washing
"caption": man throwing a frisbee on a beach, "concepts": man, frisbee, beach, sailboat, umbrella, throwing
"caption": A baby sleeps peacefully with a blanket over its head., "concepts": baby, blanket, sleeping
"caption": A man is weaving a basket with his hands., "concepts": man, basket, hands, tool, watch, shirt, apron, weaving, sitting
"caption": A young child rides a green scooter across a tiled floor., "concepts": child, scooter, floor, door, riding
"caption": Two people are parasailing under a colorful parachute over the ocean., "concepts": person, parachute, ocean, boat, mountain, parasailing
"caption": A shirtless man stands in a bathroom and speaks to the camera., "concepts": man, shower curtain, towel, picture, wall, shower curtain rod, loofah, shelf, talking
"caption": A close-up of a person's face showing their mouth and chin, with a white textured surface in the background., "concepts": person, cloth, lying
"caption": A man kicks a football on a field and runs forward., "concepts": man, football, field, goalpost, tree, running, kicking, celebrating
"caption": Two white rabbits are inside a wire cage with leaves on the floor., "concepts": rabbit, cage, leaf, standing, moving, climbing
"caption": a little girl sits on the floor playing with a string of beads, "concepts": girl, bead, string, toy, box, floor, bracelet, phone, sitting
"caption": a little boy brushes his teeth and reaches into the sink, "concepts": boy, toothbrush, sink, cabinet, towel, brushing, reaching
"caption": A silver car drives past a person walking while looking at their phone., "concepts": car, person, phone, pillar, road, building, tree, grass, canopy, speed bump, walking, looking
"caption": A man lifts a barbell from the floor to his shoulders in a weightlifting competition., "concepts": man, barbell, scoreboard, platform, curtain, banner, lifting, standing
"caption": Several kittens are inside a blue cage in a pet shop, moving around and climbing on a yellow platform., "concepts": kitten, cage, platform, pet shop, bowl, shelf, container, standing, climbing, shifting
"caption": Two kittens are inside a blue cage, with one climbing the side while the other remains seated., "concepts": kitten, cage, bowl, shelf, container, climbing, sitting
"caption": A man in a dark jacket walks up a set of outdoor stairs., "concepts": man, jacket, stairs, car, sidewalk, tree, wall, leaf, walking
"caption": A car drives past on a road beside a tree with a mottled trunk., "concepts": car, tree, road, sign, building, grass, sidewalk, driving
"caption": A brown dog stands on bricks near a doorway, then jumps down and interacts with another dog., "concepts": dog, brick, doorway, mat, car, tree, house, trash, cloth, barrel, standing, jumping, interacting
"caption": Two puppies playfully fight and chase each other on a concrete ground near a pile of clothes and a concrete pillar., "concepts": puppy, concrete pillar, clothes, plastic bag, concrete ground, bowl, person, jeans, playing, fighting, chasing, standing
"caption": Two puppies playfully fight on a white tarp in an outdoor area., "concepts": puppy, tarp, concrete, pillar, bag, bench, building, fighting
"caption": Two puppies, one black and one light-colored, playfully wrestle and chase each other on a concrete ground near a pile of fabric and a large pillar., "concepts": puppy, pillar, fabric, concrete, bag, brick, bowl, door, wall, playing, wrestling, chasing
"caption": a woman in a white shirt and cap is aiming a bow and arrow in a wooded area, "concepts": woman, bow, arrow, cap, shirt, pants, forest, standing, aiming
"caption": A little girl is cleaning a toilet in a bathroom., "concepts": girl, toilet, brush, cabinet, plant, wall, cleaning
"caption": A man parks a bicycle next to a wooden fence., "concepts": man, bicycle, fence, grass, helmet, parking
"caption": a person throws an axe at a target in an indoor facility, "concepts": person, axe, target, wall, floor, throwing
"caption": A woman sits and laughs while shaking her body back and forth in a dimly lit room with colorful lights and a dartboard in the background., "concepts": person, woman, man, dartboard, light, poster, shirt, laughing, sitting, shaking
"caption": A young elephant uses its trunk to interact with a person lying on the grass in a grassy field., "concepts": elephant, person, grass, tree, canopy, feeding trough, lying, interacting, pulling
"caption": A person lies on the grass while an elephant uses its trunk to interact with the person., "concepts": person, elephant, grass, tree, bench, lying, interacting
"caption": Two elephants playfully interact in a grassy field under a cloudy sky., "concepts": elephant, grass, sky, playing, standing
"caption": A man stands on a dark stage under a bright light, bending over then raising his arms while holding two balls., "concepts": man, ball, light, standing, raising
"caption": A person in blue clothing stands on a wooden plank atop scaffolding, spraying a wall with a hose., "concepts": person, scaffolding, plank, wall, hose, pipe, bucket, standing, spraying
"caption": A person trims a rabbit's nails with a clipper while holding it gently., "concepts": person, rabbit, nail clipper, towel, blanket, table, chair, trimming, holding, sitting
"caption": A baby is sleeping while being held., "concepts": baby, blanket, shirt, sleeping, shifting
"caption": A baby lies on a cushion and moves its hands., "concepts": baby, cushion, blanket, pillow, lying, moving
"caption": A person in a blue jacket is sitting near a striped bed with their legs moving., "concepts": person, bed, jacket, leg, carpet, moving
"caption": A man spins and throws a shot put from a circular throwing area., "concepts": man, shot put, net, field, building, circle, throwing, spinning, standing
"caption": A man fishing in a river while a bear walks through the water nearby., "concepts": man, bear, river, fishing rod, cap, jacket, rocks, tree, standing, fishing, walking
"caption": A man fly fishing in a river while a bear walks through the water nearby., "concepts": man, bear, river, fishing rod, cap, jacket, sunglasses, standing, fishing, walking
"caption": A girl stands in a room, talking and gesturing with her hands raised., "concepts": girl, tank top, ceiling light, door, picture, basketball hoop, wall, ceiling, cross, standing, talking, gesturing
"caption": A man in a blue tank top and white shorts runs across a gymnasium floor in front of a crowd., "concepts": man, crowd, floor, tank top, shorts, banner, audience, gymnasium, running
"caption": A man shoots a basketball and takes several steps back on an indoor court., "concepts": man, basketball, hoop, court, ball, bench, wall, shooting, walking, standing
"caption": A gloved hand holds a spray gun near a tiled wall with a "Sierra Vista Pools" sign., "concepts": hand, spray gun, tile, wall, glove, holding
"caption": A woman massages a man's leg while he lies on a table covered with a blanket., "concepts": person, man, woman, leg, blanket, table, towel, pillow, bracelet, massaging, lying, standing
"caption": Two small animals are grooming each other while lying on the ground., "concepts": animal, rock, plant, sand, grooming, lying
"caption": Two pigeons stand on a person's hand, moving their heads while remaining in place., "concepts": person, pigeon, hand, tree, standing, moving
"caption": A hand holds a knife against a vertical surface, showing the blade., "concepts": hand, knife, surface, holding
"caption": A blue and red bus drives on a road past a grassy hill and trees., "concepts": bus, car, road, tree, grass, sidewalk, person, streetlight, driving, walking
"caption": Two pandas playfully wrestle and climb on a wooden structure., "concepts": panda, wooden structure, tree, bamboo, log, grass, rock, wall, playing, climbing, wrestling, sitting, standing
"caption": Two pandas are playfully wrestling on a wooden platform in an enclosure., "concepts": panda, platform, wooden structure, tree, bamboo, rock, grass, fighting, climbing, lying, rolling
"caption": Two pigeons are inside a metal cage, one standing still and the other moving around., "concepts": pigeon, cage, standing, moving
"caption": A white robot with a blue light on its chest moves its arms in a dimly lit environment., "concepts": robot, light, dancing
"caption": A woman stands and rotates her arms while holding dumbbells., "concepts": woman, dumbbell, t-shirt, shorts, floor, wall, standing, rotating
"caption": Two cats play on a wooden floor near a cardboard box., "concepts": cat, cardboard box, floor, playing, lying, hiding
"caption": A woman stands at a table with a bouquet of flowers, adjusting them., "concepts": woman, bouquet, table, microphone, curtain, vase, flower, box, standing
"caption": A little girl stands near a table, talking and crumpling paper while a TV plays in the background., "concepts": girl, table, paper, tv, toy, cabinet, book, stuffed animal, tissue, standing, talking, crumpling
"caption": Two children play with hula hoops on a patio., "concepts": child, hula hoop, table, chair, bush, car, house, spinning, standing
"caption": A white vehicle is stuck in a body of water, partially submerged and stationary., "concepts": vehicle, water, grass, tree, static
"caption": A man walks away while dribbling two basketballs in a gym., "concepts": man, basketball, gym floor, door, walking, dribbling
"caption": Hands play a small toy drum set on a green table., "concepts": drum, hand, table, toy drum set, playing
"caption": a girl stands on a bed and bends backward, then stands up straight, "concepts": girl, bed, window, pillow, painting, pajamas, standing, bending
"caption": A man performs push-ups in a gym, placing one hand on a medicine ball and the other on the floor., "concepts": man, medicine ball, floor, gym equipment, t-shirt, shorts, watch, pushing
"caption": A man performs a deadlift with a barbell while another man stands nearby in a gym., "concepts": man, barbell, weight, gym, fan, flag, scoreboard, chair, pole, bench, weight plate, floor, wall, door, backpack, shirt, shorts, shoes, belt, lifting, standing
"caption": A man in black clothing performs a one-legged jump in front of a plain wall., "concepts": man, clothing, shoe, wall, jumping
"caption": A person uses a power drill to sharpen a pencil., "concepts": person, pencil, drill, tool, hand, holding, sharpening
"caption": Two bulls are fighting by pushing their heads against each other on a dirt ground., "concepts": bull, man, rope, dirt, fighting, standing, walking, squatting
"caption": A man in black clothing performs Tai Chi movements in front of a green glass wall., "concepts": man, clothing, glass wall, ground, standing
"caption": A man with blonde hair is shaking his head and touching his neck., "concepts": man, t-shirt, door, chair, table, rug, wall, tattoo, shaking, touching
"caption": A person's hands operate a Korg nanoPAD2 music controller, adjusting knobs and pressing pads., "concepts": person, device, watch, button, knob, pad, cable, adjusting, pressing, rotating
"caption": A woman in a red top and plaid shorts performs a high-knee exercise in a grassy yard., "concepts": woman, top, shorts, grass, tree, exercising
"caption": A baby lies on a wooden floor and reaches for a green building block., "concepts": baby, building block, floor, lying, reaching
"caption": A white adult rabbit nuzzles several small white and black baby rabbits inside a wire cage., "concepts": rabbit, baby rabbit, cage, blanket, hay, standing, nuzzling, lying
"caption": A white rabbit tends to its newborn kits inside a wire enclosure., "concepts": rabbit, kit, cage, hay, blanket, lying, moving, standing
"caption": A man prepares to lift a barbell in a gym., "concepts": man, barbell, gym equipment, floor, shirt, pants, mirror, standing
"caption": A blue bus drives past a tree with ferns on a street with a crosswalk., "concepts": bus, tree, fern, crosswalk, road, sign, grass, sidewalk, building, driving
"caption": A blue bus drives away from the camera on a road beside a grassy area with trees., "concepts": bus, tree, road, grass, sidewalk, crosswalk, building, sign, fence, fern, driving
"caption": A baby in a yellow walker reaches toward a mop on the floor., "concepts": baby, walker, mop, couch, stroller, floor, person, reaching, sitting
"caption": A person in a wetsuit rides a surfboard on a wave., "concepts": person, surfboard, wave, ocean, riding
"caption": a person swims freestyle in a pool lane, "concepts": person, pool, lane divider, swimming
"caption": Two rabbits, one brown and one white, are inside a metal cage with a water container., "concepts": rabbit, cage, water container, pipe, fabric, standing, moving
"caption": A hand peels a strip of tape from a wall., "concepts": hand, tape, wall, peeling
"caption": A woman washes her braided hair in a white bathtub., "concepts": woman, hair, bathtub, wall, shirt, washing, bending
"caption": A person trims a dog's nails while holding it., "concepts": person, dog, trimming
"caption": Two pandas playfully wrestle and climb over a tree branch in an enclosure., "concepts": panda, tree, branch, ground, straw, playing, wrestling, climbing, sitting
"caption": A young girl with a runny nose looks at the camera., "concepts": girl, nose, coat, shirt, standing
"caption": A woman in a blue checkered shirt speaks and gestures in a bedroom., "concepts": woman, shirt, lamp, bed, basket, door, necklace, ceiling light, wall, floor, speaking, gesturing, standing
"caption": person walking on a treadmill, "concepts": person, treadmill, shoe, blanket, walking
"caption": A boy sits on the floor holding a yellow broom., "concepts": boy, broom, dustpan, floor, shirt, shorts, sitting
"caption": A group of four ducklings moves together on dirt near a tree and a woven basket., "concepts": duckling, tree, basket, dirt, wood, brick wall, standing, walking
"caption": A close-up of a sleeping child's face, showing subtle movements as they turn slightly., "concepts": child, pillow, sleeping, turning
"caption": A man performs push-ups with his feet elevated on a red exercise ball in a gym., "concepts": man, exercise ball, gym, mirror, floor mat, shirt, pants, shoes, fan, plant, weight machine, doing push-ups, exercising, performing
"caption": A sea turtle swims near the surface of the water in an aquarium tank., "concepts": sea turtle, water, aquarium, swimming
"caption": A technician in protective gear sand blasts a floor inside a wooden room., "concepts": technician, hose, floor, window, protective suit, mask, wall, tree, standing, sand blasting, blasting
"caption": A young child wipes their nose with a tissue while a dog moves nearby in a living room., "concepts": child, dog, couch, pillow, tissue, table, floor, wall, picture, wiping, moving, sitting
"caption": A boy sits on a black leather couch, clenching his fists and shaking his head., "concepts": boy, couch, watch, shaking, clenching
"caption": A tabby cat lies on a wooden bench while another cat sits under a nearby bench; a human hand pets the cat on the bench., "concepts": cat, bench, hand, pole, tree, ground, lying, sitting, petting
"caption": A cat lies on a wooden bench while a person's hand pets it, with another cat visible under the bench., "concepts": cat, person, bench, hand, pole, tree, lying, petting, peeking
"caption": A woman lies on a red mat performing a leg exercise while another woman kneels and speaks, with both reflected in a mirror., "concepts": woman, mat, mirror, floor, tank top, pants, watch, yoga mat, lying, kneeling, speaking, lifting
"caption": A hand uses a brush to spread blue paint on a surface., "concepts": hand, brush, paint, surface, brushing
"caption": A person walks on a slackline in a park., "concepts": person, slackline, grass, fence, tree, building, shoe, walking
"caption": A man uses a long pole to clean the windows of a building., "concepts": man, building, window, pole, balcony, sky, cleaning
\end{captionconcepts}

\clearpage
\clearpage

\subsection*{C.2. Motion Localization Score}
\label{ssec:sup-gpto3pro}
\addcontentsline{toc}{subsection}{C.2. Motion Localization Score}

\begin{figure*}[h]
    \centering
   \includegraphics[width=\linewidth]{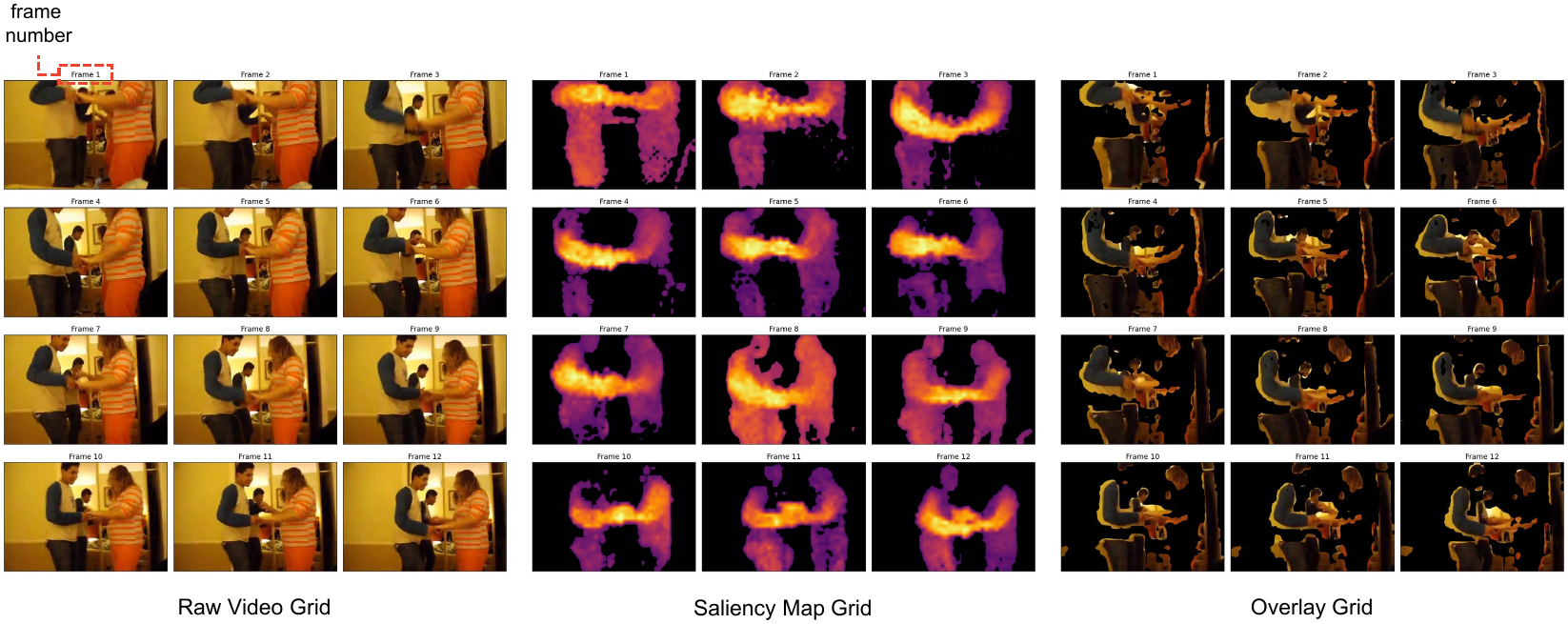}
   \caption{Example of the LLM input grid image
   }
   \label{fig:supp-evalgrid}
\end{figure*}

\noindent
Evaluating motion localization maps is a nontrivial task. To manage this, we employ the state-of-the-art large language model (LLM) OpenAI o3-pro~\cite{o3-pro}. For each video, we uniformly sample 12 frames and use them for evaluation. The 12 video frames and their corresponding saliency maps are then merged into a single grid image. Specifically, as illustrated in Fig.~\ref{fig:supp-evalgrid}, the merged image consists of 12 raw video frames on the left, 12 saliency maps in the center, and 12 overlay images on the right, all arranged in a grid. This design is motivated by two factors: OpenAI o3-pro accepts image-level inputs, and this grid-based representation enables computationally efficient evaluation. The overlay strategy is inspired by IVM~\cite{IVM}. The resulting grid image is provided to the LLM together with the instruction in \boxlabel{Evaluation Prompt}.

\ourparagraph{Metrics.}
We use five evaluation metrics:
\begin{itemize}[noitemsep, nolistsep, leftmargin=*]
    \item Spatial Localization (SL): Activations are spatially on the correct moving parts (e.g., the interacted object), rather than on the static background.
    \item Temporal Localization (TL): Activations should turn ON on the parts that are moving in that frame and stay OFF when those parts are not moving.
    \item Prompt Relevance (PR): Action-semantic alignment: activations concentrate on prompt-specified causal regions (actors' effectors and interacted objects), not static background or camera-only motion.
    \item Specificity/Sparsity (SS): Activations stay confined to necessary regions; penalize background leakage, broad spread, or near-uniform coverage across frames.
    \item Objectness/Boundary quality (OBJ): Contiguous blobs that adhere to human/object silhouettes with clean edges; penalize speckles, holes, disconnected fragments, and texture-tracing patterns.
\end{itemize}

\vspace{3pt}
For each metric, we score the output on a $1$–$5$ scale based on the number of frames that satisfy the criterion and the strength with which they do so. The scores $1$, $2$, $3$, $4$, and $5$ are then mapped to $0$, $0.25$, $0.5$, $0.75$, and $1.0$, respectively, so that each metric lies between $0$ and $1$.
The full evaluation prompt is provided in \boxlabel{Evaluation Prompt}. We set the temperature to 0 to ensure deterministic scoring.

\clearpage
\begin{evaluationprompt}
<System role="video-activation-judge">
    <Goal>
        Judge how well the provided activation maps capture motion dynamics in a 12-frame video.
        You MUST return a single XML object conforming to the OutputSchema. No extra text or JSON.
    </Goal>

    <Inputs>
        <Image id="original_grid" desc="12-frame tiled originals">REQUIRED</Image>
        <Image id="overlay_grid" desc="activation overlaid on originals">REQUIRED</Image>
        <Image id="heatmap_grid" desc="activation-only; treat intensity only, ignore color semantics">REQUIRED</Image>
        <Text id="prompt" desc="short description of actors/action/scene">REQUIRED</Text>
    </Inputs>

    <Rubric scale="1-5 integer">
        <SL>
            Definition:
                Spatial Localization: Are activations spatially on the correct moving parts (e.g., interacted object), not on static background?
            Scoring:
                1: mostly static background; moving parts rarely highlighted (>=9/12 frames wrong).
                2: background-heavy; key moving parts often missing (>=7/12 wrong).
                3: mixed; about half frames correct (≈6±2/12).
                4: mostly correct; main moving parts emphasized in >=9/12 frames, leakage small.
                5: highly accurate; >=11/12 frames focused on moving parts with minimal leakage.
        </SL>
        <TL>
            Definition:
                Temporal Localization: Activations should turn ON on the parts that are moving in that frame and stay OFF when those parts are not moving.
                Judge with original_grid + overlay_grid (use the prompt to identify target parts such as left/right foot, hands).
            Criteria:
                - Alignment Count (AC): number of frames (0-12) where activation clearly peaks on the part(s) that are actually moving in that frame.
                - Alternation (optional): if the action has a left/right or phase alternation (e.g., walking), check that the ON part alternates naturally.
            Scoring:
                1: AC <= 3, or activation looks almost the same across frames (no clear ON/OFF).
                2: AC = 4-5, or alternation mostly wrong when applicable.
                3: AC = 6-7 (mixed).
                4: AC = 8-9, and alternation generally correct when applicable.
                5: AC >= 10, and alternation clearly correct when applicable.
        </TL>
        <PR>
            Definition:
                Action-semantic alignment: activations concentrate on prompt-specified causal regions
                (actors' effectors and interacted objects), not static background or camera-only motion.
            Scoring:
                1: focuses on irrelevant/static regions; contradicts the prompt.
                2: weak; recognizes actors but misses action-specific areas.
                3: fair; about half the frames align with the described action.
                4: strong; most frames (>=9/12) emphasize action-relevant regions.
                5: very strong; consistently highlights causal regions of the action.
        </PR>
        <SS>
            Definition:
                Focus & sparsity: activations stay confined to necessary regions; penalize background leakage,
                broad spread, or near-uniform coverage across frames.
            Scoring:
                1: widespread background activation across many frames.
                2: substantial leakage (>=7/12 frames); background comparable to or larger than target.
                3: moderate leakage; mixed with target.
                4: specific; small, sporadic leakage; mostly on targets.
                5: very specific and sparse; activation confined to necessary parts.
        </SS>
        <OBJ>
            Definition:
                Shape/boundary fidelity: contiguous blobs that adhere to human/object silhouettes with clean edges;
                penalize speckles, holes, disconnected fragments, and texture-tracing patterns.
            Scoring:
                1: amorphous blobs; speckles; holes; poor connectivity.
                2: poor shape; frequent boundary leakage; disconnected pieces.
                3: fair shape; rough silhouette with noticeable leaks/holes.
                4: good; conforms to human silhouettes; connected with minor leaks.
                5: very good; coherent limb/torso blobs; crisp boundaries; minimal leakage.
        </OBJ>
        <AVG>Arithmetic mean of SL, TL, PR, SS, OBJ (1-5). Keep two decimals.</AVG>
    </Rubric>

    <Procedure>
        <Step>Use overlay_grid mainly for SL and PR; use heatmap_grid mainly for TL, SS, OBJ; cross-check with original_grid.</Step>
        <Step>Choose 2-5 strongest evidence frames (indices 1-12) and list them.</Step>
        <Step>Do not infer identities or facts outside the images. Temperature should be 0 if configurable.</Step>
        <Step>Return a SINGLE XML object exactly matching OutputSchema. No prose or markdown.</Step>
    </Procedure>

    <OutputSchema>
        <![CDATA[
            <Assessment>
                <Scores>
                    <SL>1|2|3|4|5</SL>
                    <TL>1|2|3|4|5</TL>
                    <PR>1|2|3|4|5</PR>
                    <SS>1|2|3|4|5</SS>
                    <OBJ>1|2|3|4|5</OBJ>
                </Scores>
                <AVG>1.00-5.00</AVG>
                <EvidenceFrames>
                    <Frame>int in [1,12]</Frame>
                    <Frame>...</Frame>
                </EvidenceFrames>
            </Assessment>
        ]]>
    </OutputSchema>
</System>
\end{evaluationprompt}
\clearpage

\subsection*{C.2.1. LLM-Human Correlation Study}
\label{sssec:sup-gpto3pro_llm}
\addcontentsline{toc}{subsection}{C.2.1. LLM-Human Correlation Study}

Motion Localization Score (MLS) evaluates motion localization maps using a frontier LLM, but validating the reliability of this evaluation remains necessary. Therefore, to verify the reliability of MLS, we conduct two experiments to evaluate the alignment between LLM-based judgments (OpenAI o3-pro) and human judgments.

\begin{table}[h]
\setlength{\tabcolsep}{9pt} 
\renewcommand{\arraystretch}{1.3}
\caption{Correlation coefficients between scores from 5 observers and OpenAI o3-pro.
}
\centering
\resizebox{0.5\linewidth}{!}{
    \begin{tabular}{lcccccc}
    \specialrule{0.8pt}{0pt}{2.5pt}
    Metric & SL & TL & PR & SS & OBJ & Avg. \\
    \specialrule{1pt}{1pt}{2pt}
    Pearson's $r$ (PCC) & 0.62 & 0.44 & 0.64 & 0.66 & 0.64 & 0.75 \\
    Spearman's $r_s$ & 0.52 & 0.46 & 0.59 & 0.63 & 0.69 & 0.74 \\
    \specialrule{0.8pt}{0pt}{0pt}
    \end{tabular}
}
\label{tab:supp-correlation_coeff}
\end{table}

\ourparagraph{Correlation Coefficients.} First, we compute score correlations (Pearson correlation coefficient $r$ and Spearman rank correlation coefficient $r_s$) between human and LLM scores to validate the reliability of the scores. Specifically, we use 102 video-map pairs in the MeViS dataset and compute correlation coefficients for 6 MLS components: SL, TL, PR, SS, OBJ, and Avg. Table~\ref{tab:supp-correlation_coeff} shows average correlation coefficients for 5 observers. Both Pearson and Spearman rank correlation coefficient for MLS (i.e., Avg.) exceed 0.7, indicating a strong association. 

\begin{table}[h]
\setlength{\tabcolsep}{9pt} 
\renewcommand{\arraystretch}{1.3}
\caption{Pairwise preference agreement and win rate.
}
\centering
\resizebox{0.35\linewidth}{!}{
    \begin{tabular}{ccc}
    \specialrule{0.7pt}{0pt}{2pt}
    Total & Agreement & Our wins (Win Rate) \\ 
    \specialrule{0.9pt}{1pt}{2pt}
    51 & 44.8 (88\%) & 45.6 (89\%) \\
    \specialrule{0.7pt}{0pt}{0pt}
    \end{tabular}
}
\label{tab:supp-pairwise_prefer}
\end{table}

\ourparagraph{Pairwise Preference Agreement.} 
Second, we calculate the pairwise preference agreement between ConceptAttention~\cite{ConceptAttention} and IMAP based on MLS (Avg.). Using 51 videos from the MeViS dataset, we compute the human–LLM agreement rate and win rate for 5 observers.
Table~\ref{tab:supp-pairwise_prefer} shows that the average agreement ratio among the 5 observers is 0.88, and the win rate is 0.89.

\subsection*{C.3. Motion Localization Baselines}
\label{ssec:sup-baselines}
\addcontentsline{toc}{subsection}{C.3. Motion Localization Baselines}

\noindent\textbf{ViCLIP}~\cite{ViCLIP} is a contrastive multi-modal representation learning model designed to jointly train a spatiotemporal ViT-based video encoder and a CLIP~\cite{CLIP} text encoder on the large-scale video-text dataset InternVid. 
Because ViCLIP mimics the training procedure of CLIP, it inherits the characteristics of CLIP. Since CLIP is trained with the prefix template ``a photo of a~'' for text inputs, using a single word alone as the text prompt does not yield precise saliency maps. Therefore, to obtain motion-related saliency maps, we prepend the prefix ``a motion of~'' to the motion word.

\ourparagraph{DAAM in VideoCrafter2.}
DAAM~\cite{WhatTheDAAM} is a method that aggregates cross-attention features at multiple resolutions in U-Net-based text-to-image diffusion models to obtain attribution maps for various types of tokens, such as nouns, verbs, adjectives, and adverbs. We adapt this approach to the U-Net-based text-to-video diffusion model VideoCrafter2~\cite{VideoCrafter2} in order to obtain attribution maps for desired concept tokens. In this process, we first re-noise the video to diffusion step 30 out of 50 inference steps and then condition on the target concept to compute its attribution map.

\ourparagraph{Cross Attention in Video DiT.}
By applying the concept token stream operation (Sec.~\ref{ssec:prelim-conceptattn}) proposed in ConceptAttention~\cite{ConceptAttention}, we obtain key embeddings $\vk_c$ for the concept tokens. The saliency map based on cross-attention is then computed as follows:
\begin{equation}
    \mathrm{CrossAttention}=\underset{c}{\mathrm{softmax}}\left(\vq_x \vk_c\tran \right).
\end{equation}

\ourparagraph{ConceptAttention in Video DiT.}
We follow ConceptAttention~\cite{ConceptAttention} as proposed in the original paper, where concept-token embeddings are first computed via the attention operation and then used to obtain the ConceptAttention map as follows:
\begin{equation}
    \vh_c = \mathrm{softmax}\left({\vq_{c}\vk_{xc}\tran}/{\sqrt{d}}\right) \vv_{xc},
\end{equation}
\begin{equation}
    \mathrm{ConceptAttention}=\underset{c}{\mathrm{softmax}}\left(\vh_x \vh_c\tran \right).
\end{equation}

\subsection*{C.4. Query-Key Matching Validation Experiment}
\label{ssec:sup-qkmatching}
\addcontentsline{toc}{subsection}{C.4. Query-Key Matching Validation Experiment}

The Query-Key Matching (QK-Matching; $\vq_{f_i}\vk_c\tran$) proposed in Sec.~\ref{ssec:spatial}  selects the visual token that has the highest attention score with respect to a given text key. We evaluate the accuracy of these cross-attention peaks using the VSPW dataset~\cite{VSPW}. Although VSPW provides videos at 480p resolution, the visual token embeddings in the Video DiT cross-attention layers are downsampled to $s_h \times s_w$ spatially and $t_f$ temporally (e.g., $F=13$, $H=30$, $W=45$ in CogVideoX (2B/5B) and HunyuanVideo). This makes it unsuitable to assess accuracy over all object locations. 

Instead, we select two objects per video, one near the center and one near the image boundary (foreground and background objects), and evaluate localization accuracy for these targets. For example, if the cross-attention peak for the foreground object lies within its segmentation mask, we count this as a true positive.
Evaluating the location accuracy of cross-attention peaks in this way yields an overall accuracy of $0.9554$.

\section*{D. More Experiments}
\label{sec:sup-more-exp}
\addcontentsline{toc}{section}{D. More Experiments}

We provide additional fine-grained analyses of each component of IMAP. 
Sec.~\hyperref[ssec:sup-hinorm]{D.1} investigates the selection methods of text-surrogate tokens, and Sec.~\hyperref[ssec:sup-framewise]{D.2} experimentally analyzes frame-wise selection of text-surrogate tokens. 
Sec.~\hyperref[ssec:sup-randomheads]{D.3} studies head selection based on the separation score, and Sec.~\hyperref[ssec:sup-topk]{D.4} examines the effect of the number of selected heads. Finally, Sec.~\hyperref[ssec:sup-nonstatic]{D.5} examines whether our findings remain consistent on data with camera movements.

\subsection*{D.1. Text-Surrogate Token Selection Methods}
\label{ssec:sup-hinorm}
\addcontentsline{toc}{subsection}{D.1. Text-Surrogate Token Selection Methods}

\begin{table}[h]
\setlength{\tabcolsep}{8pt} 
\renewcommand{\arraystretch}{1.3}
\caption{Quantitative analysis of various text-surrogate token selection methods
}
\centering
\resizebox{0.5\linewidth}{!}{%
    \begin{tabular}{lcccccc}
    \specialrule{0.8pt}{0pt}{2.5pt}
    Method & SL & TL & PR & SS & OBJ & Avg. \\
    \specialrule{1pt}{1pt}{2pt}
    HiNorm & 0.47 & 0.34 & 0.50 & 0.41 & 0.43 & 0.43 \\
    \rowcolor{cyan!10} QK-Matching & 0.68 & 0.48 & 0.69 & 0.61 & 0.64 & 0.62 \\
    \specialrule{0.8pt}{0pt}{0pt}
    \end{tabular}
}
\label{tab:supp-hinorm}
\end{table}

\noindent
In Sec.~\ref{ssec:spatial}, the Query-Key Matching (QK-Matching) used to select the text-surrogate token chooses a single visual token, by measuring similarity between visual-token queries and text-token keys. As an alternative, one may instead select the visual token whose embedding has the largest norm among all visual-token embeddings. We refer to this variant as \textit{HiNorm} and quantitatively compare it against QK-Matching. As shown in Table~\ref{tab:supp-hinorm}, HiNorm underperforms QK-Matching on all metrics, indicating that the visual token embedding at the text-surrogate token index does not necessarily have the largest norm.

\subsection*{D.2. Text-Surrogate Token Selection Level}
\label{ssec:sup-framewise}
\addcontentsline{toc}{subsection}{D.2. Text-Surrogate Token Selection Level}

\begin{table}[h]
\setlength{\tabcolsep}{8pt} 
\renewcommand{\arraystretch}{1.3}
\caption{Quantitative analysis of various text-surrogate token selection levels
}
\centering
\resizebox{0.5\linewidth}{!}{%
    \begin{tabular}{lcccccc}
    \specialrule{0.8pt}{0pt}{2.5pt}
    Method & SL & TL & PR & SS & OBJ & Avg. \\
    \specialrule{1pt}{1pt}{2pt}
    Video & 0.48 & 0.31 & 0.50 & 0.41 & 0.44 & 0.43 \\
    \rowcolor{cyan!10} Frame-wise & 0.68 & 0.48 & 0.69 & 0.61 & 0.64 & 0.62 \\
    \specialrule{0.8pt}{0pt}{0pt}
    \end{tabular}
}
\label{tab:supp-framewise}
\end{table}

\noindent
Because QK-Matching is defined via the similarity between visual-token queries and text-token keys, text-surrogate tokens can in principle be chosen either once per frame or once per video. In our main setting, we select one text-surrogate token per frame (Frame-wise) to account for frame-wise feature variation, but we also compare this against the alternative where a single text-surrogate token is chosen per video (Video). As shown in Table~\ref{tab:supp-framewise}, the ability to select tokens on a frame-wise basis provides a clear advantage.

\clearpage

\subsection*{D.3. Motion Head Selection}
\label{ssec:sup-randomheads}
\addcontentsline{toc}{subsection}{D.3. Motion Head Selection}

\begin{table}[h]
\setlength{\tabcolsep}{7pt} 
\renewcommand{\arraystretch}{1.3}
\caption{Quantitative results of various motion head selection methods
}
\centering
\resizebox{0.6\linewidth}{!}{%
    \begin{tabular}{lccccccc}
    \specialrule{0.8pt}{0pt}{2.5pt}
    Method & \# heads & SL & TL & PR & SS & OBJ & Avg. \\
    \specialrule{1pt}{1pt}{2pt}
    None (all heads) & 48 & 0.47 & 0.34 & 0.48 & 0.48 & 0.50 & 0.46 \\
    \specialrule{0.4pt}{1pt}{1.5pt}
    Random & 5 & 0.50 & 0.30 & 0.51 & 0.44 & 0.47 & 0.44 \\
    \rowcolor{cyan!10} Separation Score & 5 & 0.68 & 0.48 & 0.69 & 0.61 & 0.64 & 0.62 \\
    \specialrule{0.8pt}{0pt}{0pt}
    \end{tabular}
}
\label{tab:supp-randomheads}
\end{table}

\noindent
We select top-$k$ heads with high separability using the separation scores of frame-wise visual-token embeddings. To validate this effect, we compare against randomly selected heads in Table~\ref{tab:supp-randomheads}. While random heads perform similarly to using all heads, employing the separation score yields clear performance gains.

\subsection*{D.4. The Number of Selected Heads}
\label{ssec:sup-topk}
\addcontentsline{toc}{subsection}{D.4. The Number of Selected Heads}

\begin{table}[h]
\setlength{\tabcolsep}{9pt} 
\renewcommand{\arraystretch}{1.3}
\caption{Quantitative analysis of the number of selected motion heads
}
\centering
\resizebox{0.5\linewidth}{!}{%
    \begin{tabular}{ccccccc}
    \specialrule{0.8pt}{0pt}{2.5pt}
    \# Heads & SL & TL & PR & SS & OBJ & Avg. \\
    \specialrule{1pt}{1pt}{2pt}
    1 & 0.52 & 0.36 & 0.54 & 0.44 & 0.47 & 0.47 \\
    3 & 0.55 & 0.35 & 0.56 & 0.49 & 0.51 & 0.49 \\
    \rowcolor{cyan!10} 5 & 0.68 & 0.48 & 0.69 & 0.61 & 0.64 & 0.62 \\
    10 & 0.55 & 0.34 & 0.55 & 0.48 & 0.51 & 0.49 \\
    \specialrule{0.8pt}{0pt}{0pt}
    \end{tabular}
}
\label{tab:supp-topk}
\end{table}

\noindent
Since IMAP is obtained by averaging over all selected heads as in Eq.~(10), the number of heads selected is crucial. 
Table~\ref{tab:supp-topk} reports experimental results for different numbers of heads used in motion-head selection. 
When too few heads are selected, performance degrades because information from the corresponding layer is underutilized, whereas selecting too many heads dilutes the important motion features.

\subsection*{D.5. Clips with Non-Static Cameras}
\label{ssec:sup-nonstatic}
\addcontentsline{toc}{subsection}{D.5. Clips with Non-Static Cameras}

\begin{figure}[h]
\centering
\begin{minipage}[t!]{0.42\linewidth}
    \centering
    \includegraphics[width=0.65\linewidth]{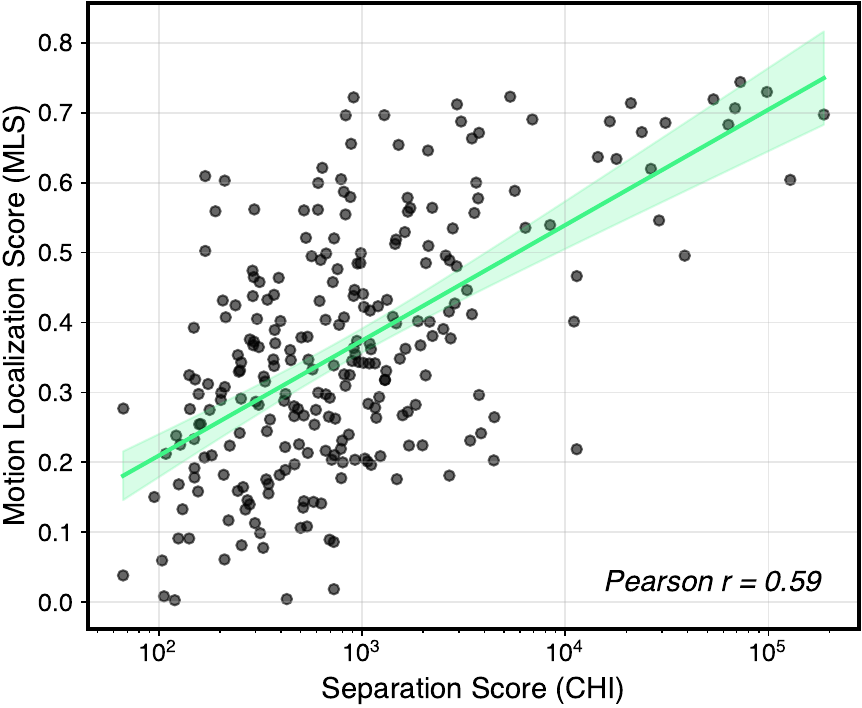}
    \caption{Scatter plot of MLS versus the Calinski–Harabasz index (CHI) for non-static videos.
    }
    \label{fig:supp-nonstatic_pcc}
\end{minipage}
\hfill
\begin{minipage}[H]{0.54\linewidth}
    \centering
    \setlength{\tabcolsep}{9pt} 
    \renewcommand{\arraystretch}{1.3}
    \captionof{table}{MLS evaluation results on non-static videos
    }
    \label{tab:supp-nonstatic_mls}
    \centering
    \resizebox{\linewidth}{!}{
        \begin{tabular}{lcccccc}
        \specialrule{1pt}{0pt}{1.5pt}
        Method & SL & TL & PR & SS & OBJ & Avg. \\
        \specialrule{1.2pt}{1pt}{1pt}
        Cross Attention & 0.26 & 0.12 & 0.28 & 0.11 & 0.15 & 0.18 \\
        ConceptAttention & 0.39 & 0.31 & 0.41 & 0.35 & 0.29 & 0.35 \\
        \rowcolor{cyan!10} IMAP & 0.53 & 0.43 & 0.48 & 0.51 & 0.55 & 0.50 \\
        \specialrule{1pt}{0pt}{0pt}
        \end{tabular}
    }
\end{minipage}
\end{figure}

\noindent
Our preprocessed MeViS dataset contains only static clips for evaluating object-centric motion localization (Sec.~\hyperref[ssec:sup-mevis]{C.1}). However, in real-world videos, camera movements such as zoom, pan, and dolly are common. Therefore, we additionally analyze and evaluate our method on videos with non-static cameras. Accordingly, we extend our evaluation to the 225 MeViS clips filtered out in Sec.~\hyperref[ssec:sup-mevis]{C.1}, which include various camera movements such as zoom, pan, and dolly.

First, following Sec.~\ref{ssec:temporal}, we randomly sample 1,500 \ourmethod{1} from attention heads and analyze the correlation between MLS and the Calinski–Harabasz index (CHI). As in the static-video setting, Fig.~\ref{fig:supp-nonstatic_pcc} shows a positive correlation between MLS and CHI, with a Pearson correlation coefficient of 0.59. Second, Table~\ref{tab:supp-nonstatic_mls} shows MLS evaluation results, which are consistent with results of static videos.

\clearpage

\section*{E. Cross-Attention-Based Video DiTs}
\addcontentsline{toc}{section}{E. Cross-Attention-Based Video DiTs}

Our IMAP does not rely on the concept-token stream (Sec.~\ref{ssec:prelim-conceptattn}). 
QK-Matching uses the query-key similarity matrix from the cross-attention operation, and motion-head selection based on the separation score is likewise independent of concept tokens. 
Therefore, IMAP can also be applied directly to verb tokens in the prompt.
This is why IMAP is applicable not only to joint-attention–based Video DiTs~\cite{CogVideoX,Hunyuanvideo} but also to cross-attention–based Video DiTs~\cite{Wan,SANA-video}.

However, because motion words cannot be externally injected, we must first perform part-of-speech tagging on the prompt to identify verb tokens. As a result, methods based on cross-attention Video DiTs cannot be directly compared with existing baselines that digest concept tokens (ViCLIP~\cite{ViCLIP}, DAAM~\cite{WhatTheDAAM}, and methods using joint-attention Video DiTs~\cite{CogVideoX,Hunyuanvideo}). Instead, in Table~\ref{tab:supp-wan}, we report a comparison only against cross-attention aggregation. Following the same protocol as in Sec.~\ref{sec:exp}, we use the MeViS dataset and tag verbs in the captions as motion tokens. For $11$ out of the $504$ videos, no verb token is detected; in these cases, the scores are set to zero.

As shown in Table~\ref{tab:supp-wan}, IMAP in Wan2.1~\cite{Wan} achieves better performance on Wan2.1-14B than on Wan2.1-1.3B, indicating that larger and stronger models enable the extraction of higher-quality saliency maps.

\begin{table}[h]
\setlength{\tabcolsep}{7pt}
\renewcommand{\arraystretch}{1.4}
\caption{Motion localization results on including five metrics
}
\centering
\resizebox{0.6\linewidth}{!}{%
    \begin{tabular}{lccccccc}
    \specialrule{0.8pt}{0pt}{3pt}
    Method & Backbone & SL & TL & PR & SS & OBJ & Avg. \\
    \specialrule{1pt}{2pt}{2pt}
    Cross Attention & Wan2.1-1.3B & 0.49 & 0.32 & 0.45 & 0.43 & 0.45 & 0.43 \\
    \rowcolor{cyan!10} IMAP & Wan2.1-1.3B & 0.50 & 0.33 & 0.52 & 0.40 & 0.40 & 0.43 \\
    \specialrule{0.4pt}{0pt}{2pt}
    Cross Attention & Wan2.1-14B & 0.50 & 0.32 & 0.51 & 0.42 & 0.41 & 0.43 \\
    \rowcolor{cyan!10} IMAP & Wan2.1-14B & \textbf{0.58} & \textbf{0.41} & \textbf{0.60} & \textbf{0.47} & \textbf{0.49} & \textbf{0.51} \\
    \specialrule{0.8pt}{0pt}{0pt}
    \end{tabular}
}
\label{tab:supp-wan}
\end{table}

\section*{F. Discussion}
\addcontentsline{toc}{section}{F. Discussion}

\subsection*{F.1. Video DiT with Decoder-Only LLM}
\label{ssec:sup-sana-video}
\addcontentsline{toc}{subsection}{F.1. Video DiT with Decoder-Only LLM}

\noindent\textbf{SANA-Video}~\cite{SANA-video} is a text-to-video generation model built on an efficient DiT backbone. SANA-Video is a cross-attention–based model rather than a joint-attention one, so external concepts cannot be injected and we must instead use the embeddings corresponding to verb-token positions in the prompt.
Meanwhile, SANA-Video adopts a decoder-only LLM as the text encoder and performs prompt enhancement by prepending the template in \boxlabel{SANA-Video Prompt Enhancement} to the input prompt.\
Therefore, an LLM-based text encoder such as Gemma~2~\cite{Gemma2} generates an ``enhanced prompt'', effectively rewriting the original input.
As a result, it becomes impossible to determine which text token embedding corresponds to which word, which prevents not only cross-attention–based attribution maps but also the application of IMAP. We hope future work will develop methods for extracting interpretable saliency maps that are compatible with such text-encoder architectures.

\begin{sanavideoprompt}
Given a user prompt, generate an 'Enhanced prompt' that provides detailed visual descriptions suitable for video generation. Evaluate the level of detail in the user prompt:
- If the prompt is simple, focus on adding specifics about colors, shapes, sizes, textures, motion, and temporal relationships to create vivid and dynamic scenes.
- If the prompt is already detailed, refine and enhance the existing details slightly without overcomplicating.
Here are examples of how to transform or refine prompts:
- User Prompt: A cat sleeping -> Enhanced: A small, fluffy white cat slowly settling into a curled position, peacefully falling asleep on a warm sunny windowsill, with gentle sunlight filtering through surrounding pots of blooming red flowers.
- User Prompt: A busy city street -> Enhanced: A bustling city street scene at dusk, featuring glowing street lamps gradually lighting up, a diverse crowd of people in colorful clothing walking past, and a double-decker bus smoothly passing by towering glass skyscrapers.
Please generate only the enhanced description for the prompt below and avoid including any additional commentary or evaluations:
User Prompt:
\end{sanavideoprompt}

\clearpage


\subsection*{F.2. Feature-Similar but Conceptually Irrelevant Tokens}
\label{ssec:sup-feature_similar}
\addcontentsline{toc}{subsection}{F.2. Feature-Similar but Conceptually Irrelevant Tokens}

\begin{figure*}[h]
  \centering
   \includegraphics[width=0.8\linewidth]{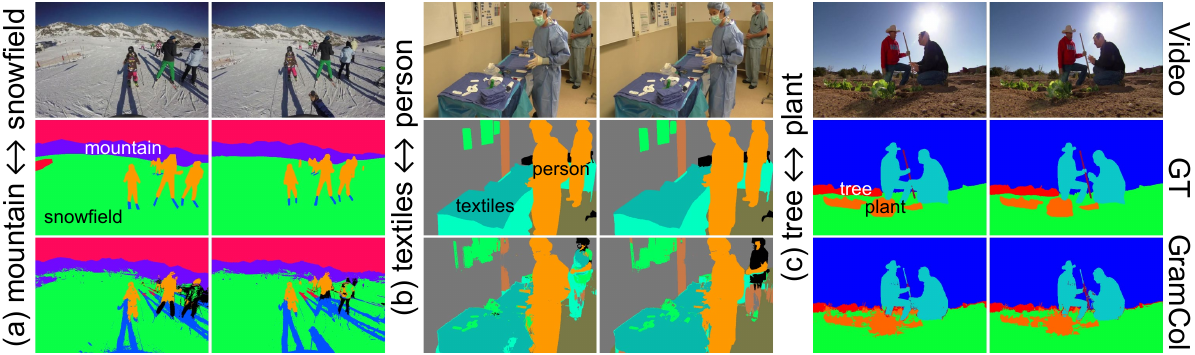}
   \caption{Feature-smilar but conceptually irrelevant cases (Repeated textures)
   }
   \label{fig:supp-similar-1}
\end{figure*}

\begin{figure*}[h]
  \centering
   \includegraphics[width=0.7\linewidth]{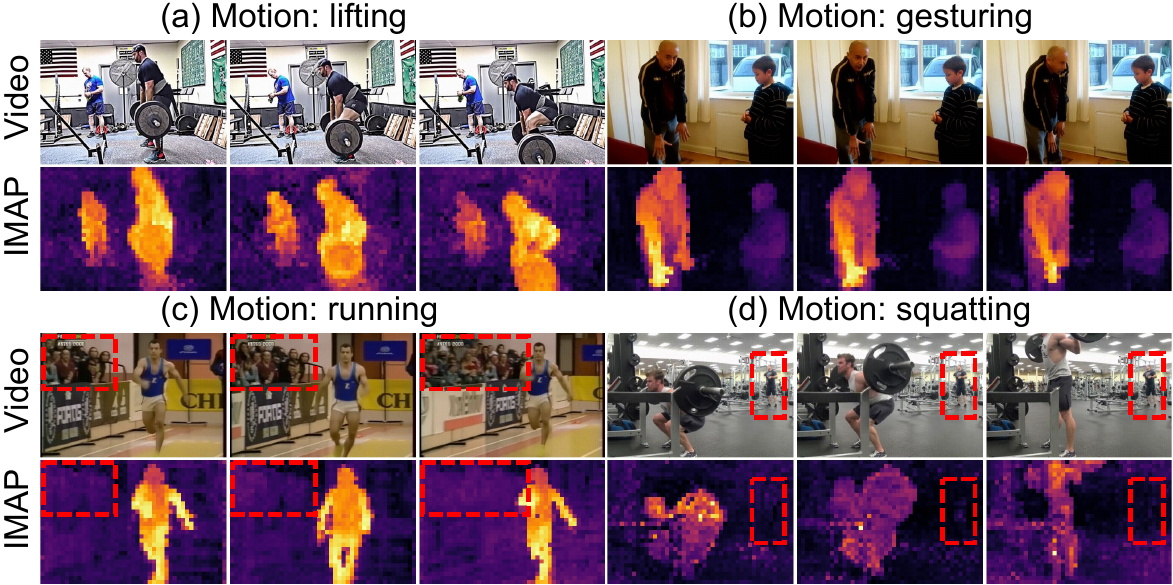}
   \caption{Feature-smilar but conceptually irrelevant cases (Different motions)
   }
   \label{fig:supp-similar-2}
\end{figure*}

Since our method uses a similarity map based on a single text-surrogate visual token, there may be a concern that the resulting saliency map could highlight regions that are feature-similar but conceptually irrelevant. To address this concern, we closely examine IMAP in two cases: (i) repeated textures across different objects, resulting in feature similarity (Fig.~\ref{fig:supp-similar-1}) and (ii) feature-similar objects exhibiting different motions (Fig.~\ref{fig:supp-similar-2}).

\noindent\textbf{Repeated Textures across Different Objects.}
We investigate cases involving feature-similar but conceptually irrelevant objects in the VSPW dataset. As shown in Fig.~\ref{fig:supp-similar-1}, \ourmethod{1} clearly distinguishes such objects even under repeated textures. For example, Fig.~\ref{fig:supp-similar-1}a shows that, despite similar snow-like appearances, \ourmethod{1} separates a mountain from a snowfield.

\noindent\textbf{Feature-Similar Objects Exhibiting Different Motions.}
We present IMAP in Fig.~\ref{fig:supp-similar-2} for cases where only one or two semantically identical objects perform the motion. In Fig.~\ref{fig:supp-similar-2}a, since \textit{lifting} can refer to lifting a barbell or a plate, both individuals are highlighted. However, in cases such as Fig.~\ref{fig:supp-similar-2}b-d, where only one object performs the motion, there is little or no highlight on the other objects.

From these observations, IMAP shows that even feature-similar objects can be distinguished when they correspond to different categories or exhibit different motions. This suggests that Video DiTs retain discriminative capacity for such concepts, and we argue that IMAP serves as a tool to examine whether Video DiTs distinctly understand a given concept.

\clearpage

\subsection*{F.3. Failure Cases \& Future Work.}
\label{ssec:sup-future_work}
\addcontentsline{toc}{subsection}{F.3. Failure Cases \& Future Work.}

\begin{figure*}[h]
  \centering
   \includegraphics[width=0.75\linewidth]{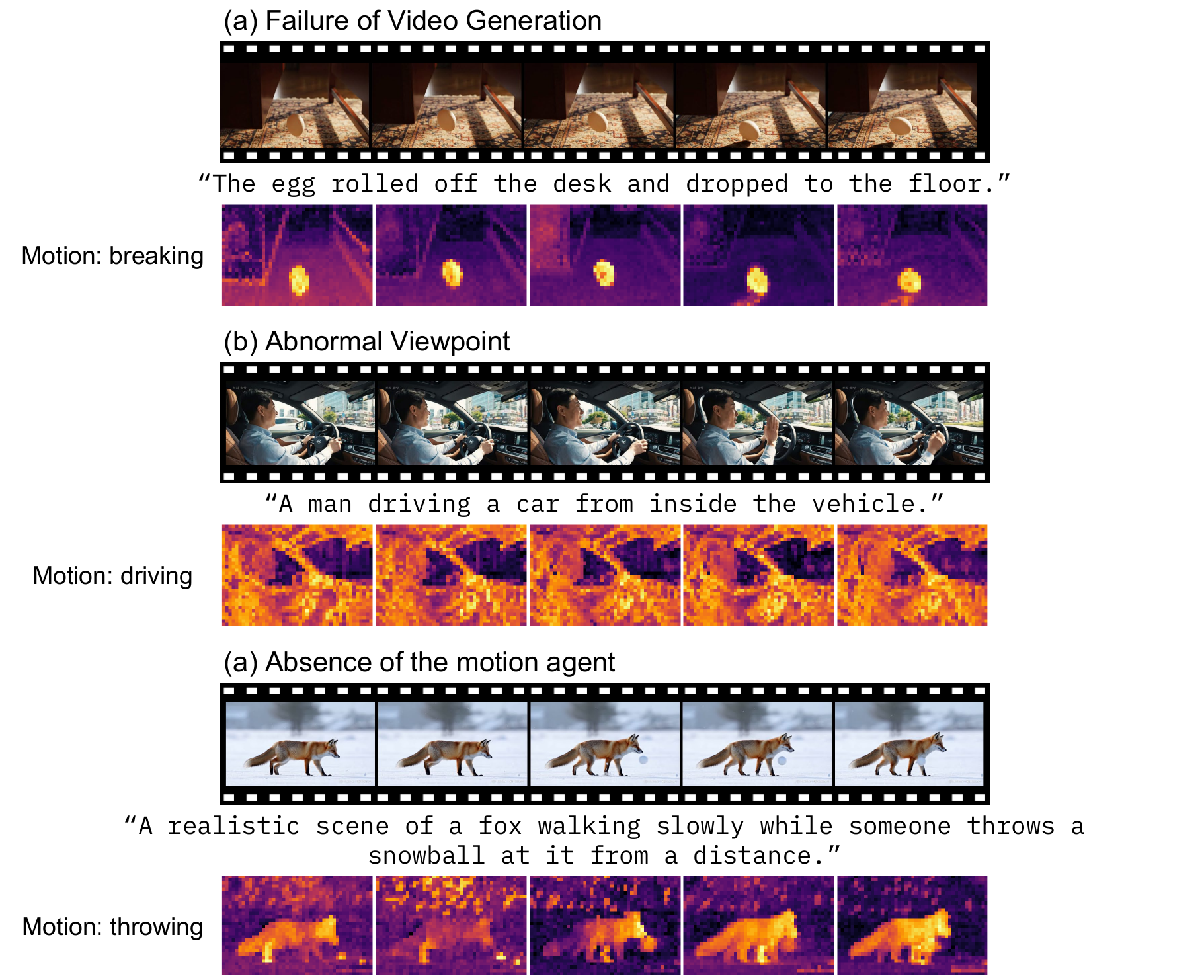}
   \caption{Failure cases. (a) The motion corresponding to the motion concept is not generated during video synthesis. (b) IMAP is extracted for the motion concept from an unconventional viewpoint. (c) The agent performing the motion is outside the camera view.
   }
   \label{fig:supp-failure}
\end{figure*}

\noindent
Since IMAP extracts interpretable saliency maps from pretrained Video DiTs, it naturally inherits failure cases stemming from the generative capability of these models. Fig.~\ref{fig:supp-failure} illustrates three types of such failures. 
\textbf{(1)} Fig.~\ref{fig:supp-failure}a shows a case where the motion corresponding to the motion concept is not generated during video synthesis. Even though the motion does not appear, IMAP still highlights the object that is likely to perform the motion. 
\textbf{(2)} Fig.~\ref{fig:supp-failure}b presents a case where the generated video depicts the motion concept from an unusual viewpoint. For example, although the agent performing \textit{driving} is a person, the video focuses on the car body instead. 
\textbf{(3)} Fig.~\ref{fig:supp-failure}c illustrates a situation in which the agent performing the motion is not generated within the video at all. In such cases, IMAP sometimes highlights the receiver of the motion (e.g., the fox catching the thrown ball), rather than the agent executing the motion (i.e., the thrower).
Although all three scenarios are failure cases of IMAP, we believe that future work can leverage IMAP as a tool for assessing the fidelity of video generation with respect to text descriptions. For example, one could inspect frame-wise variations of IMAP to evaluate whether an object-state change such as ``breaking'' is actually realized in the video. More broadly, we expect IMAP to serve as a means to diagnose and interpret physical fidelity, which remains a key challenge in video generation~\cite{PhysicsIQ}.

\section*{G. More Qualitative Results}
\addcontentsline{toc}{section}{G. More Qualitative Results}

\begin{figure*}[h]
  \centering
   \includegraphics[width=\linewidth]{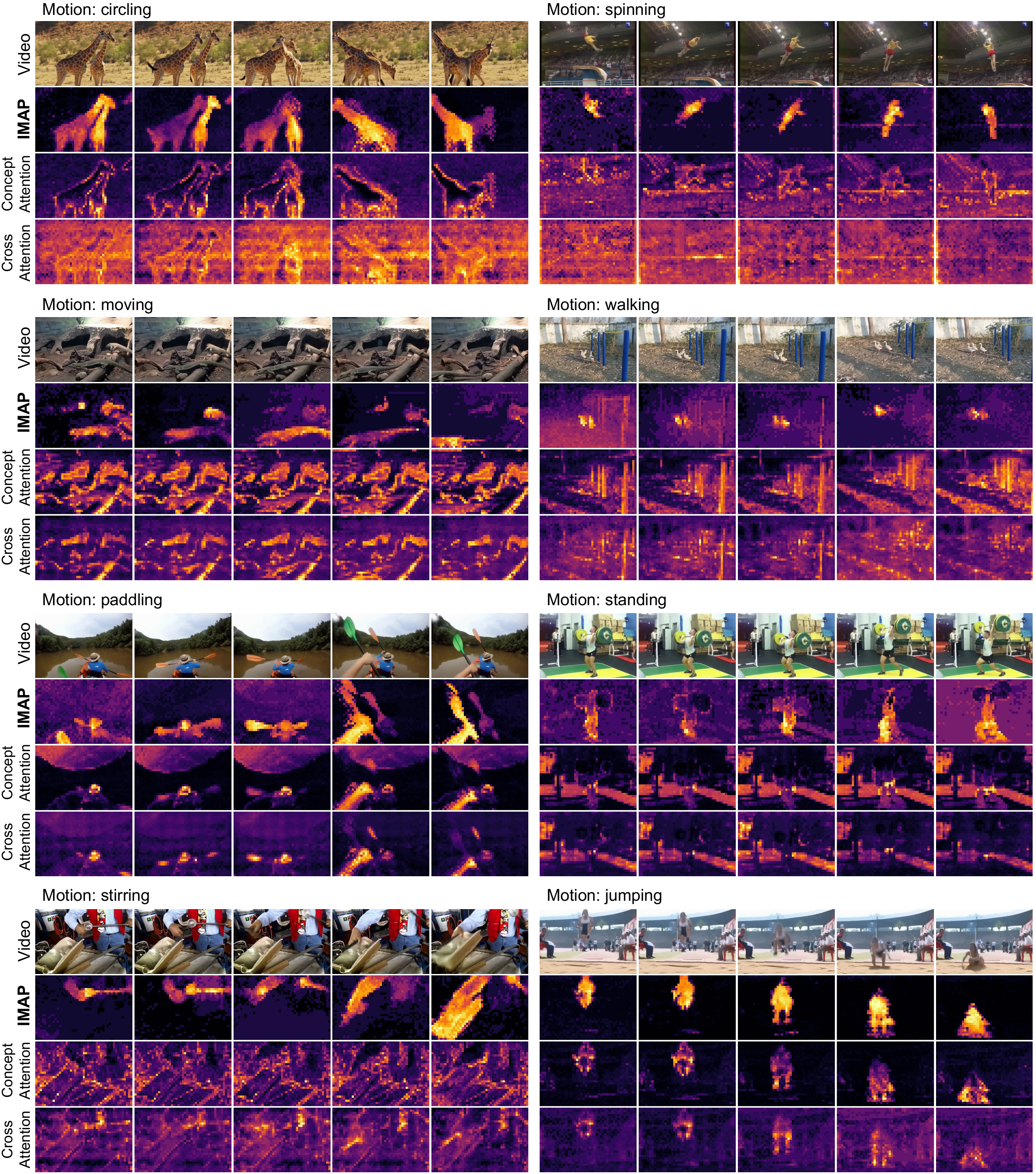}
   \caption{Qualitative comparisons of motion localization results using CogVideoX-5B on the MeViS dataset
   }
   \label{fig:supp-motion-cog5b1}
\end{figure*}

\begin{figure*}[h]
  \centering
   \includegraphics[width=\linewidth]{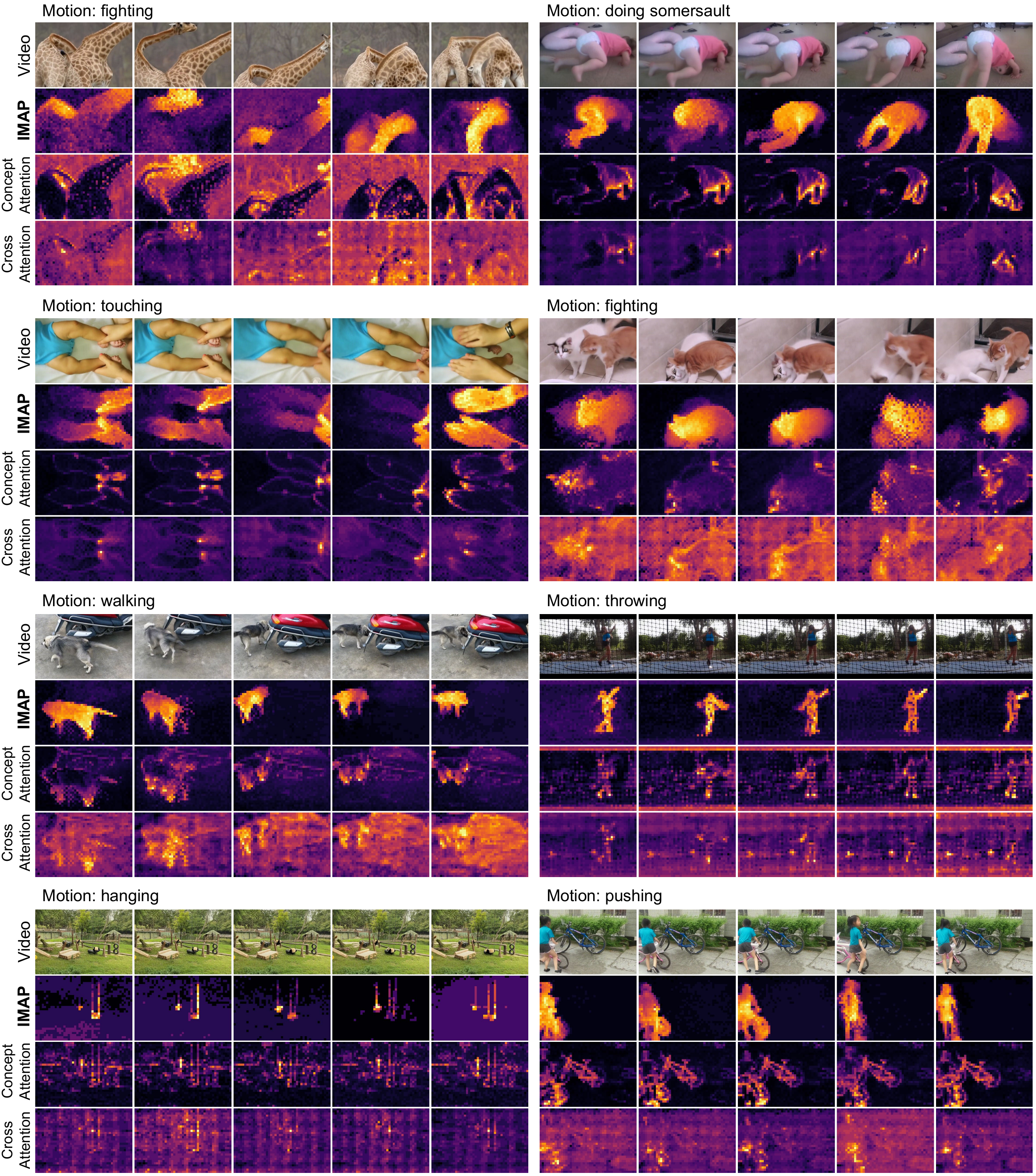}
   \caption{Qualitative comparisons of motion localization results using CogVideoX-5B on the MeViS dataset
   }
   \label{fig:supp-motion-cog5b2}
\end{figure*}

\begin{figure*}[h]
  \centering
   \includegraphics[width=\linewidth]{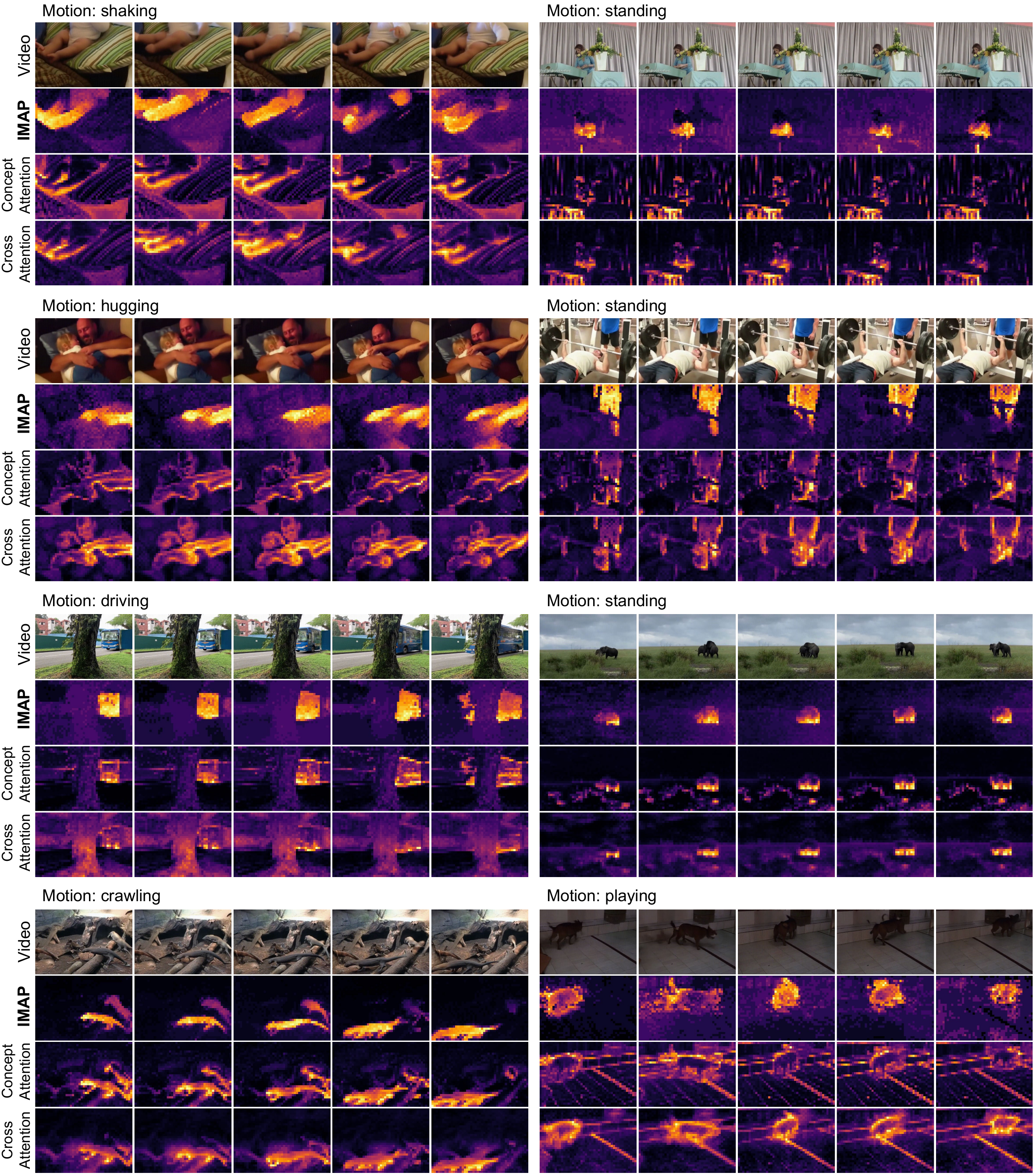}
   \caption{Qualitative comparisons of motion localization results using CogVideoX-2B on the MeViS dataset
   }
   \label{fig:supp-motion-cog2b}
\end{figure*}

\begin{figure*}[h]
  \centering
   \includegraphics[width=\linewidth]{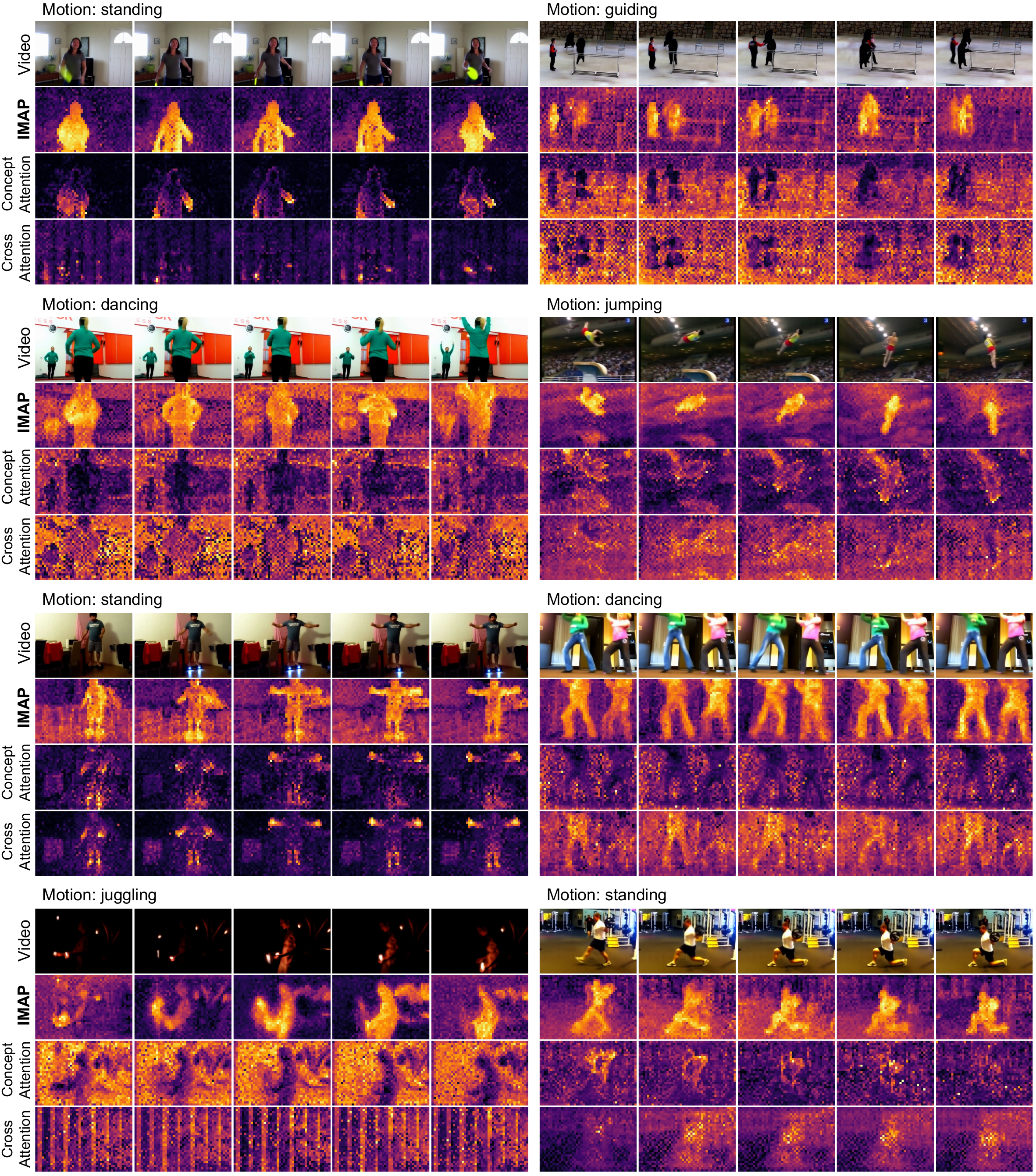}
   \caption{Qualitative comparisons of motion localization results using HunyuanVideo on the MeViS dataset
   }
   \label{fig:supp-motion-hunyuan}
\end{figure*}

\begin{figure*}[h]
  \centering
   \includegraphics[width=\linewidth]{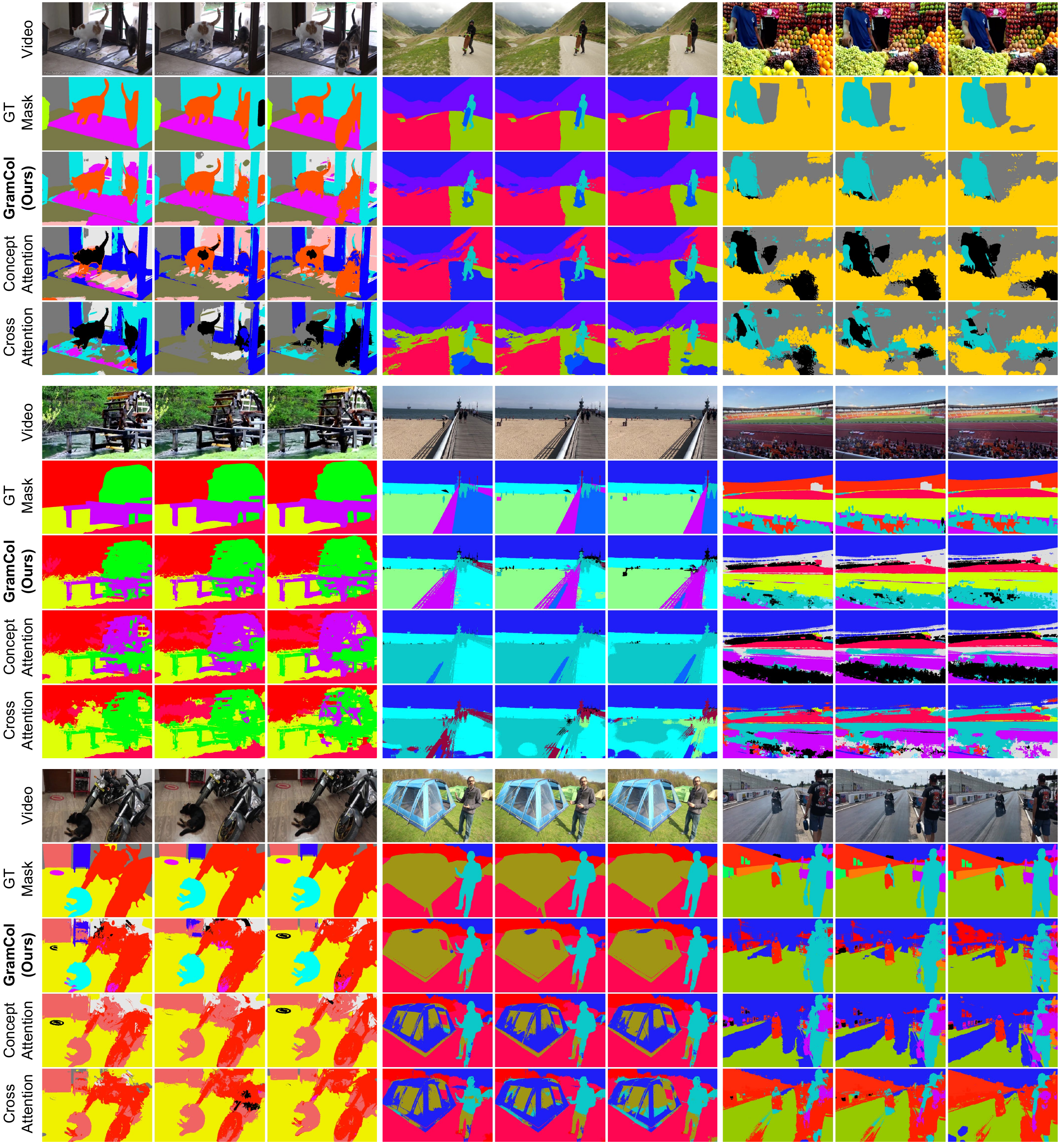}
   \caption{Qualitative comparisons of zero-shot video semantic segmentation results using CogVideoX-5B on the VSPW dataset
   }
   \label{fig:supp-vspw-1}
\end{figure*}

\begin{figure*}[h]
  \centering
   \includegraphics[width=\linewidth]{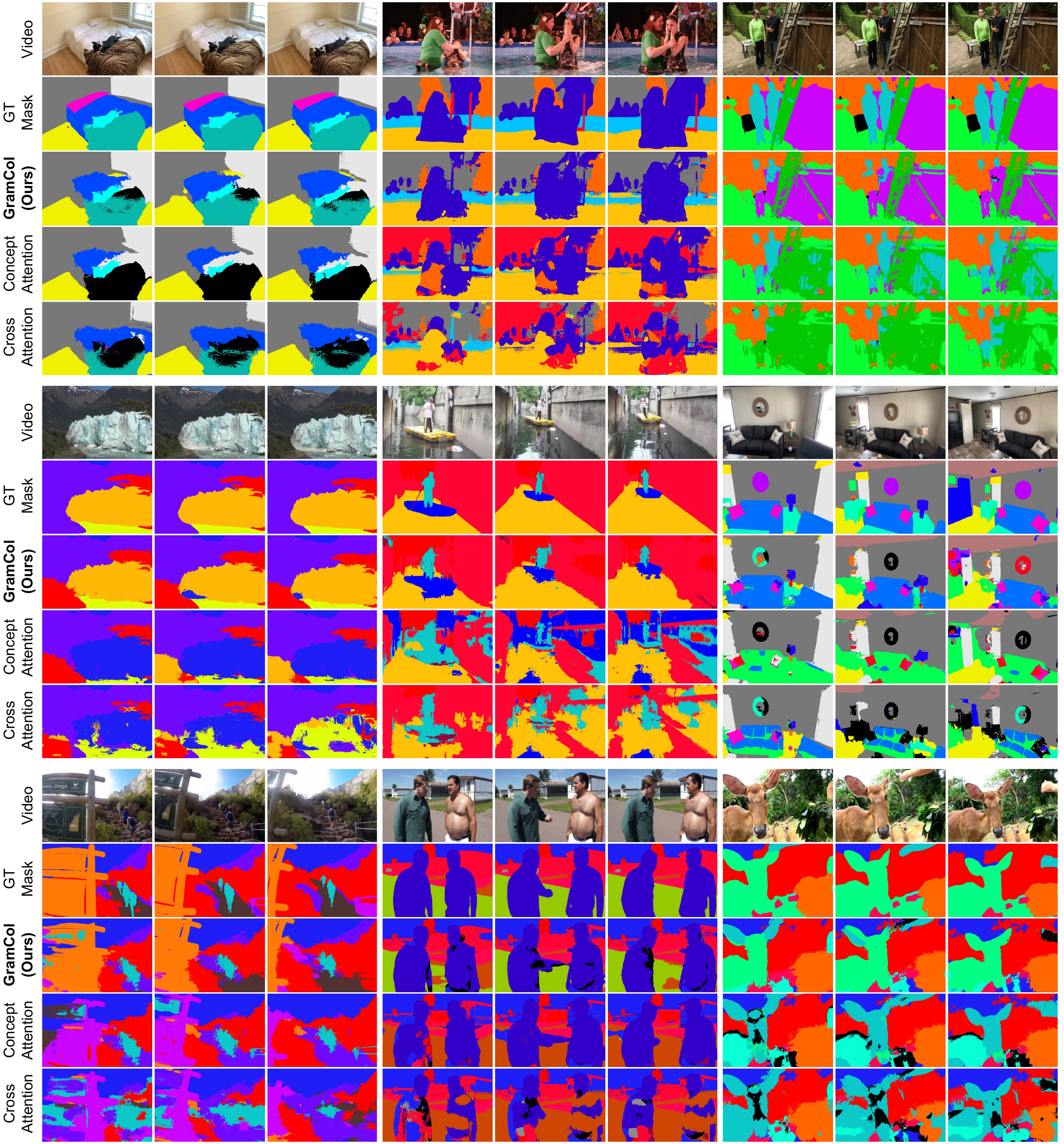}
   \caption{Qualitative comparisons of zero-shot video semantic segmentation results using CogVideoX-5B on the VSPW dataset
   }
   \label{fig:supp-vspw-2}
\end{figure*}

For more comprehensive comparisons, we provide additional IMAP examples: 
CogVideoX-5B in Fig.~\ref{fig:supp-motion-cog5b1} and Fig.~\ref{fig:supp-motion-cog5b2},  
CogVideoX-2B in Fig.~\ref{fig:supp-motion-cog2b},  
HunyuanVideo in Fig.~\ref{fig:supp-motion-hunyuan}.
We also include further examples for the zero-shot video semantic segmentation results in Sec.~\ref{ssec:vss} in Fig.~\ref{fig:supp-vspw-1} and Fig.~\ref{fig:supp-vspw-2}.

\clearpage

\clearpage

\clearpage

\clearpage

\clearpage

\clearpage

\end{document}